\documentclass{article}

\PassOptionsToPackage{numbers, compress}{natbib}

\PassOptionsToPackage{sort}{natbib}

\usepackage[preprint]{Styles/neurips_2026}

\usepackage[utf8]{inputenc} 
\usepackage[T1]{fontenc}    
\usepackage{hyperref}       
\usepackage{url}            
\usepackage{booktabs}       
\usepackage{amsfonts}       
\usepackage{nicefrac}       
\usepackage{microtype}      
\usepackage{xcolor}         

\title{Affine-invariant Cubic Newton with Weak Learners}

\author{%
  \normalfont
  \begin{tabular*}{\dimexpr\textwidth-2\tabcolsep\relax}{@{}@{\extracolsep{\fill}}cccc@{}}
    \textbf{Nikita Zozoulenko}\textsuperscript{1,*,$\dagger$} &
    \textbf{Daniel Falkowski}\textsuperscript{2,*} &
    \textbf{Thomas Cass}\textsuperscript{1} &
    \textbf{Lukas Gonon}\textsuperscript{1,3}
  \end{tabular*} \\[1ex]
  {\small \textsuperscript{1}Imperial College London \qquad
  \textsuperscript{2}Stockholm University \qquad
  \textsuperscript{3}University of St.\ Gallen} \\[0.5ex]
  {\small *Equal contribution \qquad \textsuperscript{$\dagger$}Corresponding author:
  \texttt{n.zozoulenko23@imperial.ac.uk}}
}

\usepackage{amsmath, amssymb, amsthm}
\usepackage{hyperref}       
\usepackage[capitalize,nameinlink]{cleveref}
\usepackage[most]{tcolorbox}

\usepackage{algorithm}
\usepackage{algorithmic}

\usepackage{placeins}
\usepackage{caption}
\definecolor{mycyan}{RGB}{0, 160, 180}   
\definecolor{myblue}{RGB}{0, 0, 150}     
\definecolor{mypurple}{RGB}{120, 0, 150} 
\definecolor{myred}{RGB}{210, 0, 60}     
\definecolor{myorange}{RGB}{200, 80, 0}  
\definecolor{mygray}{RGB}{100, 100, 100} 
\definecolor{proofgray}{RGB}{245, 245, 245}

\newcounter{sharedthm}[section]
\renewcommand{\thesharedthm}{\thesection.\arabic{sharedthm}}

\crefname{definition}{Definition}{Definitions}
\crefname{proposition}{Proposition}{Propositions}
\crefname{theorem}{Theorem}{Theorems}
\crefname{lemma}{Lemma}{Lemmas}
\crefname{assumption}{Assumption}{Assumptions}
\crefname{remark}{Remark}{Remarks}
\crefname{example}{Example}{Examples}

\tcbset{
    basebox/.style={
        enhanced, breakable, frame hidden,
        left=3pt, right=0pt, top=-3pt, bottom=-3pt,
        attach title to upper={\ }, 
        fonttitle=\bfseries,
        sharp corners,
        colback=white,    
        colframe=white,   
    }
}

\newtcolorbox{innerdef}[1][]{
    basebox, coltitle=mycyan, borderline west={2pt}{0pt}{mycyan},
    code={\crefalias{sharedthm}{definition}\refstepcounter{sharedthm}},
    #1
}
\newtcolorbox{innerprop}[1][]{
    basebox, coltitle=myblue, borderline west={2pt}{0pt}{myblue},
    code={\crefalias{sharedthm}{proposition}\refstepcounter{sharedthm}},
    #1
}
\newtcolorbox{innerthm}[1][]{
    basebox, coltitle=myblue, borderline west={2pt}{0pt}{myblue},
    code={\crefalias{sharedthm}{theorem}\refstepcounter{sharedthm}},
    #1
}
\newtcolorbox{innercor}[1][]{
    basebox, coltitle=myblue, borderline west={2pt}{0pt}{myblue},
    code={\crefalias{sharedthm}{corollary}\refstepcounter{sharedthm}},
    #1
}
\newtcolorbox{innerlem}[1][]{
    basebox, coltitle=mypurple, borderline west={2pt}{0pt}{mypurple},
    code={\crefalias{sharedthm}{lemma}\refstepcounter{sharedthm}},
    #1
}
\newtcolorbox{innerass}[1][]{
    basebox, coltitle=myred, borderline west={2pt}{0pt}{myred},
    code={\crefalias{sharedthm}{assumption}\refstepcounter{sharedthm}},
    #1
}
\newtcolorbox{innerrem}[1][]{
    basebox, coltitle=myorange, borderline west={2pt}{0pt}{myorange},
    code={\crefalias{sharedthm}{remark}\refstepcounter{sharedthm}},
    #1
}
\newtcolorbox{innerex}[1][]{
    basebox, coltitle=mygray, borderline west={2pt}{0pt}{mygray},
    code={\crefalias{sharedthm}{example}\refstepcounter{sharedthm}},
    #1
}

\newenvironment{definition}[1][]
  {\if\relax\detokenize{#1}\relax
     \begin{innerdef}[title={Definition \thesharedthm.}]
   \else
     \begin{innerdef}[title={Definition \thesharedthm\ (#1).}]
   \fi}
  {\end{innerdef}}

\newenvironment{proposition}[1][]
  {\if\relax\detokenize{#1}\relax
     \begin{innerprop}[title={Proposition \thesharedthm.}]
   \else
     \begin{innerprop}[title={Proposition \thesharedthm\ (#1).}]
   \fi}
  {\end{innerprop}}

\newenvironment{theorem}[1][]
  {\if\relax\detokenize{#1}\relax
     \begin{innerthm}[title={Theorem \thesharedthm.}]
   \else
     \begin{innerthm}[title={Theorem \thesharedthm\ (#1).}]
   \fi}
  {\end{innerthm}}

\newenvironment{lemma}[1][]
  {\if\relax\detokenize{#1}\relax
     \begin{innerlem}[title={Lemma \thesharedthm.}]
   \else
     \begin{innerlem}[title={Lemma \thesharedthm\ (#1).}]
   \fi}
  {\end{innerlem}}

\newenvironment{assumption}[1][]
  {\if\relax\detokenize{#1}\relax
     \begin{innerass}[title={Assumption \thesharedthm.}]
   \else
     \begin{innerass}[title={Assumption \thesharedthm\ (#1).}]
   \fi}
  {\end{innerass}}

\newenvironment{remark}[1][]
  {\if\relax\detokenize{#1}\relax
     \begin{innerrem}[title={Remark \thesharedthm.}]
   \else
     \begin{innerrem}[title={Remark \thesharedthm\ (#1).}]
   \fi}
  {\end{innerrem}}

\newenvironment{example}[1][]
  {\if\relax\detokenize{#1}\relax
     \begin{innerex}[title={Example \thesharedthm.}]
   \else
     \begin{innerex}[title={Example \thesharedthm\ (#1).}]
   \fi}
  {\end{innerex}}

\DeclareMathOperator*{\argmax}{argmax}
\DeclareMathOperator*{\argmin}{argmin}

\usepackage{graphicx}
\usepackage{subcaption}
\usepackage{tcolorbox}
\tcbuselibrary{skins,breakable}
\definecolor{codebg}{HTML}{FBFBFD}
\definecolor{codeframe}{HTML}{D0D7DE}

\tcbset{
  codebox/.style={
    colback=codebg,
    colframe=codeframe,
    boxrule=0.4pt,
    arc=2mm,
    left=2mm,right=2mm,top=1mm,bottom=1mm,
    enhanced, breakable,
  }
}
\usepackage{listings}
\definecolor{tangoKeyword}{HTML}{204A87}
\definecolor{tangoString}{HTML}{4E9A06}
\definecolor{tangoComment}{HTML}{8F5902}
\definecolor{tangoBuiltin}{HTML}{3465A4}

\lstdefinestyle{tango}{
    language=Python,
    basicstyle=\ttfamily\scriptsize,
    keywordstyle=\color{tangoKeyword}\bfseries,
    stringstyle=\color{tangoString},
    commentstyle=\color{tangoComment}\itshape,
    morestring=[s]{"""}{"""},
    emph={int,float},
    emphstyle=\color{tangoBuiltin},
    breaklines=true,
    columns=fullflexible,
    keepspaces=true,
    showstringspaces=false,
    tabsize=4
}

\begin{document}

\maketitle

\begin{abstract}
    Modern gradient boosting algorithms are based on Newton's method, which excels empirically but lacks global convergence guarantees. While recent work on Gradient-Regularized Newton (GRN) boosting achieved a global $\mathcal{O}(1/\gamma^{12}k^2)$ rate, its analysis was not able to exploit the natural Hessian geometry and suffers from a heavy $\gamma^{12}$ dependence on a worst-case `weak gradient edge' $\gamma$. In this paper, we tackle these theoretical drawbacks by generalizing the Affine-Invariant Cubic Newton (AICN) scheme to convex optimization with weak learners. Our analysis is based on average cosine values $\overline{\Theta}_k$ along the optimization path and directional improvements $\delta_k$ in the Hessian-induced norm. We show that the weak-AICN update reduces to a damped weak-Newton step with an explicit stepsize determined by $\delta_k$. Under semi-strong self-concordance, we establish a global $\mathcal{O}(1/\overline{\Theta}_k^4 k^2)$ rate, significantly improving upon the weak-GRN bound. We further prove a local linear rate, global linear convergence for Hessian-dominated losses, cosine angle lower bounds, and present weak-AdaAICN, which adaptively tunes the cubic regularization. Finally, we demonstrate the broader applicability of this framework beyond gradient boosting by showing that Newton coordinate descent emerges naturally as a special case of optimization with weak learners.
\end{abstract}

\section{Introduction}
Modern gradient boosting decision tree (GBDT) libraries such as XGBoost \cite{2016XGBoost}, LightGBM \cite{2017LightGBM}, and CatBoost \cite{2018CatBoost} use Newton's method to fit leaf weights. However, theoretical understanding of Newton boosting \cite{2000LogitBoostAdditiveLogisticRegressionAStatisticalViewOfBoosting,2014AConvergenceRateAnalysisForLogitBoostMARTAndTheirVariant,2020SnapBoostAHeterogeneousBoostingMachine,2026GradientRegularizedNewtonBoostingTreeswithGlobalConvergence} has remained limited in comparison to first-order gradient boosting \cite{1999BoostingAlgorithmsAsGradientDescent, 2000LogitBoostAdditiveLogisticRegressionAStatisticalViewOfBoosting,2006SomeTheoryForGeneralizedBoostingAlgorithms,2011GeneralizedBoostingAlgorithmsForConvexOptimization,2012APrimalDualConvergenceAnalysisOfBoosting,2017ANewPerspectiveOnBoostingInLinearRegressionViaSubgradientOptimization,2019RegularizedGradientBoosting,2020RandomizedGradientBoostingMachine,2020AcceleratingGradientBoostingMachines}. This stems from the fact that Newton's method does not have global convergence guarantees, and can diverge when initialized far from the optimum, even for smooth, strictly convex losses \cite{2007OnTheDivergenceOfLineSearchMethods,2016SimpleExamplesForTheFailureOfNewtonsMethodWithLineSearchForStrictlyConvexMinimization}. Recently \cite{2026GradientRegularizedNewtonBoostingTreeswithGlobalConvergence} proved the first global $O(1/k^2)$ rate for Newton boosting by extending the Gradient-Regularized Newton (GRN) \cite{2023RegularizedNewtonMethodWithGlobalConvergence,2024GradientRegularizationOfNewtonMethodWithBregmanDistances} scheme to weak learners. They did this via the Restricted Newton Descent framework \cite{2026GradientRegularizedNewtonBoostingTreeswithGlobalConvergence}, which poses boosting as a special case of Hilbert space convex optimization with iterates restricted to a scalable set of weak learners.

While the weak-GRN scheme successfully globalizes Newton boosting, its convergence analysis was not able to take advantage of the natural Hessian geometry of the problem. The weak-GRN bounds rely on standard Lipschitz continuity of the Hessian, and the proofs are fundamentally based on bounding the growth of Euclidean gradient norms \cite{2026GradientRegularizedNewtonBoostingTreeswithGlobalConvergence}.
Because of this geometric mismatch, the analysis does not depend on any prior established definition of weak learner edge or cosine angle \cite{2011GeneralizedBoostingAlgorithmsForConvexOptimization,2014AConvergenceRateAnalysisForLogitBoostMARTAndTheirVariant,2020RandomizedGradientBoostingMachine}. Instead, \cite{2026GradientRegularizedNewtonBoostingTreeswithGlobalConvergence} introduced the ``weak gradient edge'' $\gamma$, which measures the similarity between the exact gradient and the implied weak gradient in the Euclidean norm. 
This results in a severe $\mathcal{O}(1/\gamma^{12}k^2)$ dependence in the convergence rate, rendering the bound highly pessimistic since $\gamma$ can often be very small when using intentionally weak learners.

In this paper, we resolve these theoretical drawbacks by presenting a different global weak learner Newton scheme, namely Affine-Invariant Cubic Newton \cite{2022ADampedNewtonMethodAchievesGlobalMinus2AndLocalQuadraticConvergenceRate}, generalized to weak learners. Here, regularity is measured using semi-strong self-concordance, which is affine-invariant and allows for the analysis to take place entirely using local Hessian norms. This geometric shift allows us to measure convergence in terms of the cosine angle $\Theta_k$ between weak and exact Newton iterates. Furthermore, previous boosting papers defined their weak learner edges by taking infima over either the entire optimization sequence $k$ \cite{2011GeneralizedBoostingAlgorithmsForConvexOptimization,2014AConvergenceRateAnalysisForLogitBoostMARTAndTheirVariant,2026GradientRegularizedNewtonBoostingTreeswithGlobalConvergence}, or over the set of all possible weak learners \cite{2020RandomizedGradientBoostingMachine,2020AcceleratingGradientBoostingMachines,2020SnapBoostAHeterogeneousBoostingMachine,2012APrimalDualConvergenceAnalysisOfBoosting}. We instead develop a novel recurrence analysis that allows our bounds to depend on the \textit{average} cosine angle $\overline{\Theta}_k$ up to iteration $k$. By shifting away from a worst-case analysis, we theoretically guarantee that a single poorly-fitted weak learner does not invalidate the convergence of the whole scheme.

\textbf{Contributions:} Our main contributions are summarized as follows:
\begin{itemize}
    \item \textbf{Affine-Invariant Weak Updates:} We formulate the weak-AICN algorithm within the Restricted Newton Descent framework. By introducing the directional improvement metric $\delta_k$ in the local Hessian norm, we prove that the weak-AICN update elegantly reduces to a damped weak-Newton step with an explicit, easily computable stepsize.
    \item \textbf{Global Convergence:} Under semi-strong self-concordance, we establish a global $\mathcal{O}(1/\overline{\Theta}_k^4 k^2)$ convergence rate. This drastically improves upon the $\mathcal{O}(1/\gamma^{12} k^2)$ rate of GRN, and transitions from a worst-case bound to an average-performance bound.
    \item \textbf{Linear Rates for Specific Loss Classes:} We prove that the weak-AICN scheme achieves a local linear convergence rate for general convex losses, and a global linear convergence rate for Hessian-dominated losses (such as cross entropy loss).
    \item \textbf{Applications to Coordinate Descent:} By taking the weak learner family to be the standard basis vectors, we show that Newton coordinate descent emerges naturally as a special case, allowing us to apply our weak learner convergence theory beyond simply gradient boosting, to classical coordinate optimization. We also present an adaptive variant of weak-AICN that dynamically tunes the cubic regularization without refitting the weak learners.

    \item \textbf{Tree-Angle Lower Bounds:} For depth-$J$ decision trees, we prove explicit random lower bounds on tree alignment in a random configuration model, including a Beta-distributed stochastic lower bound and the expectation $E\!\left[(\widehat{\Theta}_{n,p,J}^{\mathrm{prox}})^2\right] \ge \frac{\min(2^J,n)}{n}$.
\end{itemize}

\subsection{Related Work}

\textbf{Boosting:}
Gradient boosting was first studied as gradient descent in $\mathbb{R}^N$ \cite{1999BoostingAlgorithmsAsGradientDescent, 2000LogitBoostAdditiveLogisticRegressionAStatisticalViewOfBoosting}, and later generalized to functional spaces \cite{2011GeneralizedBoostingAlgorithmsForConvexOptimization}. The first boosting convergence rate of $\mathcal{O}((\log k)^{-\frac{1}{2}})$ was proved in \cite{2006SomeTheoryForGeneralizedBoostingAlgorithms}, later improved to $\mathcal{O}(1/k)$ under smoothness \cite{2014AConvergenceRateAnalysisForLogitBoostMARTAndTheirVariant}, and to $\mathcal{O}(1/k^2)$ via Nesterov momentum \cite{2020AcceleratingGradientBoostingMachines}. Recently, the $\mathcal{O}(1/k^2)$ rate was also achieved by the weak-GRN boosting scheme \cite{2026GradientRegularizedNewtonBoostingTreeswithGlobalConvergence}. For strongly convex losses, a linear rate was derived in \cite{2011GeneralizedBoostingAlgorithmsForConvexOptimization}, and later by \cite{2020RandomizedGradientBoostingMachine} via coordinate descent by enumerating all possible weak learners. Newton boosting was first introduced by LogitBoost \cite{2000LogitBoostAdditiveLogisticRegressionAStatisticalViewOfBoosting} for logistic loss and later popularized by XGBoost \cite{2016XGBoost} for general losses. Theory for vanilla Newton boosting remains limited, with guarantees only in the strongly convex setting for small learning rates \cite{2014AConvergenceRateAnalysisForLogitBoostMARTAndTheirVariant, 2020SnapBoostAHeterogeneousBoostingMachine, 2026GradientRegularizedNewtonBoostingTreeswithGlobalConvergence}.

\textbf{Newton's method in convex optimization:}
Newton’s method has quadratic local convergence \cite{2018LecturesOnConvexOptimization} but lacks global guarantees \cite{2007OnTheDivergenceOfLineSearchMethods,2016SimpleExamplesForTheFailureOfNewtonsMethodWithLineSearchForStrictlyConvexMinimization}, motivating regularized variants. Cubic regularized Newton achieves global $\mathcal{O}(1/k^2)$ rates under Lipschitz Hessians \cite{2006CubicRegularizationOfNewtonMethodAndItsGlobalPerformance, 2008AcceleratingTheCubicRegularizationOfNewtonsMethodOnConvexProblems}, but requires solving a costly subproblem each iteration. GRN matches this rate without such subproblems via adaptive $\ell_2$ regularization \cite{2023RegularizedNewtonMethodWithGlobalConvergence,2024GradientRegularizationOfNewtonMethodWithBregmanDistances}. More recently, affine-invariant approaches such as AICN \cite{2022ADampedNewtonMethodAchievesGlobalMinus2AndLocalQuadraticConvergenceRate} achieve the same global $\mathcal{O}(1/k^2)$ rate, together with local quadratic convergence, as a damped Newton method with explicit computable stepsize. For a broader historical overview, including earlier suboptimal rates and the development of self-concordance, we refer to \cite{2023RegularizedNewtonMethodWithGlobalConvergence,2022ADampedNewtonMethodAchievesGlobalMinus2AndLocalQuadraticConvergenceRate}. Finally, we want to mention \cite{2024InexactTensorMethodsAndTheirApplicationToStochasticConvexOptimization,2025SimpleStepsizeForQuasiNewtonMethodsWithGlobalConvergenceGuarantees} which studies inexact (quasi-)Newton's method with inexact Hessians, rather than with weak learners.

\section{Setup and Optimization with Exact and Weak Learners}

In this paper we work in a general Hilbert space setting, following \cite{2026GradientRegularizedNewtonBoostingTreeswithGlobalConvergence}. Let $(\mathcal{H}, \|\cdot\|)$ be a real Hilbert space, and $L: \mathcal{H}\to\mathbb{R}$ be a convex and twice differentiable loss function with positive definite Hessian. Let $\eta >0$ be the learning rate (constant dampening factor). The goal of Newton's method is to produce a sequence of estimates $F_k$ minimizing the loss $L(F_k)$. It does this iteratively by minimizing a second order Taylor expansion of the loss at each iteration $k$:
\begin{align*}
    L(F_k + f) &\approx
     L(F_k) + \left\langle f,g_k \right\rangle+ \frac{1}{2}\left\langle f, H_k[f]  \right\rangle 
    =: L(F_k) + Q_k(f),
\end{align*}
where $g_k = \nabla L(F_k)$ is the gradient, $H_k = \nabla^2 L(F_k)$ is the Hessian operator, and $Q_k(f)$ is the second-order approximation of $L$ centered on $F_k$. The Newton step is derived by differentiating $Q_k$ and setting $0 =\nabla Q_k(f) = g_k + H_k f_{k+1}^{\mathrm{N}}$, yielding
\begin{equation}
    f_{k+1}^{\mathrm{N}}
    \ 
    :=
    \ 
    -H_k^{-1}[g_k]
    \
    =
    \argmin_{f\in\mathcal{H}} Q_k(f).
\end{equation}
The exact Newton iterates are then defined as $F_{k+1}^{\mathrm{N}} = F_k^{\mathrm{N}} + \eta f_{k+1}^{\mathrm{N}} = \sum_{t=1}^{k+1} \eta f_t^{\mathrm{N}}$, assuming $F_0^{\mathrm{N}} = 0$.

\subsection{Restricted Newton Descent}

In this paper, we work in the Restricted Newton Descent framework \cite{2026GradientRegularizedNewtonBoostingTreeswithGlobalConvergence}, where the updates $f$ are restricted to a set of weak learners $f \in  \mathcal{F}\subseteq \mathcal{H}$. The exact Newton update $f^{\mathrm{N}}_{k+1}$ was defined by minimizing $Q_k(f)$ over the whole space $\mathcal{H}$. The weak learner $f_{k+1}^{\mathrm{wN}}$ is instead defined as
\begin{equation}
    f_{k+1}^{\mathrm{wN}} \in \argmin_{f\in\mathcal{F}} Q_k(f),
    \qquad \qquad
    Q_k(\mathcal{F}) = \min_{f\in\mathcal{F}} Q_k(f).
\end{equation}
Here $Q_k(\mathcal{F})$ is the minimally achievable weak learner quadratic Taylor loss. For boosting, $\mathcal{F}$ is typically the set of decision trees of a maximum depth. We use the following standard assumption:

\begin{assumption}[Scalable weak learner family \cite{2020AcceleratingGradientBoostingMachines}]\label{ass:scalableWeakLearnerFamily}
The weak-learner family $\mathcal{F}\subseteq \mathcal{H}$ is closed under scalar multiplication:
if $f\in\mathcal{F}$ and $t\in\mathbb{R}$ then $t f\in\mathcal{F}$.
\end{assumption}

\subsection{Hessian norms, cosine angle, and directional improvement}

In this subsection we introduce the basic tools and definitions needed for our weak learner analysis. The notion of local Hessian norm and its dual norm, defined below, will play a central role.

\begin{definition}[Hessian-induced norm and dual norm]\label{def:hessianNorm}
We define the \emph{Hessian-induced inner product and norm} as
\[
\langle u,v\rangle_{H_k} \ :=\ \langle u, H_k v\rangle,
\qquad \qquad
\|u\|_{H_k} \ :=\ \langle u,u\rangle_{H_k}^{1/2} \ =\ \langle H_k u,u\rangle^{1/2},
\]
with corresponding \emph{dual norm} defined by
\[
\|v\|_{H_k}^* \ :=\ \|v\|_{H_k^{-1}} \ =\ \langle H_k^{-1} v, v\rangle^{1/2}.
\]
\end{definition}
\begin{remark}
In a general infinite-dimensional Hilbert space, the notation $H_k^{-1}$ should be interpreted as requiring $H_k$ to be \emph{boundedly invertible} at the points of interest (equivalently, $H_k\succeq \mu I$ for some $\mu>0$
at those points). In finite dimensions this is automatic from $H_k\succ 0$.
\end{remark}

Next, we define the fundamental metrics governing our weak learner convergence bounds. In addition to the cosine angle $\Theta_k$ between exact and weak Newton directions \cite{2026GradientRegularizedNewtonBoostingTreeswithGlobalConvergence}, we introduce a new but related metric, denoted by $\delta_k$. This quantity measures the directional alignment to the exact Newton direction in the local Hessian geometry.

\begin{definition}[Cosine angle and directional improvement]\label{def:CosineAngle}
    Let $f^{\mathrm{N}}_{k+1}$ and  $f^{\mathrm{wN}}_{k+1}$ be the exact and weak Newton directions, respectively, computed at iteration $k$. The weak learner \emph{cosine angle} is defined as
    \begin{equation}
        \Theta_k
        :=
        \frac{\langle f_{k+1}^{\mathrm{N}},f_{k+1}^{\mathrm{wN}}\rangle_{H_k}}{\|f_{k+1}^{\mathrm{N}}\|_{H_k}\,\|f_{k+1}^{\mathrm{wN}}\|_{H_k}} = \operatorname{cosangle}_{\|\cdot\|_{H_k}}(f_{k+1}^{\mathrm{N}}, f_{k+1}^{\mathrm{wN}}).
    \end{equation}
    For any nonzero $f\in \mathcal{H}$, define the \emph{directional improvement} and \emph{maximal directional improvement} as
    \begin{equation}
        \delta_k(f)
        \ :=\
        \frac{-\langle g_k,f\rangle}{\|f\|_{H_k}}
        \ = \
        \left\langle  f_{k+1}^{\mathrm{N}}, \frac{f}{\|f\|_{H_k}} \right\rangle_{H_k},
        \qquad \quad
        \delta_k(\mathcal{F}) = \max_{f \in \mathcal{F} \setminus \{0\}} \delta_k(f).
    \end{equation}
\end{definition}
Before we introduce the weak-AICN scheme, we first need to present a key result related to the weak-Newton update which our analysis depends on. The following proposition shows that, instead of minimizing the quadratic approximation \(Q_k(f)\), we may equivalently maximize the directional improvement \(\delta_k(f)\). We present the proof in \cref{app:QuadraticMinimization}. This differs substantially from the approach in \cite{2022ADampedNewtonMethodAchievesGlobalMinus2AndLocalQuadraticConvergenceRate} as a tool for AICN analysis, as well as the weak-Newton results of \cite{2026GradientRegularizedNewtonBoostingTreeswithGlobalConvergence} which only required elementary properties of cosine angles.

\begin{proposition}[$Q$-$\delta$ duality]\label{prop:deltak_Qk}
For any non-zero direction \(\phi^\ast \in \mathcal{F}\) with
\(\delta_k(\phi^\ast)\geq 0\), we have
\[
\phi^\ast \in \argmax_{f\in\mathcal{F}\setminus\{0\}} \delta_k(f)
\quad\Longleftrightarrow\quad
\frac{\delta_k(\phi^\ast)}{\|\phi^\ast\|_{H_k}}\,\phi^\ast
\in \argmin_{f\in \mathcal{F}} Q_k(f).
\]
In particular,
\(
Q_k(\mathcal{F}) = -\frac12 \delta_k(\mathcal{F})^2.
\)
Moreover, every minimizer \(f_{k+1}^{\mathrm{wN}} \in \argmin_{f\in \mathcal{F}} Q_k(f)\) satisfies
\[
-\langle g_k, f_{k+1}^{\mathrm{wN}} \rangle
=
\|f_{k+1}^{\mathrm{wN}}\|_{H_k}^2,
\qquad \qquad
\|f_{k+1}^{\mathrm{wN}}\|_{H_k}
=
\delta_k(\mathcal{F}).
\]
If \(f_{k+1}^{\mathrm{wN}} \neq 0\), then also
\[
\delta_k(f_{k+1}^{\mathrm{wN}})
=
\|f_{k+1}^{\mathrm{wN}}\|_{H_k}
=
\delta_k(\mathcal{F}).
\]
In particular, we have the identities $\Theta_k \|g_k\|_{H_k}^* = \delta_k(\mathcal{F})$ and $\Theta_k = \frac{\|f_{k+1}^{\mathrm{wN}}\|_{H_k}}{\|f_{k+1}^{\mathrm{N}}\|_{H_k}}$.
\end{proposition}

\section{AICN with Weak Learners}\label{sec:AICNWithWeakLearners}

We are now ready to define the weak-AICN scheme. Rather than simply minimizing the local quadratic Taylor bound $Q_k$, global methods like Cubic Regularized Newton (CRN) \cite{2006CubicRegularizationOfNewtonMethodAndItsGlobalPerformance,2008AcceleratingTheCubicRegularizationOfNewtonsMethodOnConvexProblems} and AICN \cite{2022ADampedNewtonMethodAchievesGlobalMinus2AndLocalQuadraticConvergenceRate} instead add an additional cubic term to the minimization step, resulting in a \emph{global bound} to be minimized. CRN measured this extra cubic term in the Euclidean norm, resulting in an optimization procedure which required solving an inner expensive subproblem at each iteration. AICN instead added a cubic term using the Hessian-induced norm, which resulted in the AICN update reducing to a damped Newton step. Below, we define our weak-AICN update:

\begin{definition}[Weak-AICN]\label{def:weakAICN}
The weak-AICN iterates $F_k$ with smoothness constant $C_{\mathrm{est}}$ and fixed learning rate $\eta>0$ are defined recursively via the cubic model
\begin{align*}
    F_{k+1}\ :=\ F_k + \eta f_{k+1}^{\mathrm{wAICN}},
    \qquad \qquad
    f_{k+1}^{\mathrm{wAICN}} = \argmin_{f\in \mathcal{F}} M_k(f),
\end{align*}
where $F_0\in\mathcal{H}$ is a given initial value and 
\begin{align*}
    M_k(f)
    \;  := \; 
    \langle g_k,f\rangle + \frac12\|f\|_{H_k}^2 + \frac{C_{\mathrm{est}}}{6}\|f\|_{H_k}^3 
    \;  = \;  
    Q_k(f) + \frac{C_{\mathrm{est}}}{6}\|f\|_{H_k}^3.
\end{align*}
\end{definition}

The fundamental objects in our analysis are the quadratic model $Q_k$, the cubic model $M_k$, the directional improvement $\delta_k$ and the weak learner cosine angle $\Theta_k$. Using only \cref{def:weakAICN} and \cref{prop:deltak_Qk}, we are able to prove the following theorem, which says that weak-AICN reduces to a damped weak-Newton step, with an explicit stepsize governed by the quantity $C_{\mathrm{est}}\delta_k(f^{\mathrm{wN}}_{k+1})$. The resulting algorithm is summarized in \cref{alg:wAICN}. All proofs are deferred to \cref{app:WeakAICNProofs}.

\begin{theorem}[w-AICN is damped w-Newton]\label{theorem:wAICN_is_wNewton}
The weak-AICN iterates $f_{k+1}^{\mathrm{wAICN}}$ are of the form
\[
f_{k+1}^{\mathrm{wAICN}}
=
\alpha_k\, f_{k+1}^{\mathrm{wN}},
\]
with explicit stepsize determined by the quantity $C_{\mathrm{est}} \delta_k$, where $\delta_k := \delta_k(\mathcal{F}) = \|f_{k+1}^{\mathrm{wN}}\|_{H_k}$:
\[
\alpha_k
\ :=\
\frac{-1+\sqrt{1+2C_{\mathrm{est}}\delta_k}}{C_{\mathrm{est}}\delta_k}
\ =\
\frac{2}{1+\sqrt{1+2C_{\mathrm{est}}\delta_k}}.
\label{eq:F-aicn-alpha}
\]
Moreover, we have $0<\alpha_k \leq 1$, the identity  $\frac{1}{2}C_{\mathrm{est}}\delta_k\alpha_k^2 = 1-\alpha_k$, and that $\alpha_k\uparrow 1$ as $\delta_k\downarrow 0$.
\end{theorem}

\begin{algorithm}[]
\caption{Weak-AICN}
\label{alg:wAICN}
\begin{algorithmic}[1]
    \STATE \textbf{Input:} Loss $L:\mathcal{H}\to\mathbb{R}$. Learning rate $\eta$. Weak learner family $\mathcal{F}$. Cubic regularization $C_{\mathrm{est}}$. Initial value $F_0$.
    \FOR{$k = 0, 1, 2, \dots$}
        \STATE $g_k, \: H_k \leftarrow \nabla L(F_k), \: \nabla^2L(F_k)$
        \STATE $f^{\mathrm{wN}}_{k+1} \leftarrow \argmin_{f \in \mathcal{F}} Q_k(f)$
        \STATE $\delta_k, \; \alpha_k \leftarrow \sqrt{|\langle g_k, f^{\mathrm{wN}}_{k+1}\rangle|}, \; \frac{2}{1+\sqrt{1+2C_{\mathrm{est}}\delta_k}}$
        \STATE $F_{k+1} \leftarrow F_k + \eta \alpha_k f^{\mathrm{wN}}_{k+1}$
    \ENDFOR
\end{algorithmic}
\end{algorithm}

Unlike exact AICN \cite[Theorem 1]{2022ADampedNewtonMethodAchievesGlobalMinus2AndLocalQuadraticConvergenceRate}, which defines its dampening using the dual gradient norm $\|g_k\|_{H_k}^*$, our weak learner formulation reveals that the directional improvement $\delta_k$ is the true driving mechanism of the dampening. The exact case can be recovered via the identity $\Theta_k \|g_k\|_{H_k}^* = \delta_k$ from \cref{prop:deltak_Qk}. Crucially, \cref{theorem:wAICN_is_wNewton} is purely algebraic and requires no extra regularity assumptions on the loss, allowing us to derive a bound on the cubic model, presented in \cref{prop:weak_cubic_improvement_theta_factor}.

\begin{proposition}[Weak Improvement $\Theta^2_k$ Factor ]\label{prop:weak_cubic_improvement_theta_factor}
Let $\alpha_k$ be the weak-AICN dampening, and write $M_k(\mathcal{F}) := \min_{f\in\mathcal{F}} M_k(f)$. We have the following weak learner bound:
\begin{equation}
    M_k(\mathcal{F}) 
    \ = \
    \frac{\alpha_k(\alpha_k-4)}{6} \| f^{\mathrm{wN}}_{k+1}\|_{H_k}^2    
    \ \leq\
    \Theta_k^2\, M_k(\mathcal{H}).
\end{equation}
\end{proposition}

\subsection{Semi-strong Self-concordance and Weak Contracting-point Inequality}

To establish our global convergence guarantees, we must impose regularity assumptions on the loss function. In the second-order optimization literature, global methods such as Cubic Regularized Newton \cite{2006CubicRegularizationOfNewtonMethodAndItsGlobalPerformance,2008AcceleratingTheCubicRegularizationOfNewtonsMethodOnConvexProblems} and Gradient-Regularized Newton (GRN) \cite{2023RegularizedNewtonMethodWithGlobalConvergence,2024GradientRegularizationOfNewtonMethodWithBregmanDistances,2026GradientRegularizedNewtonBoostingTreeswithGlobalConvergence} typically assume that the Hessian operator is Lipschitz continuous. In this paper, we instead adopt the regularity assumption of the classical AICN method \cite{2022ADampedNewtonMethodAchievesGlobalMinus2AndLocalQuadraticConvergenceRate}, assuming the loss $L$ satisfies \textit{semi-strong self-concordance}. This allows us to bound the variation of the Hessian strictly within the local geometry induced by the Hessian itself, and has the advantage that our bounds are affine-invariant with no dependence on choice of basis. We formally state this definition below in the language of Hessian norms.

\begin{definition}[Semi-strong self-concordance \cite{2022ADampedNewtonMethodAchievesGlobalMinus2AndLocalQuadraticConvergenceRate}]\label{def:semi-strong-self-concordance}
For a bounded linear operator 
$A : \mathcal H \to \mathcal H$, define the \textbf{Hessian-induced affine-invariant operator norm} as
\[
\|A\|_{\operatorname{aff-op},H_k}
\ :=\ \ 
\sup_{u\neq 0}
\frac{\|Au\|_{H_k^{-1}}}{\|u\|_{H_k}}. \label{eq:affine-invariant-operator-norm}
\]
Equivalently,
\(
    \|A\|_{\operatorname{aff-op},H_k} 
    = \bigl\| H_k^{-1/2}\, A \, H_k^{-1/2} \bigr\|_{\operatorname{op}},
\)
where $\|\cdot\|_{\operatorname{op}}$ is the standard operator norm.
The convex loss $L$ is said to be \textbf{semi-strongly self-concordant (SSC)} with constant $C_{\mathrm{ssc}}>0$ if for all $F\in \mathcal{H}$ and all $f\in \mathcal{H}$,
writing $H_F:=\nabla^2 L(F)$,
\[
\|\nabla^2 L(F+f)-\nabla^2 L(F)\|_{\operatorname{aff-op},H_F}
\ \le\
C_{\mathrm{ssc}}\|f\|_{H_F}.
\label{eq:semi-strong-self-concordance}
\]
\end{definition}


Assuming SSC yields the following bound on the second-order Taylor error, which most subsequent analysis will depend on. A direct consequence 
is that the cubic model $M_k(f)$ being minimized by AICN is a global bound, rather than a local bound like the quadratic approximation $Q_k(f)$.

\begin{proposition}[Affine-invariant cubic Taylor bound \cite{2022ADampedNewtonMethodAchievesGlobalMinus2AndLocalQuadraticConvergenceRate}]\label{cubic-taylor-bound}
Assume $L$ is SSC. Fix $F\in \mathcal{H}$ and write $H_F:=\nabla^2 L(F)$. Then for all $f\in \mathcal{H}$,
\begin{equation}
\bigl|L(F+f)-L(F)-Q_F(f)\bigr|
\ \le\
\frac{C_{\mathrm{ssc}}}{6}\|f\|_{H_F}^3.
\label{eq:F-t3}
\end{equation}
\end{proposition}

By evaluating this global upper bound at the weak-AICN update, we are able to prove that the weak-AICN update successfully contracts the objective function toward any arbitrary comparison point $Y \in \mathcal{H}$. Crucially, the strength of this contraction is scaled by the squared weak learner cosine angle $\Theta_k^2$ and learning rate $\eta$.

\begin{proposition}[\textbf{Weak Contracting-point Inequality}]\label{weak-aicn-contracting-point}
Let the loss $L$ be SSC and consider weak-AICN iterates with $C_{\mathrm{est}} \geq C_{\mathrm{ssc}}$ and learning rate $0<\eta \leq 1$. Then for every $Y\in \mathcal{H}$,
\begin{equation}
    L(F_{k+1})
    \ \leq \ 
    (1-\eta\Theta_k^2) L(F_k) + \eta\Theta_k^2
    \left( L(Y) + \frac{C_{\mathrm{est}}}{3}\|Y-F_k\|_{H_k}^3 \right).
    \label{eq:w-cp}
\end{equation}
\end{proposition}

\subsection{Global and Local Convergence of AICN with Weak Learners}

Before we state our convergence theorems, we first present two recurrence lemmas used in our analysis. These lemmas allow us to improve the weak learner bounds from a \emph{worst-case} analysis to an \emph{average performance} analysis. Prior works define their weak learner edges either as an infimum over all iterations $k$ \cite{2011GeneralizedBoostingAlgorithmsForConvexOptimization,2014AConvergenceRateAnalysisForLogitBoostMARTAndTheirVariant,2026GradientRegularizedNewtonBoostingTreeswithGlobalConvergence}, or as an infimum over all possible weak learners
\cite{2020RandomizedGradientBoostingMachine,2020AcceleratingGradientBoostingMachines,2020SnapBoostAHeterogeneousBoostingMachine,2012APrimalDualConvergenceAnalysisOfBoosting}. This is problematic, since while weak learner conditions are generally stable on a rolling-window basis, they can be noisy from iteration to iteration, with worst-case values often close to 0. By using these recurrences, our final convergence rates depend on the path-averaged cosine angle instead of the infimum, proving that a single poorly-fitted weak learner does not degrade the theoretical guarantee of the whole optimization process.

\vspace{-7pt}\begin{lemma}\label{lemmaAlpha3div2Recursion} 
    Let $\Delta_k$ be a non-negative sequence such that $\Delta_{k+1} \leq \Delta_k - c_k\Delta_k^{\frac{3}{2}}$ for a sequence of constants $c_k \geq 0$. Then for all $k \geq 1$ with $\sum_{i=0}^{k-1}c_i>0$
    \begin{equation}
        \Delta_{k} \leq \frac{4}{\left( 2c_0 + \sum_{i=0}^{k-1} c_i \right)^2} \leq \frac{4}{\overline{c_{k}}^2 k^2},
    \end{equation}
    where $\overline{c_k}$ is the average $\overline{c_k} := \frac{1}{k} \sum_{i=0}^{k-1} c_i$.
\end{lemma}
\begin{lemma}\label{lemmaLinearRecursion}
Let $\{\Delta_k\}_{k\geq 0}$ be a non-negative sequence such that $\Delta_{k+1} \leq (1-c_k)\,\Delta_k$ for a sequence of constants $c_k \in [0,1]$. Then for all $k \geq 1$,
\begin{equation}
    \Delta_k 
    \;\leq\;
     \Delta_0\prod_{i=0}^{k-1} (1-c_i)
    \;\leq\;
    \Delta_0 \exp\!\left(-\sum_{i=0}^{k-1} c_i\right)
    \;\leq\;
    \Delta_0 e^{-k\overline{c_k}},
\end{equation}
where $\overline{c_k} := \frac{1}{k}\sum_{i=0}^{k-1} c_i$.
\end{lemma}

We impose the following boundedness assumption which is standard for affine-invariant and self-concordance analysis, including for proving the global $\mathcal{O}(1/k^2)$ rate for the exact AICN scheme \cite{2022ADampedNewtonMethodAchievesGlobalMinus2AndLocalQuadraticConvergenceRate}. Note that this assumption automatically is satisfied if $L$ is strongly convex and smooth, or strongly convex with Lipschitz continuous Hessian (see \cref{remark:sufficient_condition_for_bounded_local_radius}).

\begin{assumption}[Bounded local radius]\label{F-aicn-bounded-radius}
Let $F^* = \argmin_{F\in \mathcal{H}} L(F)$ and assume the weak-AICN iterates satisfy a uniform bound
\[
\sup_{k\geq 0}\ \|F_k-F^*\|_{H_k}\ \le\ D < \infty.
\label{eq:F-aicn-bd}
\]
\end{assumption}

Applying our recurrence lemmas to the weak contracting-point inequality (\cref{weak-aicn-contracting-point}), we are able to establish the following global convergence of weak-AICN.

\begin{theorem}[Global convergence]\label{thm:aicn_global_rate}
Let $L$ be SSC and consider weak-AICN iterates $F_k$ with $C_{\mathrm{est}} \geq C_{\mathrm{ssc}}$ and learning rate $\eta \in (0, 1]$. Define $\Delta_k := L(F_k)-L(F^*)$ and let $k_0$ be the first iteration such that $\Delta_{k_0} \leq C_{\mathrm{est}} D^3$. Define the path-averaged cosine edges $\overline{\Theta_k^2} := \frac{1}{k}\sum_{i=0}^{k-1} \Theta_i^2$ and $\overline{\Theta_{k_0, k}^2} := \frac{1}{k-k_0}\sum_{i=k_0}^{k-1} \Theta_i^2$. Weak-AICN exhibits an initial global linear phase, followed by a global $\mathcal{O}(1/k^2)$ phase. If $\Delta_0 \leq C_{\mathrm{est}}D^3$, then $k_0=0$. If $k_0>1$, the length of the initial phase is bounded by
\(
k_0 < \frac{3}{2\eta\overline{\Theta_{k_0-1}^2}}\ln\left(\frac{\Delta_0}{C_{\mathrm{est}}D^3}\right) + 1
\). The convergence rates are:

\vspace{-2pt}
\begin{enumerate}
    \item Initial Linear Regime ($1 \leq k < k_0$): \hspace{25pt}
    \(
     \Delta_k \leq \Delta_0 \exp\left(-\frac{2}{3}\eta k \overline{\Theta_k^2}\right).
    \)
    \item Global $\mathcal{O}(1/k^2)$ Regime ($k > k_0$):\hspace{30pt}
    \(
     \Delta_k \leq \frac{9\,C_{\mathrm{est}}\,D^3}{\eta^2 (\overline{\Theta_{k_0, k}^2})^2 (k-k_0)^2}.
    \)
\end{enumerate}
\end{theorem}

Next, we establish the local convergence of weak-AICN. While the global regime is governed by the minimization of the cubic upper bound, the local regime is instead characterized by the contraction of the gradient in the dual Hessian norm. Due to the extra error introduced by the weak learners, we are only able to prove a local linear rate---similar to prior boosting works \cite{2011GeneralizedBoostingAlgorithmsForConvexOptimization,2020RandomizedGradientBoostingMachine,2014AConvergenceRateAnalysisForLogitBoostMARTAndTheirVariant,2020SnapBoostAHeterogeneousBoostingMachine,2026GradientRegularizedNewtonBoostingTreeswithGlobalConvergence}---rather than the quadratic rates found in the classical finite-dimensional convex optimization literature \cite{2022ADampedNewtonMethodAchievesGlobalMinus2AndLocalQuadraticConvergenceRate}. Our local rate's dependence on the squared edge $\Theta_k^2$ is consistent with prior work referenced above.

\begin{theorem}[Local Linear Rate]
    \label{thm:local_linear}
    Let the loss $L$ be SSC with learning rate $\eta \in (0, 1]$ and $C_{\mathrm{est}} \geq C_{\mathrm{ssc}}$. If the weak-AICN iterates satisfy the local condition $ \|g_t\|_{H_t}^* \leq \frac{\Theta_t}{6C_{\mathrm{est}}}$ for $t = 0, \dots, k-1$, then the gradient contracts in the dual Hessian norm:
    \begin{equation}
        \|g_{t+1}\|^*_{H_{t+1}} \leq (1 - \rho_t) \|g_t\|^*_{H_t},
    \end{equation}
    where $\rho_t = \frac{1}{4}\eta \alpha_t \Theta_t^2$. Consequently, we obtain a local linear rate of convergence via the bound
    \begin{equation}
        L(F_k) - L(F^*) 
        \; \leq \;
        \|g_k\|_{H_k}^* D
        \; \leq \;
        \|g_0\|_{H_0}^* D \exp\left(-k \overline{\rho_k}\right),
    \end{equation}
    where $\overline{\rho_k} := \frac{1}{k}\sum_{t=0}^{k-1} \rho_t = \frac{\eta}{4k}\sum_{t=0}^{k-1} \alpha_t \Theta_t^2$ is the path-averaged contraction rate.
\end{theorem}

\subsection{Global Convergence of weak-AICN for Hessian-dominated Losses}

Weak-AICN can obtain global linear rates for the class of Hessian-dominated losses, a property recently introduced in the Restricted Newton Descent framework \cite{2026GradientRegularizedNewtonBoostingTreeswithGlobalConvergence} motivated by the LogitBoost analysis \cite{2014AConvergenceRateAnalysisForLogitBoostMARTAndTheirVariant}. The canonical example of such a loss is binary and categorical cross entropy with $c=1$. Note that Hessian dominance alone does not imply SSC; in particular,
unregularized binary log-loss is not even self-concordant \cite{2010SelfConcordantAnalysisForLogisticRegression}. However, $\ell_2$-regularized cross entropy satisfies both Hessian dominance and SSC, since it is strongly convex and smooth with a Lipschitz continuous Hessian.

\begin{definition}[Hessian-dominance \cite{2026GradientRegularizedNewtonBoostingTreeswithGlobalConvergence}]\label{defHessianDominatedLoss}
    Assume $L_\star:=\inf_{F\in\mathcal H}L(F)>-\infty$. We say that the loss sequence $L(F_k)$ is \textbf{Hessian-dominated} if there exists a constant $c>0$ such that $\|f^\mathrm{N}_{k+1}\|^2_{H_k} \geq c\bigl(L(F_k)-L_\star\bigr)$ for all $k$.
\end{definition}

\begin{theorem}[Linear weak-AICN rate for Hessian-dominated loss]\label{thm:hessDominatedLinear}
Let $L$ be SSC and consider weak-AICN iterates with $C_{\mathrm{est}} \geq C_{\mathrm{ssc}}$ and learning rate $\eta \in (0, 1]$. If $L$ further satisfies \cref{defHessianDominatedLoss}, then the weak-AICN iterates satisfy
\begin{equation}
    L(F_{k+1})-L_\star \leq \left( 1 - \rho_k \right)\bigl(L(F_k)-L_\star\bigr),
\end{equation}
where $\rho_k = c \eta \Theta^2_k\left(1 - \frac{\eta}{2}\right) \alpha_k$.
\end{theorem}

This result is notable because the analogous linear rate for vanilla Newton boosting typically requires a restricted learning rate $\eta \in (0, \frac{2\mu}{S})$ under strict $\mu$-strong convexity and $S$-smoothness assumptions \cite{2026GradientRegularizedNewtonBoostingTreeswithGlobalConvergence}. In contrast, \cref{thm:hessDominatedLinear} holds for all learning rates $\eta\leq 1$, trading strong convexity for a sufficiently large self-concordance constant $C_{\mathrm{est}} \geq C_{\mathrm{ssc}}$.

\section{Applications of Weak-AICN}
In this section, we present two application areas of weak-AICN and optimization with weak learners.
\subsection{Newton boosting}
\label{sec:gradient-boosting}
The primary application of weak learner analyses is gradient tree boosting. Given a training dataset $X = \{x_1, \dots, x_n\} \subset \mathbb{R}^p$ and targets $y_i$, we work in the Hilbert space $\mathcal{H}_X = \mathcal{L}^2(\widehat{\nu}_n)$ of square-integrable functions equipped with the empirical measure $\widehat{\nu}_n = \frac{1}{n}\sum_{i=1}^n \delta_{x_i}$. The boosting objective is to minimize the empirical risk $L(F) = \frac{1}{n}\sum_{i=1}^n l(F(x_i), y_i)$, where $l$ is the data-point-wise loss. This space can be identified with $\mathbb{R}^n$, equipped with the inner product $\langle f, h\rangle_X = \frac{1}{n}\sum_{i=1}^n f(x_i)h(x_i)$ and induced norm $\|f\|_X^2 = \frac{1}{n}\sum_{i=1}^n f(x_i)^2$. When the weak learner family $\mathcal{F}$ is restricted to the set of depth-$J$ decision trees $\mathcal{T}_J$, the classical Newton boosting updates used by libraries like XGBoost are recovered \cite{2026GradientRegularizedNewtonBoostingTreeswithGlobalConvergence}. Since weak-AICN is simply damped weak-Newton, implementing our scheme into existing methodologies is very simple, requiring only computing the stepsize $\alpha_k$ in \cref{theorem:wAICN_is_wNewton}. Next, we show that the regularity assumptions required for weak-AICN carry over naturally to $L$ from the base loss $l$. See \cref{app:SSC} for proof.

\begin{theorem}[SSC inheritance for empirical risks]\label{thm:ssc_inheritance}
    Assume that for each target $y_i$, the scalar function $l(\cdot, y_i)$ is semi-strongly self-concordant with constant $C_l$. Then the empirical risk $L(F) = \frac{1}{n}\sum_{i=1}^n l(F(x_i), y_i)$ is semi-strongly self-concordant on $\mathcal{L}^2(\widehat{\nu}_n)$ with constant \( C_{\mathrm{ssc}}(L) \le \sqrt{n}\,C_l \).
\end{theorem}

To achieve convergence via \cref{thm:aicn_global_rate}, the weak learners must maintain a non-zero cosine angle $\Theta_k$ along the optimization path. Unlike the weak gradient edge $\gamma$ used in weak-GRN's \cite{2026GradientRegularizedNewtonBoostingTreeswithGlobalConvergence} bounds, which is difficult to verify theoretically, the strict positivity of $\Theta_k$ follows directly from the identity $\Theta_k \|g_k\|_{H_k}^* = \delta_k(\mathcal{F})$ established in \cref{prop:deltak_Qk}. This relationship guarantees that any fitted decision tree achieving a strictly positive improvement in loss automatically has $\Theta_k > 0$. The real difficulty, however, lies in quantitative control. Establishing universal, deterministic lower bounds for $\Theta_k$ over all possible datasets is impossible in general. For instance, consider the parity function $y_i = (-1)^{\sum_j (x_i)_j}$ on the Boolean hypercube $X = \{0,1\}^p$; any decision tree of depth $J < p$ leaves at least one coordinate free per leaf, outputting exactly zero on average and yielding a worst-case cosine angle of exactly $\Theta_k = 0$. Because deterministic worst-case bounds are fatally pessimistic for general tree weak learners, analyzing tree alignment requires shifting from a worst-case perspective to a probabilistic one.

Since the exact Hessian-induced cosine angle $\Theta_k$ is presently too hard to analyze directly, we instead study a random proxy. Concretely, assume
$X_1,\dots,X_n \stackrel{\mathrm{iid}}{\sim}\mathrm{Unif}([0,1]^p)$ and let $\mathsf Q$ be any metric model which, conditional on $X$, samples a positive definite self-adjoint operator $A:\mathcal H_X\to\mathcal H_X$. Write
\[
\langle f,h\rangle_A:=\langle f,Ah\rangle_X,
\qquad
S_A(\mathcal H_X):=\{f\in\mathcal H_X:\|f\|_A=1\},
\]
and draw \(U_A\mid(X,A)\sim\mathrm{Unif}(S_A(\mathcal H_X))\). We define the random depth-$J$ metric proxy by
\[
\widehat{\Theta}_{n,p,J}^{\mathrm{prox},\mathsf Q}
:=
\max_{g\in\mathcal T_J(X)\setminus\{0\}}
\frac{\langle U_A,g\rangle_A}{\|g\|_A}.
\]
The identity-metric proxy is recovered by taking \(A=I\) deterministically. The point of this construction is to give a tractable no-signal baseline for random tree alignment in a realized Hessian geometry. The following theorem gives a lower bound for this quantity, uniformly over the metric model \(\mathsf Q\), which is dyadic with respect to tree depth $J$. See \cref{app:tree-proxy-experiments} for the geometric setup and proof.  

\begin{theorem}[Dyadic tree-angle lower bounds]\label{thm:tree-average-main}
Let $n\ge2$, $p\ge1$, and $J\ge0$, and let $\mathsf Q$ be any metric model as above. If $2^J<n$, then there exists a random variable
\[
Z_J\sim \mathrm{Beta}\!\left(2^{J-1},\frac{n-2^J}{2}\right)
\]
such that $(\widehat{\Theta}_{n,p,J}^{\mathrm{prox},\mathsf Q})^2 \ge Z_J$ almost surely. In particular,
\(
\mathbb E\!\left[(\widehat{\Theta}_{n,p,J}^{\mathrm{prox},\mathsf Q})^2\right]
\ge \frac{\min(2^J,n)}{n}
\)
and
\(
\mathbb E[\widehat\Theta^{\mathrm{prox},\mathsf Q}_{n,p,J}]
\ge
\frac{
\Gamma\!\left(\frac{2^J+1}{2}\right)
\Gamma\!\left(\frac n2\right)
}{
\Gamma\!\left(2^{J-1}\right)
\Gamma\!\left(\frac{n+1}{2}\right)
}.
\)
If $2^J\ge n$, then $\widehat{\Theta}_{n,p,J}^{\mathrm{prox},\mathsf Q}=1$ almost surely.
\end{theorem}

\begin{remark}
    \Cref{thm:alternative_tree_angle} in the Appendix gives an alternative, potentially simpler and more intuitive, signal-plus-noise formulation of \cref{thm:tree-average-main}. There, a tree-aligned signal plus isotropic Gaussian noise in the Hessian geometry yields the same squared-angle expectation lower bound.
\end{remark}

\subsection{Newton Coordinate Descent}\label{sec:NewtonCoordinateDescent}

To the best of our knowledge, prior theoretical frameworks for optimization with weak learners have exclusively been used for gradient boosting. However, the Restricted Newton Descent framework is fundamentally space-agnostic. By simply changing the choice of the base space $\mathcal{H}$ and the weak learner family $\mathcal{F}$, we can recover classical algorithms from finite-dimensional convex optimization.

Coordinate descent is a classical optimization paradigm \cite{2015CoordinateDescentAlgorithmsWright}, with modern analyses spanning randomized updates \cite{2012EfficiencyOfCoordinateDescentMethodsOnHugeScaleOptimizationProblemsNesterov, 2014IterationComplexityOfRandomizedBlockCoordinateDescentMethodsForMinimizingACompositeFunction} and second-order subspace methods \cite{2019RSNRandomizedSubspaceNewton,2018RandomizedBlockCubicNewtonMethod,2022LetsMakeBlockCoordinateDescentConvergeFaster}. In our framework, let the base space be $\mathcal{H} = \mathbb{R}^d$ equipped with the standard basis $\{e_1, \dots, e_d\}$. If we let the weak learner family be the set of all scaled basis vectors, $\mathcal{F} = \{ce_i : c\in\mathbb{R}, 1\leq i \leq d\}$, we recover Newton Coordinate Descent. To compute the corresponding weak update, one has to maximize the directional improvement $\delta_k(\varphi)$ over the basis vectors. For a single coordinate direction $\pm e_i$, this evaluates to
\begin{align}
    \delta_k(\pm e_i)
    \ = \
    \frac{\mp\langle g_k, e_i\rangle}{\|e_i\|_{H_k}} 
    \ = \ 
    \frac{\mp(g_k)_i}{\sqrt{(H_k)_{ii}}}.
\end{align}
Therefore, finding the optimal weak direction reduces to selecting the coordinate $i^*$ that maximizes $i^* = \argmax_{i} |(g_k)_i| / \sqrt{(H_k)_{ii}}$. This is a one-dimensional Newton analogue of the Gauss--Southwell--Lipschitz selection rule \cite{2015CoordinateDescentConvergesFasterWithTheGaussSouthwellRuleThanRandomSelection}, with $(H_k)_{ii}$ acting as the local curvature. By \cref{prop:deltak_Qk}, the resulting weak Newton step is exactly a 1D scalar Newton step along the chosen coordinate:
\begin{align}
    f_{k+1}^{\mathrm{wN}}
    \ = \
    \frac{|\delta_k(e_{i^*})|}{\|e_{i^*}\|_{H_k}} \operatorname{sign}(-(g_k)_{i^*})\, e_{i^*}
    \ = \ 
    -\frac{(g_k)_{i^*}}{(H_k)_{i^*i^*}} e_{i^*}.
\end{align}
Calculating this update requires only the gradient and the diagonal of the Hessian, avoiding the $\mathcal{O}(d^3)$ cost of full matrix inversion. Furthermore, our general weak learner angle machinery immediately yields the following lower bound for the coordinate descent cosine angle. See \cref{app:CoordinateDescent} for proof.

\begin{proposition}\label{prop:coordinate_descent_lower_bound}
    For Newton Coordinate Descent ($\mathcal{F} = \{ce_i : c\in\mathbb{R}, 1\leq i \leq d\}$), the weak learner cosine angle is deterministically lower bounded by
    \begin{equation}
        \Theta_k \geq \frac{1}{\sqrt{d \kappa_k}},
    \end{equation}
    where $d$ is the dimension and $\kappa_k = \lambda_{\max}(H_k)/\lambda_{\min}(H_k)$ is the condition number of the Hessian.
\end{proposition}

Consequently, applying weak-AICN with coordinate weak learners immediately yields a globally convergent, damped Newton Coordinate Descent scheme. This algorithm shares some connections with a recent low-rank and sketched subspace AICN method \cite{2025SketchAndProjectMeetsNewtonMethodGlobalConvergenceWithLowRankUpdates}. However, our method arises here organically as a special case of optimization with weak learners, inheriting all theoretical results from the weak-AICN framework. In the finite-dimensional convex optimization literature, it is common to use adaptive versions of algorithms where regularity constants such as $C_{\mathrm{est}}$ are adjusted dynamically during optimization \cite{2023RegularizedNewtonMethodWithGlobalConvergence}. We present one such modification to weak-AICN in \cref{app:AdaptiveWeakAICN}.

\subsection{Supplementary Experiments}

\begin{figure}[htbp]
    \centering
    \begin{subfigure}[b]{0.45\textwidth}
        \centering
        \includegraphics[width=\textwidth]{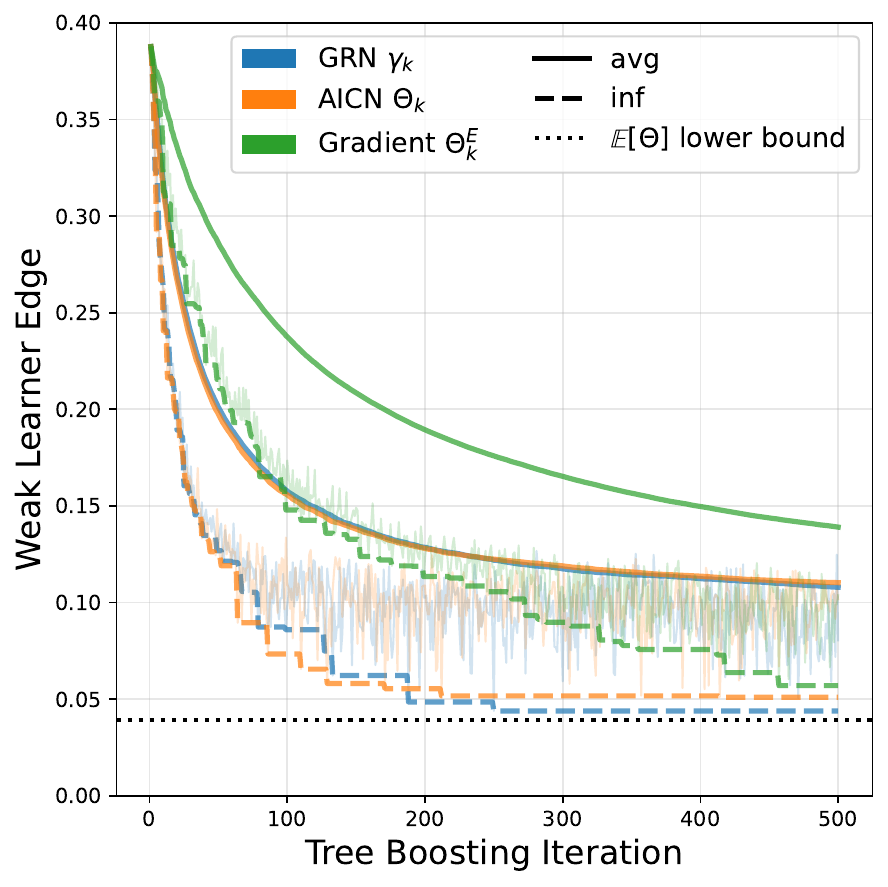}
    \end{subfigure}
    \hfill
    \begin{subfigure}[b]{0.45\textwidth}
        \centering
        \includegraphics[width=\textwidth]{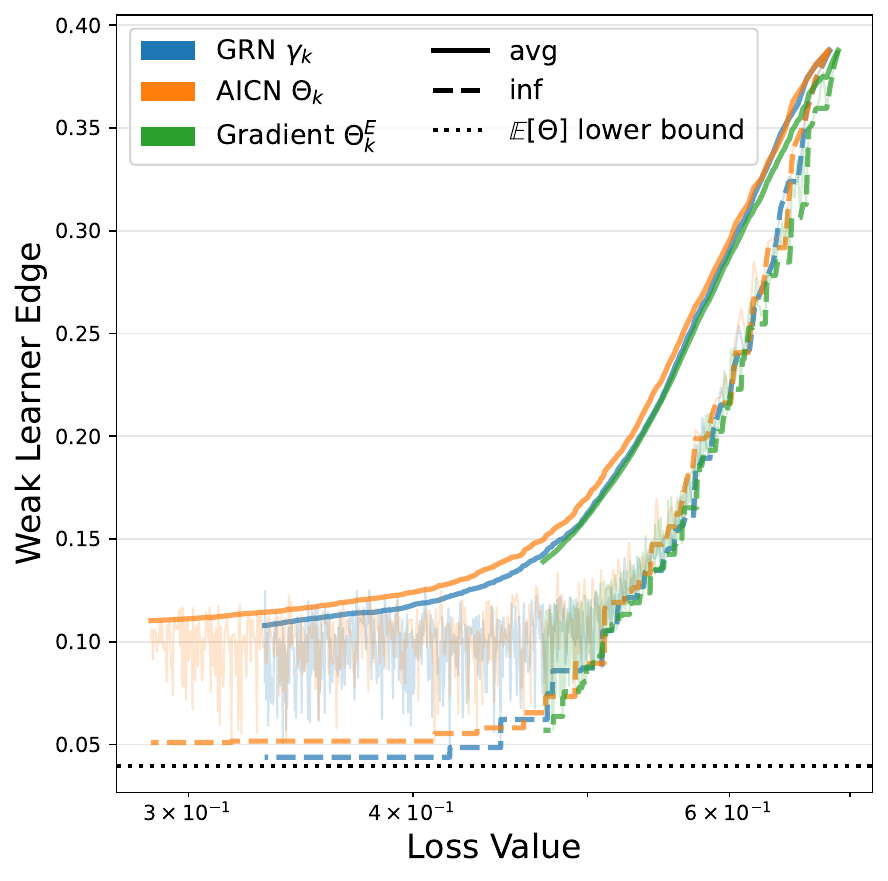}
    \end{subfigure}
    \caption{Comparison of weak learner edges for tree boosting on HIGGS ($n=10000,\ J=4$): Hessian cosine angle $\Theta_k$ for weak-AICN, weak gradient edge $\gamma_k$ for weak-GRN, and Euclidean cosine angle $\Theta_k^E$ for first-order boosting. \textbf{Left:} Edges against boosting iterations. \textbf{Right:} Edges against training loss. Solid lines show arithmetic running averages; dashed lines show running infima. The dotted line shows the proxy expectation lower bound from \cref{thm:tree-average-main} as a reference value.}
    \label{fig:higgs_weak_learner_edge_comparisons}
\end{figure}

We provide three sets of experiments, with full experimental details in the appendix. The goal of these experiments is not to claim state-of-the-art predictive performance, but rather to empirically validate and supplement theoretical findings presented in this paper. All code was run on a single CPU, and is available at \url{github.com/nikitazozoulenko/global-newton-boosting}. Our experiments in this paper include:

\begin{enumerate}
    \item \textbf{Adaptive weak-AICN (\cref{app:AdaptiveWeakAICN}):} We empirically validate our adaptive modification of the (weak-)AICN algorithm. Enabled by the simplification of the cubic model bound in \cref{prop:weak_cubic_improvement_theta_factor}, this adaptive scheme dynamically tunes the regularity constant without requiring expensive inner-loop subproblems or re-evaluations of the Hessian or gradient.
    
    \item \textbf{Tree Boosting Convergence and Weak Learner Edges (\cref{app:ExperimentsGRNComparison}):} Following the experimental setup of \cite{2026GradientRegularizedNewtonBoostingTreeswithGlobalConvergence}, we compare weak-AICN with weak-GRN, vanilla Newton boosting, and first-order boosting, using decision trees as weak learners. We demonstrate the non-convergence of vanilla Newton boosting under the Charbonnier loss, compare weak learner edges on the HIGGS dataset, and analyze their relationship with tree depth. \Cref{fig:higgs_weak_learner_edge_comparisons} compares the Hessian cosine angle $\Theta_k$ used by weak-AICN with the weak gradient edge $\gamma_k$ used by weak-GRN. These quantities are comparable as cosine-like weak learner edges defined in different geometries, and take similar empirical values. This supports the comparison of our global $\mathcal{O}(1/(\overline{\Theta}_k^4 k^2))$ rate with weak-GRN's $\mathcal{O}(1/(\gamma^{12}k^2))$ rate, showing substantially better dependence on weak learner alignment. Furthermore, our bound uses a path-averaged cosine angle, whereas weak-GRN uses the worst-case edge $\gamma=\inf_t\gamma_t$. The full experimental setup and metric definitions are provided in \cref{app:ExperimentsGRNComparison}.
    
    \item \textbf{Random Tree Angle Proxy Experiments (\cref{app:tree-angle-experiments}):} We evaluate the dyadic tree-angle lower bounds and structured-signal hypotheses in relation to \cref{thm:tree-average-main}.
\end{enumerate}

\section*{Limitations}

While weak-AICN is simple to implement as a damped weak-Newton step, and improves the currently best known weak learner edge dependence in the global $\mathcal{O}(1/k^2)$ rate from $\gamma^{12}$ to $\overline{\Theta}_k^4$, this geometric shift comes with a trade-off. Because weak-AICN operates entirely within the local Hessian geometry, it can struggle with ill-conditioned problems if static $\ell_2$ regularization is not added, unlike weak-GRN which handles such problems seamlessly via adaptive gradient-based $\ell_2$ penalties. Furthermore, in practical gradient boosting applications with small learning rates, there is often little empirical difference between vanilla Newton boosting and globally convergent variants. Nevertheless, global schemes remain strictly necessary to theoretically guarantee convergence. In this context, our weak-AICN analysis helps explain why libraries like XGBoost perform well in practice, as our theory guarantees that the standard Newton boosting algorithm will safely converge provided the learning rate $\eta$ does not exceed the theoretically derived weak-AICN dampening factor $\alpha_k$.

\section*{Conclusion}

In this paper, we introduced weak-AICN, a globally convergent second-order optimization scheme with weak learners.
By analysing directional improvements $\delta_k$ in the local Hessian norm, we proved that the weak-AICN update reduces to a damped weak-Newton step with an explicit, easily computable stepsize, without requiring any regularity assumptions on the loss function. Building upon this update and measuring regularity through semi-strong self-concordance, we established an affine-invariant global $\mathcal{O}(1/\overline{\Theta}_k^4 k^2)$ convergence rate. This significantly improves the weak learner edge dependence compared to prior best-known bounds for the weak-GRN scheme, and transitions theoretical guarantees from a pessimistic worst-case infimum to an average-performance bound. We further established a local linear rate for general convex losses, and a global linear rate for Hessian-dominated losses. To address the impossibility of deterministic worst-case bounds for decision trees, we derived explicit Beta-distributed stochastic lower bounds on tree alignment in a random configuration model, including an expectation bound that improves dyadically with tree depth. Finally, by showing that Newton coordinate descent emerges naturally when restricting the weak learner family to standard basis vectors, we demonstrated that this framework extends beyond gradient boosting, connecting our work and the field to the broader convex optimization literature.

\section*{Acknowledgements}
NZ has been supported by the Roth Scholarship at Imperial College London. DF was supported by the Göran Gustafsson Foundation for Research in Natural Sciences and Medicine via Alexander Berglund's 2024 Göran Gustafsson Prize. The work of TC was supported in part by UK Research and Innovation (UKRI) through the Engineering and Physical Sciences Research Council (EPSRC) via Programme Grants [Grant No. UKRI1010: High order mathematical and computational infrastructure for streamed data that enhance contemporary generative and large language models]. We acknowledge computational resources and support provided by the Imperial College Research Computing Service (DOI: \texttt{10.14469/hpc/2232}). For the purpose of open access, the authors have applied a Creative Commons Attribution (CC BY) licence to any Author Accepted Manuscript version arising.

{
\small
\bibliographystyle{plainnat}
\bibliography{references}
}

\clearpage
\appendix

\crefalias{section}{appendix}
\crefalias{subsection}{appendix}
\crefname{appendix}{Appendix}{Appendices}

\section{Proofs for Weak Learner Direction Selection}\label{app:QuadraticMinimization}

To prove \cref{prop:deltak_Qk}, we first study the behaviour of the quadratic model $Q_k(f)$ along a single ray $t \mapsto ft,\ t \in \mathbb{R}$. Because our weak learner family $\mathcal{F}$ is closed under scalar multiplication, minimizing over $\mathcal{F}$ is equivalent to minimizing over all rays generated by directions in $\mathcal{F}$:

\begin{lemma}[Ray Minimization of the Quadratic Model] \label{lem:ray_min}
Fix any nonzero direction $f \in \mathcal{H}$. The minimum of the quadratic model $Q_k(tf)$ over the ray $\{tf: t \in \mathbb{R}\}$ is given by
\begin{equation}
    \min_{t \in \mathbb{R}} Q_k(tf) = -\frac{1}{2}\delta_k(f)^2, 
    \quad 
    \text{attained at}
    \quad
    t^*(f) = \frac{\delta_k(f)}{\|f\|_{H_k}}.
\end{equation}
\end{lemma}
\begin{proof}
Let $h(t) := Q_k(tf) = t\langle g_k, f\rangle + \frac{1}{2}t^2\|f\|_{H_k}^2$. This is a scalar quadratic function in $t$. Because $\|f\|_{H_k}^2 > 0$, it is strictly convex and its unique global minimum is achieved where the derivative is zero:
\begin{equation}
    h'(t) = \langle g_k, f\rangle + t\|f\|_{H_k}^2 = 0 
    \qquad
    \implies
    \qquad
    t^* = \frac{-\langle g_k, f\rangle}{\|f\|_{H_k}^2} = \frac{\delta_k(f)}{\|f\|_{H_k}},
\end{equation}
where the last equality follows from the definition $\delta_k(f) = \frac{-\langle g_k, f\rangle}{\|f\|_{H_k}}$. Substituting $t^*$ back into the objective yields the minimal value
\begin{align}
    h(t^*) &= \left( \frac{-\langle g_k, f\rangle}{\|f\|_{H_k}^2} \right) \langle g_k, f\rangle + \frac{1}{2} \left( \frac{-\langle g_k, f\rangle}{\|f\|_{H_k}^2} \right)^2 \|f\|_{H_k}^2 \\
    &= -\frac{\langle g_k, f\rangle^2}{\|f\|_{H_k}^2} + \frac{1}{2}\frac{\langle g_k, f\rangle^2}{\|f\|_{H_k}^2} = -\frac{1}{2} \left( \frac{-\langle g_k, f\rangle}{\|f\|_{H_k}} \right)^2 = -\frac{1}{2}\delta_k(f)^2.
\end{align}
\end{proof}

With this lemma established, the proof of \cref{prop:deltak_Qk} follows naturally.

\begin{proof}[Proof of \cref{prop:deltak_Qk}]
Since \(\mathcal{F}\) is closed under scalar multiplication, we can rewrite
\[
Q_k(\mathcal{F})
=
\min_{f\in \mathcal{F}} Q_k(f)
=
\min_{\phi\in \mathcal{F}\setminus\{0\}}\left(\min_{t\in\mathbb{R}} Q_k(t\phi)\right).
\]
Applying \cref{lem:ray_min} to the inner minimization gives
\[
Q_k(\mathcal{F})
=
\min_{\phi\in \mathcal{F}\setminus\{0\}}
\left(-\frac12 \delta_k(\phi)^2\right)
=
-\frac12 \sup_{\phi\in \mathcal{F}\setminus\{0\}} \delta_k(\phi)^2.
\]
Because \(\delta_k(-\phi)=-\delta_k(\phi)\) and \(\mathcal{F}\) is closed under sign, we have
\[
\sup_{\phi\in \mathcal{F}\setminus\{0\}} \delta_k(\phi)^2
=
\delta_k(\mathcal{F})^2.
\]
Therefore,
\[
Q_k(\mathcal{F}) = -\frac12 \delta_k(\mathcal{F})^2.
\]

Now fix a nonzero direction \(\phi^\ast\in \mathcal{F}\) with \(\delta_k(\phi^\ast)\ge0\).

If \(\phi^\ast \in \argmax_{f\in \mathcal{F}\setminus\{0\}} \delta_k(f)\), then
\(\delta_k(\phi^\ast)=\delta_k(\mathcal{F})\), and \cref{lem:ray_min} yields
\[
Q_k\!\left(\frac{\delta_k(\phi^\ast)}{\|\phi^\ast\|_{H_k}}\phi^\ast\right)
=
-\frac12 \delta_k(\phi^\ast)^2
=
-\frac12 \delta_k(\mathcal{F})^2
=
Q_k(\mathcal{F}).
\]
Hence
\[
\frac{\delta_k(\phi^\ast)}{\|\phi^\ast\|_{H_k}}\phi^\ast
\in \argmin_{f\in \mathcal{F}} Q_k(f).
\]

Conversely, suppose
\[
f^\ast := \frac{\delta_k(\phi^\ast)}{\|\phi^\ast\|_{H_k}}\phi^\ast
\in \argmin_{f\in \mathcal{F}} Q_k(f).
\]
Then
\[
-\frac12 \delta_k(\phi^\ast)^2
=
Q_k(f^\ast)
=
Q_k(\mathcal{F})
=
-\frac12 \delta_k(\mathcal{F})^2.
\]
Thus \(\delta_k(\phi^\ast)^2=\delta_k(\mathcal{F})^2\). Since both quantities are nonnegative,
\[
\delta_k(\phi^\ast)=\delta_k(\mathcal{F}),
\]
so
\[
\phi^\ast \in \argmax_{f\in \mathcal{F}\setminus\{0\}} \delta_k(f).
\]

Finally, let \(f_{k+1}^{\mathrm{wN}} \in \argmin_{f\in \mathcal{F}} Q_k(f)\). If \(f_{k+1}^{\mathrm{wN}}=0\), then
\(Q_k(\mathcal{F})=0\), hence \(\delta_k(\mathcal{F})=0\), and the stated identities are immediate. Assume now that
\(f_{k+1}^{\mathrm{wN}}\neq0\). Since the whole ray \(\{t f_{k+1}^{\mathrm{wN}} : t\in\mathbb{R}\}\) lies in \(\mathcal{F}\),
the scalar function \(t\mapsto Q_k(t f_{k+1}^{\mathrm{wN}})\) is minimized at \(t=1\). By  \cref{lem:ray_min},
its unique minimizer is
\[
t^\ast
=
\frac{\delta_k(f_{k+1}^{\mathrm{wN}})}{\|f_{k+1}^{\mathrm{wN}}\|_{H_k}},
\]
so \(t^\ast=1\), i.e.
\[
\delta_k(f_{k+1}^{\mathrm{wN}})
=
\|f_{k+1}^{\mathrm{wN}}\|_{H_k}.
\]
Using the value formula from \cref{lem:ray_min} together with
\(Q_k(\mathcal{F})=-\frac12 \delta_k(\mathcal{F})^2\), we obtain
\[
-\frac12 \|f_{k+1}^{\mathrm{wN}}\|_{H_k}^2
=
-\frac12 \delta_k(f_{k+1}^{\mathrm{wN}})^2
=
Q_k(f_{k+1}^{\mathrm{wN}})
=
Q_k(\mathcal{F})
=
-\frac12 \delta_k(\mathcal{F})^2.
\]
Hence
\[
\|f_{k+1}^{\mathrm{wN}}\|_{H_k}
=
\delta_k(\mathcal{F}),
\]
and therefore also
\[
\delta_k(f_{k+1}^{\mathrm{wN}})
=
\delta_k(\mathcal{F}).
\]
Thus,
\[
-\langle g_k, f_{k+1}^{\mathrm{wN}}\rangle
=
\delta_k(f_{k+1}^{\mathrm{wN}})\,\|f_{k+1}^{\mathrm{wN}}\|_{H_k}
=
\|f_{k+1}^{\mathrm{wN}}\|_{H_k}^2.
\]
Combining the above one obtains that the cosine angle equivalently takes the form
\[
    \Theta_k
    =
    \frac{\langle f_{k+1}^{\mathrm{N}},f_{k+1}^{\mathrm{wN}}\rangle_{H_k}}{\|f_{k+1}^{\mathrm{N}}\|_{H_k}\,\|f_{k+1}^{\mathrm{wN}}\|_{H_k}}
    =
    \frac{\delta_k(\mathcal{F})}{\|f_{k+1}^{\mathrm{N}}\|_{H_k}}
    =
    \frac{\|f_{k+1}^{\mathrm{wN}}\|_{H_k}}{\|f_{k+1}^{\mathrm{N}}\|_{H_k}}.
\]
Finally, using the definition of the dual norm applied to the gradient $g_k = -H_k f_{k+1}^{\mathrm{N}}$, and the previous cosine angle identity, one obtains
\[
 \|g_k\|_{H_k}^* 
 =
 \|H_k^{-\frac{1}{2}}H_k f_{k+1}^{\mathrm{N}}\|
 = 
 \|H_k^{\frac{1}{2}} f_{k+1}^{\mathrm{N}}\|
 =
 \|f_{k+1}^{\mathrm{N}}\|_{H_k}
 =
 \frac{\delta_k(\mathcal{F})}{\Theta_k}
\]
This completes the proof.
\end{proof}

\clearpage

\section{Weak-AICN Proofs}\label{app:WeakAICNProofs}

\subsection{Proof of Weak-AICN is Damped weak-Newton}

\begin{proof}[Proof of \cref{theorem:wAICN_is_wNewton}] If $\delta_k(\mathcal{F})=0$, then \cref{prop:deltak_Qk} gives $Q_k(\mathcal{F})=0$. Since $0 \in \mathcal{F}$, $Q_k(0) = 0$, and
\begin{equation}
    M_k(f)
    =
    Q_k(f)+\frac{C_{\mathrm{est}}}{6}\|f\|_{H_k}^3 
    \geq 
    0 
    = 
    M_k(0),
\end{equation}
for all $f\in\mathcal{F}$, $M_k$ is also minimized at $0$. In this case every weak Newton minimizer is
$f_{k+1}^{\mathrm{wN}}=0$, and $\alpha_k=1$.

Assume now that $\delta_k(\mathcal{F})>0$, and fix a direction $f \in \mathcal{F} \setminus \{0\}$. Since $\mathcal{F}$ is scalable, $-f \in \mathcal{F}$. For any $t\geq0$,
\begin{equation}
    M_k(tf)
= -\delta_k(f)\,t\|f\|_{H_k}
  + \frac{1}{2} t^2\|f\|_{H_k}^2
  + \frac{C_{\mathrm{est}}}{6} t^3 \|f\|_{H_k}^3,
\end{equation}
and $M_k(tf)-M_k(-tf) = -2\,\delta_k(f)\,t\|f\|_{H_k}$.
Therefore, for each fixed radius $t\geq0$, the smaller of the two values $M_k(tf)$ and
$M_k(-tf)$ is attained at the sign for which $\delta_k(f)\geq0$. Thus minimizing $M_k$ over the scalable family $\mathcal{F}$
is equivalent to first orienting the direction so that $\delta_k(f) \geq 0$, and then minimizing over $t\geq 0$. Let
\begin{equation}
    r := t\|f\|_{H_k}\ge0,
    \qquad
    m_\delta(r) 
    := 
    -\delta r + \frac12 r^2 + \frac{C_{\mathrm{est}}}{6}r^3.
\end{equation}
For ease of notation, we will minimize over $r$ instead of $t$. Along an oriented direction $f$ with $\delta_k(f) \geq 0$, we have $M_k(tf)=m_{\delta_k(f)}(r).$
For fixed \(\delta\ge0\),
\begin{equation}
    m_\delta'(r) = -\delta + r + \frac{C_{\mathrm{est}}}{2}r^2,
    \qquad
    m_\delta''(r) = 1 + C_{\mathrm{est}}r,
\end{equation}
so \(m_\delta\) is strictly convex on \([0,\infty)\). Its unique minimizer on $[0, \infty)$ is \(r^*(0)=0\) when
\(\delta=0\), and for \(\delta>0\) it is the unique positive solution of $m'_\delta(r) = 0$, namely
\[
r^*(\delta)
=
\frac{-1+\sqrt{1+2C_{\mathrm{est}}\delta}}{C_{\mathrm{est}}}
=
\frac{2\delta}{1+\sqrt{1+2C_{\mathrm{est}}\delta}}.
\]

Define
\[
\phi(\delta) := \min_{r\ge0} m_\delta(r) = m_\delta(r^*(\delta)).
\]
We claim that \(\phi\) is strictly decreasing on \([0,\infty)\). If \(0<\delta_1<\delta_2\) and
\(r_1:=r^*(\delta_1)\), then \(r_1>0\) and
\[
\phi(\delta_2)
\le
m_{\delta_2}(r_1)
=
m_{\delta_1}(r_1) - (\delta_2-\delta_1)r_1
=
\phi(\delta_1) - (\delta_2-\delta_1)r_1
<
\phi(\delta_1).
\]
Also \(\phi(0)=0\), while for every \(\delta>0\),
\[
\phi(\delta)=m_\delta(r^*(\delta))<m_\delta(0)=0=\phi(0).
\]
Thus \(\phi\) is strictly decreasing on \([0,\infty)\). 

Consequently, since $\phi(\delta_k(f)) = \min_{t\geq0} M_k(tf)$, minimizing \(M_k\) over \(\mathcal{F}\) is equivalent to maximizing \(\delta_k(f)\). By
\cref{prop:deltak_Qk}, $\delta_k(f)$ is maximized by the weak Newton direction $f_{k+1}^{\mathrm{wN}}$, which additionally satisfies $\|f_{k+1}^{\mathrm{wN}}\|_{H_k} = \delta_k(f_{k+1}^{\mathrm{wN}}) = \delta_k(\mathcal{F})$.
Therefore the minimizer of \(M_k\) is
\[
r^*(\delta_k)\,\frac{f_{k+1}^{\mathrm{wN}}}{\|f_{k+1}^{\mathrm{wN}}\|_{H_k}}
=
\alpha_k f_{k+1}^{\mathrm{wN}},
\qquad
\alpha_k := \frac{r^*(\delta_k)}{\delta_k}
=
\frac{2}{1+\sqrt{1+2C_{\mathrm{est}}\delta_k}}.
\]
\newpage
Finally, as for the $\alpha^2_k$ identity, substituting \(r=\alpha_k\delta_k\) into the optimality condition
\[
-\delta_k + r + \frac{C_{\mathrm{est}}}{2}r^2 = 0
\]
gives
\[
\frac{C_{\mathrm{est}}\delta_k}{2}\alpha_k^2 = 1-\alpha_k.
\]
This concludes the proof.
\end{proof}

\subsection{Proof of Weak Improvement $\Theta^2_k$ Factor}

\begin{proof}[Proof of \cref{prop:weak_cubic_improvement_theta_factor}]
By \cref{theorem:wAICN_is_wNewton} we can write
\begin{align}
    M_k(f^{\mathrm{wAICN}}_{k+1})
    &=
    \langle g_k, f^{\mathrm{wAICN}}_{k+1}\rangle + \frac{1}{2}\|f^{\mathrm{wAICN}}_{k+1}\|^2_{H_k} + \frac{C_{\mathrm{est}}}{6} \|f^{\mathrm{wAICN}}_{k+1}\|^3_{H_k} \\
    &=
    -\alpha_k \|f^{\mathrm{wN}}_{k+1}\|_{H_k}^2 + \frac{\alpha^2_k}{2} \|f^{\mathrm{wN}}_{k+1}\|_{H_k}^2 + \frac{C_{\mathrm{est}}\alpha_k^3}{6}\|f^{\mathrm{wN}}_{k+1}\|_{H_k}^3 \\
    &= 
    \alpha_k\|f^{\mathrm{wN}}_{k+1}\|_{H_k}^2 \left(  -1 + \frac{\alpha_k}{2} + \frac{C_{\mathrm{est}} \alpha_k^2}{6}\|f^{\mathrm{wN}}_{k+1}\|_{H_k} \right) \\
    &=
    \alpha_k\|f^{\mathrm{wN}}_{k+1}\|_{H_k}^2 \left(  -1 + \frac{\alpha_k}{2} + \frac{1-\alpha_k}{3} \right) \\
    &=
    \frac{\alpha_k(\alpha_k-4)}{6}\|f^{\mathrm{wN}}_{k+1}\|_{H_k}^2
\end{align}
where we also used \cref{prop:deltak_Qk} for the inner product with the gradient, and \cref{theorem:wAICN_is_wNewton} for the $C_{\mathrm{est}}\alpha_k^2$ identity. This proves the first part. Next, define 
\begin{equation}
    \alpha(s):=\frac{2}{1+\sqrt{1+2C_{\mathrm{est}}s}}\in(0,1],
    \qquad \quad
    \psi(s)
    \ :=\
    \frac{\alpha(s)\bigl(4-\alpha(s)\bigr)}{6}\,s^2
    .
\end{equation}
Since $s\mapsto \alpha(s)$ is decreasing and $a\mapsto a(4-a)$ is increasing on $(0,1]$,
the prefactor $\frac{\alpha(s)(4-\alpha(s))}{6}$ is decreasing in $s$. Hence for any $t\leq s$,
\[
\psi(t)
=
\frac{\alpha(t)\bigl(4-\alpha(t)\bigr)}{6}\,t^2
\ \ge\
\frac{\alpha(s)\bigl(4-\alpha(s)\bigr)}{6}\,(t)^2
=
\left(\frac{t}{s}\right)^2\psi(s).
\]
Applying this with the choices
\[
s := \|f_{k+1}^{N}\|_{H_k},
\qquad
t := \|f_{k+1}^{\mathrm{wN}}\|_{H_k},
\]
yields
\[
\psi(t)\geq \Theta_k^2 \psi(s).
\]
Here we used that \(t\leq s\) and
\[
\Theta_k
=
\frac{\langle f_{k+1}^{N}, f_{k+1}^{\mathrm{wN}}\rangle_{H_k}}
{\|f_{k+1}^{N}\|_{H_k}\|f_{k+1}^{\mathrm{wN}}\|_{H_k}}
=
\frac{\|f_{k+1}^{\mathrm{wN}}\|_{H_k}}{\|f_{k+1}^{N}\|_{H_k}}
=
\frac{t}{s},
\]
which follows from \cref{prop:deltak_Qk}.
Since \(M_k(\mathcal{F})=-\psi(t)\) and \(M_k(\mathcal{H})=-\psi(s)\), this yields
\[
M_k(\mathcal{F})\leq \Theta_k^2 M_k(\mathcal{H}),
\]
as claimed.
\end{proof}

\subsection{Proof of Weak Contracting-point Inequality}

\begin{proof}[Proof of \cref{weak-aicn-contracting-point}]
By the upper Taylor bound \eqref{eq:F-t3} from \cref{cubic-taylor-bound} applied to the weak-AICN update, we obtain
\begin{align}
    L(F_{k+1}) 
    &\leq  
    L(F_k) + M_k(\eta f^{\mathrm{wAICN}}_{k+1}) \\
    &\leq 
    L(F_k) + \eta M_k( f^{\mathrm{wAICN}}_{k+1}) \\
    &\leq 
    L(F_k) + \eta \Theta_k^2 M_k(f^{AICN}_{k+1}) \\
    &=
    (1-\eta \Theta_k^2) L(F_k) + \eta \Theta_k^2 \left( L(F_k) + \min_{f \in \mathcal{H}} M_k(f) \right), \label{eq:wCP-proof1}
\end{align}
where the second inequality follows by Jensen's inequality (sum of linear and convex terms), and the third by \cref{prop:weak_cubic_improvement_theta_factor}. Additionally, applying the lower Taylor bound \eqref{eq:F-t3} at $f = Y-F_k$ yields 
\begin{equation*}
    L(F_k) + Q_k(Y-F_k) \leq L(Y) + \frac{C_{\mathrm{ssc}}}{6}\|Y-F_k\|_{H_k}^3.
\end{equation*}
Adding the cubic penalty $\frac{C_{\mathrm{est}}}{6}\|Y-F_k\|_{H_k}^3$ to both sides to form $M_k(Y-F_k)$ and applying the assumption $C_{\mathrm{ssc}} \leq C_{\mathrm{est}}$, we obtain
\begin{align}
    L(F_k) + M_k(Y-F_k) 
    &\leq L(Y) + \frac{C_{\mathrm{ssc}} + C_{\mathrm{est}}}{6} \|Y-F_k\|_{H_k}^3 \notag \\
    &\leq L(Y) + \frac{C_{\mathrm{est}}}{3} \|Y-F_k\|_{H_k}^3. \label{eq:wCP-proof2}
\end{align}
Inserting \eqref{eq:wCP-proof2} into \eqref{eq:wCP-proof1} yields the conclusion.
\end{proof}

\subsection{Proof of Recurrence Lemmas}

\begin{proof}[Proof of \cref{lemmaAlpha3div2Recursion}]
    If any $\Delta_i=0$, all subsequent terms vanish and the bound is immediate for those terms. It therefore suffices to prove the bound for $k$ such that $\Delta_0,\dots,\Delta_k>0$.
    The proof is based on a telescoping sum argument. We start by writing the difference between two consecutive terms as
    \begin{align}
        \frac{1}{\sqrt{\Delta_{k+1}}} - \frac{1}{\sqrt{\Delta_k}}
        &= \frac{\sqrt{\Delta_k} - \sqrt{\Delta_{k+1}}}{\sqrt{\Delta_k \Delta_{k+1}}} \\
        &= \frac{\Delta_k - \Delta_{k+1}}{\sqrt{\Delta_k \Delta_{k+1}} (\sqrt{\Delta_k} + \sqrt{\Delta_{k+1}})}.
    \end{align}
    Substituting the recurrence $\Delta_k - \Delta_{k+1} \geq c_k \Delta_k^{3/2}$ into the numerator gives
    \begin{equation}
        \frac{1}{\sqrt{\Delta_{k+1}}} - \frac{1}{\sqrt{\Delta_k}} \geq \frac{c_k \Delta_k^{3/2}}{\sqrt{\Delta_k \Delta_{k+1}} (\sqrt{\Delta_k} + \sqrt{\Delta_{k+1}})}.
    \end{equation}
    Because $c_k \geq 0$ and $\Delta_k \geq 0$ by assumption, the sequence is non-increasing ($\Delta_{k+1} \leq \Delta_k$), and we obtain that
    \begin{align}
        \frac{1}{\sqrt{\Delta_{k+1}}} - \frac{1}{\sqrt{\Delta_k}}
        \geq \frac{c_k \Delta_k^{3/2}}{\sqrt{\Delta_k^2} (\sqrt{\Delta_k} + \sqrt{\Delta_k})}
        = \frac{c_k}{2}.
    \end{align}
    Summing these terms from $i=0$ to $k-1$ yields the telescoping sum
    \begin{equation}
        \frac{1}{\sqrt{\Delta_k}} \geq \frac{1}{\sqrt{\Delta_0}} + \frac{1}{2}\sum_{i=0}^{k-1} c_i.
    \end{equation}
    We must also bound the initial term $\Delta_0$. To this end, observe that the recurrence at $k=0$ implies $0\leq \Delta_1 \leq \Delta_0 - c_0\Delta_0^\frac{3}{2}$. Rearranging this inequality yields $1/\sqrt{\Delta_0} \geq c_0$. Inserting this into the summation leads to
    \begin{equation}
        \frac{1}{\sqrt{\Delta_k}} \geq c_0 + \frac{1}{2}\sum_{i=0}^{k-1} c_i = \frac{1}{2}\left( 2c_0 + \sum_{i=0}^{k-1} c_i \right).
    \end{equation}
    The result follows immediately by squaring and inverting the inequality.
\end{proof}

\begin{proof}[Proof of \cref{lemmaLinearRecursion}]
    Recursively applying the hypothesis $\Delta_{i+1} \leq (1-c_i)\Delta_i$ for $i=k-1, \dots, 0$ immediately yields
    \begin{equation}
        \Delta_k \leq \Delta_0 \prod_{i=0}^{k-1} (1-c_i).
    \end{equation}
    Using the standard inequality $1-x \leq e^{-x}$, valid for all real $x$, we bound each term in the product:
    \begin{equation}
        \prod_{i=0}^{k-1} (1-c_i) \leq \prod_{i=0}^{k-1} e^{-c_i} = \exp\!\left(-\sum_{i=0}^{k-1} c_i\right).
    \end{equation}
    Substituting the average $\overline{c_k}$ into the exponent completes the proof.
\end{proof}

\subsection{Proof of Global Convergence with Weak Learners}

\begin{proof}[Proof of \cref{thm:aicn_global_rate}]
Let $\Delta_k:=L(F_k)-L(F^*)\geq 0$. We apply \eqref{eq:w-cp} with
\[
    Y := (1-\beta)F_k+\beta F^*,\qquad \beta\in[0,1].
\]
By convexity, 
\[
L(Y)\leq (1-\beta)L(F_k)+\beta L(F^*)=L(F^*)+(1-\beta)\Delta_k.
\]
Also, $Y-F_k=\beta(F^*-F_k)$, so $\|Y-F_k\|_{H_k}\leq \beta D$ by \eqref{eq:F-aicn-bd}.
Thus \eqref{eq:w-cp} implies
\begin{equation}\label{eq:recurrence-gen}
    \Delta_{k+1}
    \le
    (1 - \eta \Theta_k^2 \beta)\Delta_k + \eta \Theta_k^2\frac{C_{\mathrm{est}}}{3} \beta^3 D^3.
\end{equation}
First, note that $\Delta_{k+1} \leq \Delta_k$ which follows from setting $\beta=0$ above. Next, we choose
$$
\beta := \min\left( 1, \sqrt{\frac{\Delta_k}{C_{\mathrm{est}}D^3}}\right).
$$

\textbf{Case 1: Linear Regime ($\beta=1$):} Since $\Delta_k$ is monotone, $\beta$ will remain equal to one until the condition $\Delta_k < C_{\mathrm{est}}D^3$ is met, where it will then remain strictly less than one. In the case $\beta=1$, we have that $\Delta_k \geq C_{\mathrm{est}}D^3$. Factoring $\Delta_k$ out of \eqref{eq:recurrence-gen} yields
\[
\Delta_{k+1} \leq \left(1 - \eta\Theta^2_k + \eta\Theta^2_k \frac{C_{\mathrm{est}}D^3}{3\Delta_k}\right)\Delta_k \leq \left(1 - \frac{2}{3}\eta\Theta^2_k\right)\Delta_k.
\]
Applying \cref{lemmaLinearRecursion} to this multiplicative recurrence gives the linear rate:
\[
\Delta_k \leq \Delta_0 \exp\left(-\frac{2}{3}\eta k \overline{\Theta_k^2}\right).
\]
By definition, $k_0$ is the first iteration where $\Delta_{k_0} \leq C_{\mathrm{est}}D^3$. If $k_0>1$, then $\Delta_{k_0-1} > C_{\mathrm{est}}D^3$. Evaluating our linear rate at $k = k_0 - 1$ and chaining these inequalities yields
\[
C_{\mathrm{est}}D^3 < \Delta_{k_0-1} \leq \Delta_0 \exp\left(-\frac{2}{3}\eta (k_0-1) \overline{\Theta_{k_0-1}^2}\right).
\]
Dividing by $\Delta_0$, taking the logarithm of both sides, and rearranging for $k_0$ gives the stated bound on the length of the initial phase.

\textbf{Case 2: Global Regime ($\beta < 1$):}
Once $\Delta_{k_0} \leq C_{\mathrm{est}}D^3$, monotonicity implies $\Delta_k \leq C_{\mathrm{est}}D^3$ for all $k \geq k_0$. Thus for $k \geq k_0$, we have $\beta = \sqrt{\frac{\Delta_k}{C_{\mathrm{est}}D^3}}$.
Substituting this $\beta$ into \eqref{eq:recurrence-gen} yields
\[
\Delta_{k+1}
\leq
\Delta_k - \frac{2\eta\Theta_k^2}{3\sqrt{C_{\mathrm{est}}D^3}}\Delta_k^{3/2}.
\]
Let $\gamma_t := \Delta_{k_0+t}$ for $t \geq 0$. The sequence satisfies $\gamma_{t+1} \leq \gamma_t - c_t \gamma_t^{3/2}$ with sequence constants $c_t = \frac{2\eta\Theta_{k_0+t}^2}{3\sqrt{C_{\mathrm{est}}D^3}}$. Applying \cref{lemmaAlpha3div2Recursion} directly gives
\[
\gamma_t \leq \frac{4}{\overline{c_t}^2 t^2} = \frac{9\,C_{\mathrm{est}}\,D^3}{\eta^2 (\overline{\Theta_{k_0, k_0+t}^2})^2 t^2}.
\]
Substituting $t = k - k_0$ back into the inequality concludes the proof.
\end{proof}

\begin{remark}
    If $\eta=1$ and $\mathcal{F} = \mathcal{H}$, then $f^{\mathrm{wN}}_{k+1} = f^{\mathrm{N}}_{k+1}$ and $\Theta_k = 1$. Applying \eqref{eq:recurrence-gen} at $k=0$ with $\beta=1$ yields $\Delta_1 \leq \frac{1}{3}C_{\mathrm{est}}D^3 < C_{\mathrm{est}}D^3$. This implies that $k_0 \leq 1$ when not using weak learners.
\end{remark}

\newpage
\subsection{Proof of Local Linear Convergence with Weak Learners}


\begin{lemma}\label{lemma:SC-hess-ineq}
    Write $\delta_k := \delta_k(f_{k+1}^{\mathrm{wN}})$. If $ C_{\mathrm{est}} \|\eta f_{k+1}^{\mathrm{wAICN}}\|_{H_k} = C_{\mathrm{est}} \eta \alpha_k \delta_k < 1$ then
    \begin{align}
        \left( 1- C_{\mathrm{est}} \eta \alpha_k \delta_k \right)^2 H_{k+1}
        \preceq
        H_k
        \preceq
        \frac{1}{\left( 1- C_{\mathrm{est}} \eta \alpha_k \delta_k \right)^2} H_{k+1}
    \end{align}
\end{lemma}
\begin{proof}
    Follows from \cite[Theorem 5.1.7]{2018LecturesOnConvexOptimization} applied to the weak-AICN update $f_{k+1}^{\mathrm{^{wAICN}}}$ since semi-strong self-concordance implies self-concordance.
\end{proof}

\begin{proof}[Proof of \cref{thm:local_linear}]
    The weak-AICN update is $F_{k+1} = F_k + s_k$, where $s_k = \eta \alpha_k f^{\mathrm{wN}}_{k+1} = \eta f^{\mathrm{wAICN}}_{k+1}$. Using the fundamental theorem of calculus, we expand the gradient at the next weak-AICN iterate:
    \begin{align}
        g_{k+1} 
        &=
        g_k + \int_0^1 \nabla^2 L(F_k + t s_k) s_k \, dt \\
        &=
        \underbrace{g_k + H_k s_k}_{v_{\mathrm{lin}}} + \underbrace{\int_0^1 \left( \nabla^2 L(F_k + t s_k) - H_k \right) s_k \, dt}_{e_k}.
    \end{align}
    We analyze the norms of $v_{\mathrm{lin}}$ and $e_k$ separately.

    \textbf{1. Bound on the Linear Term $v_{\mathrm{lin}}$:}
    Recall that $g_k = -H_k f^{\mathrm{N}}_{k+1}$. Thus, $v_{\mathrm{lin}} = -H_k f^{\mathrm{N}}_{k+1} + \eta \alpha_k H_k f^{\mathrm{wN}}_{k+1}$. Taking squared dual norms we see that
    \begin{align}
        \|v_{\mathrm{lin}}\|_{H_k}^{*2} &= \| -f^{\mathrm{N}}_{k+1} + \eta \alpha_k f^{\mathrm{wN}}_{k+1} \|_{H_k}^2 \\
        &= \|f^{\mathrm{N}}_{k+1}\|_{H_k}^2 + \eta^2 \alpha_k^2 \|f^{\mathrm{wN}}_{k+1}\|_{H_k}^2 - 2\eta\alpha_k \langle f^{\mathrm{N}}_{k+1}, f^{\mathrm{wN}}_{k+1} \rangle_{H_k}.
    \end{align}
    Write $\delta_k = \delta_k(f_{k+1}^{\mathrm{wN}})$ for brevity. By \cref{prop:deltak_Qk} we have the identities $\langle f^{\mathrm{N}}_{k+1}, f^{\mathrm{wN}}_{k+1} \rangle_{H_k} = \delta_k^2$ and   $\|f^{\mathrm{wN}}_{k+1}\|_{H_k} = \delta_k = \Theta_k \|f^{\mathrm{N}}_{k+1}\|_{H_k}$. Substituting these into the equation yields
    \begin{align}
        \|v_{\mathrm{lin}}\|^{*2}_{H_k} &= \left(\frac{\delta_k}{\Theta_k}\right)^2 + \eta^2 \alpha_k^2 \delta_k^2 - 2\eta \alpha_k \delta_k^2 \\
        &=
        \left(\frac{\delta_k}{\Theta_k}\right)^2 \left[ 1 - \Theta_k^2 \eta \alpha_k (2 - \eta \alpha_k) \right] \\
        &=
        \|g_k\|_{H_k}^{*2} \left[ 1 - \Theta_k^2 \eta \alpha_k (2 - \eta \alpha_k) \right].
    \end{align}
    Taking square roots and using the bounds $\sqrt{1-x} \leq 1-\frac{x}{2}$ and $\eta \alpha_k \leq 1$ gives
    \begin{align}
        \|v_{\mathrm{lin}}\|_{H_k}^* 
        &=
        \|g_k\|_{H_k}^* \sqrt{1 - \Theta_k^2 \eta \alpha_k (2 - \eta \alpha_k)} \\
        &\leq
        \|g_k\|_{H_k}^* \left( 1 - \Theta_k^2 \eta \alpha_k \frac{2 - \eta \alpha_k}{2} \right) \\
        &\leq
        \|g_k\|_{H_k}^* \left( 1 - \frac{\Theta_k^2 \eta \alpha_k}{2} \right).
    \end{align}

    \textbf{2. Bound on the Error Term $e_k$:}
    Using the definition of the Hessian-induced affine-invariant operator norm and semi-strong self-concordance we find that
    \begin{align}
        \|e_k\|_{H_k}^* &\leq \int_0^1 \|\left( \nabla^2 L(F_k + t s_k) - H_k \right) s_k \|_{H_k}^* \, dt \\
        &\leq \int_0^1 \|\nabla^2 L(F_k + t s_k) - H_k \|_{\mathrm{aff-op}, H_k} \|s_k\|_{H_k} \, dt \\
        &\leq \int_0^1 C_{\mathrm{est}} \| t s_k \|_{H_k} \|s_k\|_{H_k} \, dt \\
        &= \frac{C_{\mathrm{est}}}{2} \|s_k\|_{H_k}^2.
    \end{align}
    Substituting $\|s_k\|_{H_k} = \eta \alpha_k \delta_k$ and $\|g_k\|^*_{H_k} = \frac{\delta_k}{\Theta_k}$ gives the final bound
    \begin{equation}
        \label{eq:e_k_bound}
        \|e_k\|_{H_k}^* 
        \leq
        \|g_k\|^*_{H_k} \frac{C_{\mathrm{est}}}{2} \eta^2 \alpha_k^2 \delta_k \Theta_k.
    \end{equation}

    \textbf{3. Change of Metric:}
    We must measure the gradient $g_{k+1}$ in the norm induced by $H_{k+1}$. Let $r_k := \|s_k\|_{H_k} = \eta \alpha_k \delta_k$. By \cref{lemma:SC-hess-ineq} we have for $C_{\mathrm{est}} r_k < 1$ and all $u\in\mathcal{H}$ that
    \begin{equation}
        \|u\|_{H_{k+1}}^* \leq \frac{1}{1 - C_{\mathrm{est}} r_k} \|u\|_{H_k}^*.
    \end{equation}
    Applying this to $u = g_{k+1} = v_{\mathrm{lin}} + e_k$ and substituting the above bounds yields
    \begin{align}
        \|g_{k+1}\|_{H_{k+1}}^* 
        &\leq
        \frac{1}{1 - C_{\mathrm{est}} \eta \alpha_k \delta_k} \left( \|v_{\mathrm{lin}}\|_{H_k}^* + \|e_k\|_{H_k}^* \right) \\
        &\leq
        \frac{\|g_k\|_{H_k}^*}{1 - C_{\mathrm{est}} \eta \alpha_k \delta_k} \left( 1 - \frac{ \Theta_k^2 \eta \alpha_k}{2} + \frac{1}{2}C_{\mathrm{est}}\delta_k \Theta_k \eta^2 \alpha_k^2 \right) \\
        &=: \|g_k\|_{H_k}^* \frac{\mathcal{N}}{\mathcal{D}}.
    \end{align}
    \textbf{4. Contraction:}
    Denote the numerator and denominator of the fraction multiplying $\|g_k\|_{H_k}^*$ by $\mathcal{N}$ and $\mathcal{D}$ respectively.
    We proceed by analyzing the difference $\mathcal{D} - \mathcal{N}$.
    \begin{align}
        \mathcal{D} - \mathcal{N} 
        &= \left( 1 - C_{\mathrm{est}}\eta\alpha_k\delta_k \right) - \left( 1 - \frac{\Theta_k^2 \eta \alpha_k}{2} + \frac{1}{2}C_{\mathrm{est}}\delta_k \Theta_k \eta^2 \alpha_k^2 \right) \\
        &= \frac{\Theta_k^2 \eta \alpha_k}{2} - C_{\mathrm{est}}\eta\alpha_k\delta_k - \frac{1}{2}C_{\mathrm{est}}\delta_k \Theta_k \eta^2 \alpha_k^2 \\
        &= \eta \alpha_k \left[ \frac{\Theta_k^2}{2} - C_{\mathrm{est}}\delta_k \left( 1 + \frac{\eta \alpha_k \Theta_k}{2} \right) \right].
    \end{align}
    Since $\eta, \alpha_k, \Theta_k \in (0, 1]$, we have $1 + \frac{\eta \alpha_k \Theta_k}{2} \leq \frac{3}{2}$. Substituting this bound and the hypothesis $\delta_k \leq \frac{\Theta_k^2}{6 C_{\mathrm{est}}}$ gives
    \begin{align}
        \mathcal{D} - \mathcal{N} 
        &\geq \eta \alpha_k \left[ \frac{\Theta_k^2}{2} - \frac{3}{2}C_{\mathrm{est}}\delta_k \right] \\
        &\geq \eta \alpha_k \left[ \frac{\Theta_k^2}{2} - \frac{3}{2}C_{\mathrm{est}} \left( \frac{\Theta_k^2}{6 C_{\mathrm{est}}} \right) \right] \\
        &= \frac{1}{4} \eta \alpha_k \Theta_k^2.
    \end{align}
    Finally, since $\delta_k \leq \frac{1}{6 C_{\mathrm{est}}}$ and $\eta \alpha_k \leq 1$, the denominator satisfies $\mathcal{D} = 1 - C_{\mathrm{est}}\eta\alpha_k\delta_k \geq 5/6$, and specifically $\mathcal{D} \leq 1$. Therefore,
    \begin{equation}
        \frac{\mathcal{N}}{\mathcal{D}} = 1 - \frac{\mathcal{D}-\mathcal{N}}{\mathcal{D}} \leq 1 - (\mathcal{D}-\mathcal{N}) \leq 1 - \frac{1}{4} \eta \alpha_k \Theta_k^2.
    \end{equation}
    Setting $\rho_k = \frac{1}{4} \eta \alpha_k \Theta_k^2$ completes the proof of the one-step contraction.

    \textbf{5. Final Linear Convergence:} The inner product on $\mathcal{H}$ can be bounded by the product of the Hessian-induced norm and its dual. For any $u, v \in \mathcal{H}$ we have by Cauchy-Schwarz inequality
    \begin{equation}
        \langle u, v \rangle = \langle H_k^{-1/2} u, H_k^{1/2} v \rangle \leq \|H_k^{-1/2} u\| \|H_k^{1/2} v\| = \|u\|_{H_k}^* \|v\|_{H_k}.
    \end{equation}
    Applying this to $g_k$ and $(F_k - F^*)$, after using the convexity of $L$, yields
    \begin{equation}
        L(F_k) - L(F^*) 
        \leq
        \langle g_k, F_k - F^* \rangle
        \leq
        \|g_k\|_{H_k}^* \|F_k - F^*\|_{H_k}
        \leq 
        \|g_k\|_{H_k}^* D,
    \end{equation}
    where the final inequality follows from \cref{F-aicn-bounded-radius}. Finally, applying \cref{lemmaLinearRecursion} to the multiplicative recurrence $\|g_{k+1}\|^*_{H_{k+1}} \leq (1 - \rho_k) \|g_k\|^*_{H_k}$ gives the result.
\end{proof}

\begin{remark}[Sufficient conditions for the bounded local radius]\label{remark:sufficient_condition_for_bounded_local_radius}

As previously discussed, a bounded-radius condition of the form
\cref{F-aicn-bounded-radius} is also used in classical AICN analysis
\cite{2022ADampedNewtonMethodAchievesGlobalMinus2AndLocalQuadraticConvergenceRate}.
However, simple sufficient conditions for this bound follow
from standard regularity conditions on $L$.
Under the assumptions of \cref{weak-aicn-contracting-point},
taking $Y=F_k$ gives $L(F_{k+1})\leq L(F_k)$.
If $L$ is additionally $\mu$-strongly convex, then
\[
    \|F_k-F^*\|^2\leq R_0^2
    :=\frac{2}{\mu}\bigl(L(F_0)-L(F^*)\bigr).
\]
If $L$ is also $S$-smooth, then
\[
    \|F_k-F^*\|_{H_k}^2
    \leq S\|F_k-F^*\|^2
    \leq SR_0^2,
\]
hence \cref{F-aicn-bounded-radius} holds with $D=\sqrt{S}\,R_0$.
Alternatively, if its Hessian is $M$-Lipschitz, then
\[
    \|H_k\|_{\mathrm{op}}
    \leq \|\nabla^2L(F^*)\|_{\mathrm{op}}+M\|F_k-F^*\|
    \leq \|\nabla^2L(F^*)\|_{\mathrm{op}}+MR_0,
\]
and hence
\[
    \|F_k-F^*\|_{H_k}^2
    \leq \|H_k\|_{\mathrm{op}}\|F_k-F^*\|^2
    \leq \bigl(\|\nabla^2L(F^*)\|_{\mathrm{op}}+MR_0\bigr)R_0^2
    =:D^2.
\]
For examples of radius assumptions in stochastic zeroth-order
and bandit convex optimization, see
\cite{2024StochasticZerothOrderOptimizationUnderStronglyConvexityAndLipschitzHessianMinimaxSampleComplexity,2023ASecondOrderMethodForStochasticBanditConvexOptimisation};
note that these settings and results are not directly comparable to ours.
\end{remark}

\subsection{Proof of Linear Rate for Hessian-dominated Losses}

\begin{proof}[Proof of \cref{thm:hessDominatedLinear}]
Write $\delta_k = \delta_k(f^{\mathrm{wN}}_{k+1})$. We start by applying the affine-invariant cubic bound (\cref{cubic-taylor-bound}) to the weak AICN update, which by \cref{theorem:wAICN_is_wNewton} takes the form $f_{k+1}^{\mathrm{wAICN}} = \alpha_k f_{k+1}^{\mathrm{wN}}$ gives
\begin{align*}
    L(F_k) - L(F_{k+1}) 
    &\geq -Q_k(\eta f_{k+1}^{\mathrm{wAICN}}) - \frac{C_{\mathrm{est}}}{6}\|\eta f_{k+1}^{\mathrm{wAICN}}\|_{H_k}^3 \\
    &= \eta \alpha_k \delta_k^2 - \frac{1}{2} (\eta \alpha_k)^2 \delta_k^2 - \frac{C_{\mathrm{est}}}{6} (\eta \alpha_k)^3 \delta_k^3 \\
    &= \eta \alpha_k \delta_k^2 \left( 1 - \frac{1}{2} \eta \alpha_k - \eta^2 \left[ \frac{C_{\mathrm{est}}\delta_k}{6}\alpha_k^2 \right] \right).
\end{align*}
To simplify the bracketed term, we recall from \cref{theorem:wAICN_is_wNewton} that $\alpha_k$ satisfies
\(
    \frac{C_{\mathrm{est}}\delta_k}{6}\alpha_k^2 = \frac{1-\alpha_k}{3}.
\)
Substituting this back into the descent inequality, we obtain
\begin{align*}
    L(F_k) - L(F_{k+1}) 
    &\geq \eta \alpha_k \delta_k^2 \left( 1 - \frac{1}{2} \eta \alpha_k - \eta^2 \frac{1-\alpha_k}{3} \right) \\
    &= \eta \alpha_k \delta_k^2 \left( 1 - \frac{\eta^2}{3} - \alpha_k \left[ \frac{\eta}{2} - \frac{\eta^2}{3} \right] \right) \\
    &\geq \eta \alpha_k \delta_k^2 \left( 1 - \frac{\eta}{2} \right) \\
    &= \eta \alpha_k \Theta_k^2 \|f^{N}_{k+1}\|_{H_k}^2 \left( 1 - \frac{\eta}{2} \right)
\end{align*}
The final inequality follows because $\eta \in (0, 1]$ implies $\frac{\eta}{2} \geq \frac{\eta^2}{3}$, meaning the term in square brackets is non-negative. Since we subtract this term scaled by $\alpha_k < 1$, the expression is minimized as $\alpha_k \to 1$. Finally, applying the Hessian-dominance condition $\|f^{N}_{k+1}\|_{H_k}^2 \geq c\bigl(L(F_k)-L_\star\bigr)$ and rearranging yields the result.
\end{proof}



\clearpage

\section{Semi-Strong Self-Concordance in the Boosting Space}\label{app:SSC}

To prove the inheritance of the SSC property for empirical risks (\cref{thm:ssc_inheritance}), we first explicitly state the SSC condition for 1D scalar functions.

\begin{lemma}[One-dimensional SSC]\label{lem:ssc_1d}
Let $l:\mathbb{R}\to\mathbb{R}$ be twice differentiable with $l''(z)>0$. The semi-strong self-concordance condition (\cref{def:semi-strong-self-concordance}) is equivalent in one dimension to:
\[
\frac{|l''(z+t)-l''(z)|}{l''(z)} \le C_l\,|t|\sqrt{l''(z)} \qquad \forall z,t\in\mathbb{R}.
\]
\end{lemma}
\begin{proof}
In one dimension, the Hessian operator reduces to the positive scalar $h_z = l''(z)$. The affine-invariant operator norm of a scalar $a$ is therefore $\|a\|_{\mathrm{aff\mbox{-}op}, h_z} = \|h_z^{-1/2} a h_z^{-1/2}\|_{\mathrm{op}} = |a|/h_z$. The Hessian norm of $t$ is $\|t\|_{h_z} = |t|\sqrt{h_z}$. Substituting these into the definition of SSC yields the result.
\end{proof}

\begin{proof}[Proof of \cref{thm:ssc_inheritance}]
We operate in the space $\mathcal{H} = \mathbb{R}^n$ equipped with the empirical inner product $\langle u, v\rangle_X = \frac{1}{n}\sum_{i=1}^n u_i v_i$. The empirical risk is $L(F) = \frac{1}{n}\sum_{i=1}^n l(F(x_i), y_i)$. For any iterate $F \in \mathcal{H}$ and direction $f \in \mathcal{H}$, the Hessian operator $H_F := \nabla^2 L(F)$ and its variation $\Delta := \nabla^2 L(F+f) - \nabla^2 L(F)$ act as diagonal linear operators on $\mathcal{H}$. Specifically, their representations in the standard basis are diagonal matrices with entries $d_i = l''(F(x_i), y_i)$ and $\Delta_i = l''(F(x_i)+f(x_i), y_i) - l''(F(x_i), y_i)$, respectively.

For any diagonal operator, the operator norm induced by $\langle \cdot, \cdot \rangle_X$ is simply the maximum absolute value of its entries. Therefore, the affine-invariant operator norm is
\[
    \|\Delta\|_{\mathrm{aff\mbox{-}op}, H_F} 
    \ = \ \|H_F^{-1/2} \Delta H_F^{-1/2}\|_{\mathrm{op}} 
    \ = \ \max_{1 \le i \le n} \frac{|\Delta_i|}{d_i}.
\]
Applying the 1D SSC condition (\cref{lem:ssc_1d}) to the base loss $l(\cdot, y_i)$, we bound each component by
\[
    \frac{|\Delta_i|}{d_i} 
    \ \le \ C_l |f(x_i)| \sqrt{l''(F(x_i), y_i)} 
    \ = \ C_l |f(x_i)| \sqrt{d_i}.
\]
Taking the maximum over all $i=1,\dots,n$, we obtain
\[
    \|\Delta\|_{\mathrm{aff\mbox{-}op}, H_F} \le C_l \max_{1 \le i \le n} \bigl(|f(x_i)|\sqrt{d_i}\bigr).
\]
Finally, we relate this maximum to the local Hessian norm $\|f\|_{H_F}$. By definition, the squared Hessian norm is the empirical average $\|f\|_{H_F}^2 = \frac{1}{n} \sum_{i=1}^n d_i f(x_i)^2$. Since the maximum element of a set of non-negative numbers is bounded by their sum, we have
\[
    \max_{1 \le i \le n} \bigl(d_i f(x_i)^2\bigr) 
    \ \le \ \sum_{i=1}^n d_i f(x_i)^2 
    \ = \ n \left( \frac{1}{n}\sum_{i=1}^n d_i f(x_i)^2 \right) 
    \ = \ n \|f\|_{H_F}^2.
\]
Taking the square root and substituting this into our operator norm bound yields
\[
    \|\Delta\|_{\mathrm{aff\mbox{-}op}, H_F} \le \sqrt{n} C_l \|f\|_{H_F},
\]
proving that the empirical risk $L$ is semi-strongly self-concordant with constant $C_{\mathrm{ssc}}(L) \le \sqrt{n} C_l$.
\end{proof}

\clearpage
\section{Newton Coordinate Descent: Lower Bound Proofs and Experiments}\label{app:CoordinateDescent}

\subsection{Proof of Newton Coordinate Descent Cosine Angle Lower Bound}
\begin{proof}[Proof of \cref{prop:coordinate_descent_lower_bound}]
    Recall from \cref{prop:deltak_Qk} that the cosine angle can be expressed as the ratio of directional improvements $\Theta_k = \delta_k(f_{k+1}^{\mathrm{wN}}) / \delta_k(f_{k+1}^{\mathrm{N}})$. We bound the numerator and denominator separately.

    \textbf{1. Lower bound on the weak improvement:}
    For coordinate weak learners, the maximal directional improvement is achieved at the optimal coordinate $i^*$. Using the standard Rayleigh quotient bound $(H_k)_{ii} \leq \lambda_{\max}(H_k)$, we have:
    \begin{equation} \label{eq:cd_num_bound}
        \delta_k(f_{k+1}^{\mathrm{wN}}) 
        \ = \ \max_{1 \leq i \leq d} \frac{|(g_k)_i|}{\sqrt{(H_k)_{ii}}} 
        \ \geq \ \max_{1 \leq i \leq d} \frac{|(g_k)_i|}{\sqrt{\lambda_{\max}(H_k)}} 
        \ = \ \frac{\|g_k\|_\infty}{\sqrt{\lambda_{\max}(H_k)}}.
    \end{equation}

    \textbf{2. Upper bound on the exact improvement:}
    By \cref{prop:deltak_Qk}, the exact directional improvement is exactly the dual Hessian norm of the gradient. Using the eigenvalue bound $H_k^{-1} \preceq \lambda_{\min}(H_k)^{-1} I$ and the vector norm equivalence $\|g_k\|_2 \leq \sqrt{d}\|g_k\|_\infty$, we obtain:
    \begin{equation}
        \delta_k(f^{\mathrm{N}}_{k+1}) 
        \ = \ \sqrt{\langle g_k, H_k^{-1} g_k \rangle} 
        \ \leq \ \sqrt{\frac{1}{\lambda_{\min}(H_k)} \|g_k\|_2^2} 
        \ \leq \ \sqrt{\frac{d}{\lambda_{\min}(H_k)}} \|g_k\|_\infty.
    \end{equation}

    \textbf{3. Final Ratio:}
    Substituting these two bounds into the cosine angle identity yields:
    \begin{equation}
        \Theta_k 
        \ = \ \frac{\delta_k(f_{k+1}^{\mathrm{wN}})}{\delta_k(f^{\mathrm{N}}_{k+1})} 
        \ \geq \ \frac{\|g_k\|_\infty / \sqrt{\lambda_{\max}(H_k)}}{\sqrt{d/\lambda_{\min}(H_k)} \|g_k\|_\infty} 
        \ = \ \sqrt{\frac{\lambda_{\min}(H_k)}{d \lambda_{\max}(H_k)}} 
        \ = \ \frac{1}{\sqrt{d \kappa_k}}.
    \end{equation}
    This completes the proof.
\end{proof}

\subsection{Algorithm for Newton Coordinate Descent}

For completeness and ease of use, we detail the exact computation for the Newton coordinate weak learner below.

\begin{algorithm}[H]\label{alg:NewtonCoordinateDescent}
\caption{Computing $f^{\mathrm{wN}}_{k+1}$ for coordinate weak learners}
\begin{algorithmic}[1]
\STATE \textbf{Input:} Diagonal of Hessian $\operatorname{diag}(H_k)$ and gradient $g_k$
\STATE $A \leftarrow g_k \odot g_k \oslash \operatorname{diag}(H_k)$
\STATE $i^* \leftarrow \argmax_{1 \leq i \leq d} |A_i|$
\STATE $f^{\mathrm{wN}}_{k+1} \leftarrow -\frac{(g_k)_{i^*}}{(H_k)_{i^*i^*}} e_{i^*}$
\end{algorithmic}
\end{algorithm}

\subsection{Adaptive Weak-AICN Algorithm}\label{app:AdaptiveWeakAICN}

In the classical finite-dimensional convex optimization literature, adaptive variants of Newton algorithms are often preferred in practice. Rather than relying on a fixed, global regularity constant (which can be difficult to compute or overly conservative), adaptive schemes dynamically estimate the local regularity during optimization. While the original authors of exact-AICN \cite{2022ADampedNewtonMethodAchievesGlobalMinus2AndLocalQuadraticConvergenceRate} did not explore such an algorithm, we provide an adaptive variant of weak-AICN below, which we call weak-AdaAICN.

Weak-AdaAICN operates using a backtracking line search that dynamically adjusts the local estimate $C_k$. At each iteration, if the empirical decrease in the loss does not match the decrease predicted by the cubic model, we double the estimate $C_k$ and recompute the stepsize $\alpha_k$. Crucially, this check is highly computationally efficient. Because \cref{prop:weak_cubic_improvement_theta_factor} allows us to explicitly evaluate the minimum of the cubic model as $M_k(\alpha_k f_{k+1}^{\mathrm{wN}}) = \frac{\alpha_k(\alpha_k-4)}{6}\delta_k^2$, validating the descent condition requires only scalar operations and a single loss evaluation, completely bypassing the need to re-evaluate the Hessian or the weak learner. 

From a theoretical standpoint, \cref{cubic-taylor-bound} guarantees that the required descent condition is satisfied whenever $C_k \ge C_{\mathrm{ssc}}$. As a result, the adaptive constant will never grow indefinitely and is bounded by $C_k \le 2\max(C_0, C_{\mathrm{ssc}})$.

\begin{algorithm}[H]
\caption{Adaptive Weak AICN}
\begin{algorithmic}[1]
\STATE \textbf{Input:} Convex loss $L$, Weak learner set $\mathcal{F}$, Learning rate $\eta \in (0, 1]$, Initial regularity constant $C_{0} > 0$ and initial value $F_0$.
\FOR{$k = 0, 1, \dots$}
    \STATE Compute gradient $g_k = \nabla L(F_k)$ and Hessian $H_k = \nabla^2 L(F_k)$.
        
    \STATE $f^{\mathrm{wN}}_{k+1} \leftarrow \operatorname{argmin}_{f \in \mathcal{F}} \left( \langle g_k, f \rangle + \frac{1}{2}\langle f, H_k f \rangle \right)$

    \STATE $\delta_k \leftarrow \sqrt{|\langle g_k, f^{\mathrm{wN}}_{k+1} \rangle |}$

    \REPEAT 
        \STATE Increase estimate  $C_{k} \leftarrow 2 C_{k}$
        \STATE $\alpha_k \leftarrow \frac{2}{1+\sqrt{1+2 C_k \delta_k}}$
        \STATE $M_k \leftarrow \frac{\alpha_k(\alpha_k-4)}{6} \delta_k^2$
        \STATE $F_{test} \leftarrow F_k + \alpha_k  f^{\mathrm{wN}}_{k+1}$
        
    \UNTIL{$L(F_{test}) \leq L(F_k) +  M_k$}
    
    \STATE Update $F_{k+1} \leftarrow F_k + \eta \alpha_k f^{\mathrm{wN}}_{k+1}$
    \STATE Decrease estimate $C_{k+1} \leftarrow C_{k} / 4$
\ENDFOR
\end{algorithmic}
\end{algorithm}

\subsection{Experimental Validation of weak-AdaAICN}

We implement weak-AdaAICN as an OPTAMI-compatible \cite{2025OPTAMI} PyTorch optimizer \cite{2019PyTorch} to validate its effectiveness against the non-adaptive AICN algorithm \cite{2022ADampedNewtonMethodAchievesGlobalMinus2AndLocalQuadraticConvergenceRate}. Following the experimental setup of the original AICN paper, we benchmark the methods on Nesterov's second-order lower bound function \cite{2021ImplementableTensorMethodsInUnconstrainedConvexOptimizationSecondOrderLowerBoundFunction}, with optional $\ell_2$ regularization $\mu$. For $x \in \mathbb{R}^d$, the objective is defined as
\begin{align}
    L(x) = \frac{1}{3} \sum_{i=1}^{d-1} |x_i - x_{i+1}|^3 + \frac{1}{3} |x_d|^3 - x_1 + \frac{\mu}{2}\|x\|_2^2.
\end{align}

For exactness and reproducibility, this objective can be efficiently implemented in PyTorch using the following snippet:

\begin{tcolorbox}[codebox]
\begin{lstlisting}[style=tango]
import torch

def nesterov_general_lower_bounded_fun(x, p_method: int = 2, mu: float = 1e-3):
    """
    Args:
        x: Input tensor of shape (d,)
        p_method: The order of the optimization method (p). 
                  e.g., Newton is p=2, Tensor is p>=3.
        mu: optional l2-reg for stability (not used by Nesterov).
    """
    power = p_method + 1
    diffs = torch.abs(x[:-1] - x[1:])
    chain_term = torch.sum(diffs**power) / power
    last_term = (torch.abs(x[-1])**power) / power
    linear_term = -x[0]
    l2_reg = (mu / 2) * torch.sum(x**2)
    return chain_term + last_term + linear_term + l2_reg
\end{lstlisting}
\end{tcolorbox}

We evaluate the adaptive scheme in two distinct settings. First, we consider the exact (full-space) setting, recovering exact AICN by setting the weak learner family to the full space $\mathcal{F} = \mathcal{H} = \mathbb{R}^d$. Second, we consider Newton Coordinate Descent (as introduced in \cref{sec:NewtonCoordinateDescent}) by choosing the weak learner family to be the set of scaled standard basis vectors, $\mathcal{F}_{\mathrm{CD}} = \{c e_i : c \in \mathbb{R}, 1 \leq i \leq d\}$. 

For both settings, we compare three specific optimizer configurations:
\begin{enumerate}
    \item Non-adaptive AICN with an optimally tuned regularity constant ($C_{\mathrm{best}}$), representing the smallest empirical constant that ensures a monotonic decrease in loss.
    \item Non-adaptive AICN with a conservative constant choice ($C_{\mathrm{conservative}}$).
    \item Our proposed AdaAICN scheme initialized with the conservative guess ($C_{\mathrm{conservative}}$).
\end{enumerate}
For the exact full-space experiments, we set $C_{\mathrm{best}}=81$ and $C_{\mathrm{conservative}}=1000$. For the coordinate descent experiments, we set $C_{\mathrm{best}}=16$ and $C_{\mathrm{conservative}}=100$. We set the problem dimension to $d=100$ for the exact setting and $d=20$ for the coordinate descent setting. All methods are initialized from the same random starting point. We set the regularization parameter to $\mu=10^{-3}$ and learning rate $\eta=1.0$.

\begin{figure}[htbp]
\vskip 0.3in
    \centering
    \begin{subfigure}[b]{0.48\textwidth}
        \centering
        \includegraphics[width=\textwidth]{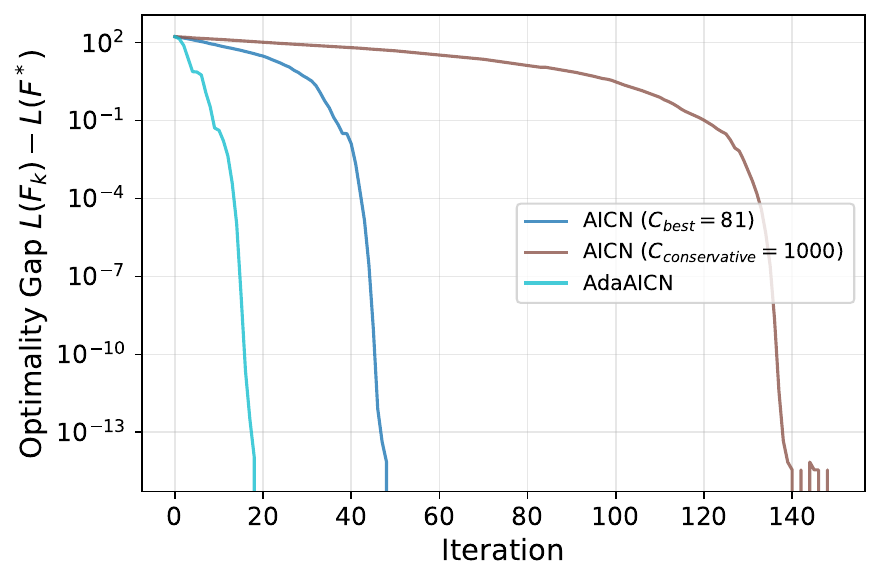} 
    \end{subfigure}
    \hfill 
    \begin{subfigure}[b]{0.48\textwidth}
        \centering
        \includegraphics[width=\textwidth]{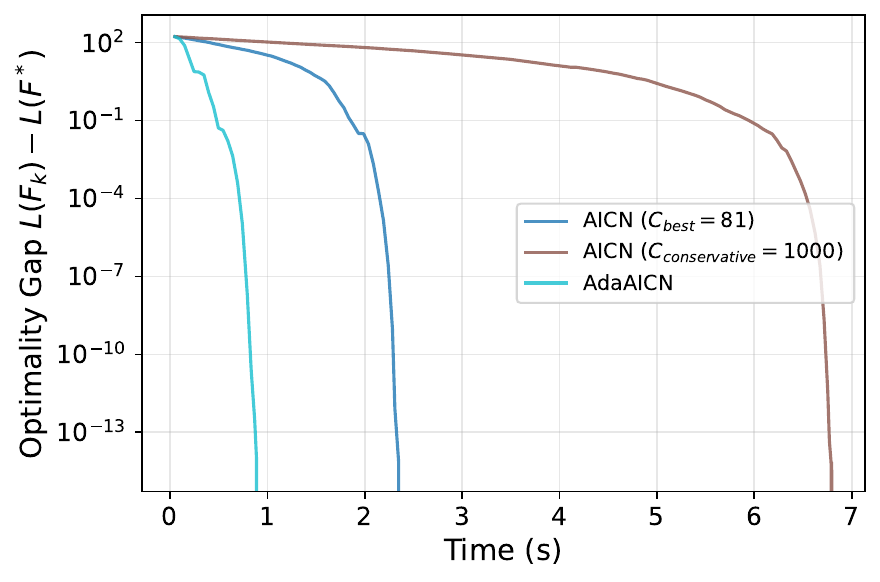} 
    \end{subfigure}
    
    \caption{Performance of exact AdaAICN versus non-adaptive AICN on Nesterov's second-order lower bound function ($d=100$). \textbf{Left:} Loss against the number of optimization iterations. \textbf{Right:} Loss against wall-clock time (seconds).}
    \label{fig:AdaAICN}
\vskip 0.3in
\end{figure}

\begin{figure}[htbp]
\vskip 0.3in
    \centering
    \begin{subfigure}[b]{0.48\textwidth}
        \centering
        \includegraphics[width=\textwidth]{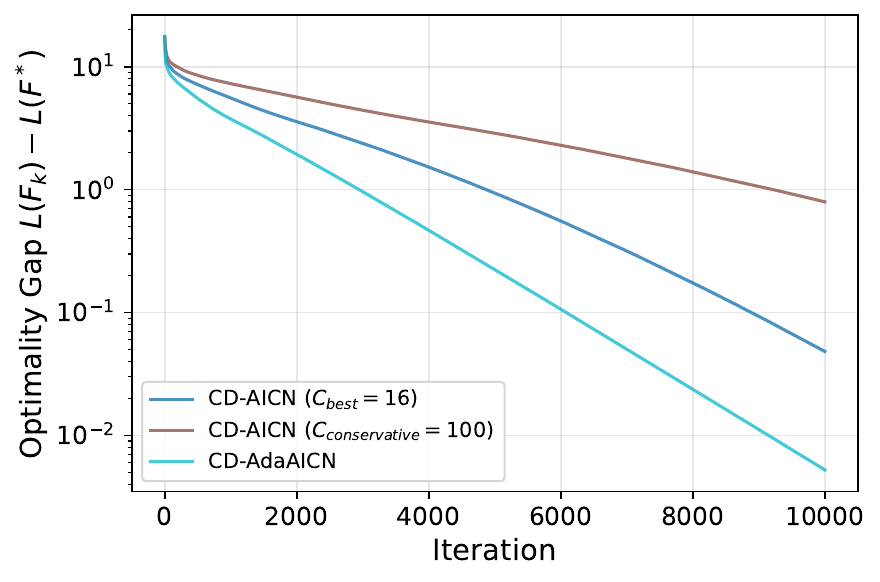} 
    \end{subfigure}
    \hfill 
    \begin{subfigure}[b]{0.48\textwidth}
        \centering
        \includegraphics[width=\textwidth]{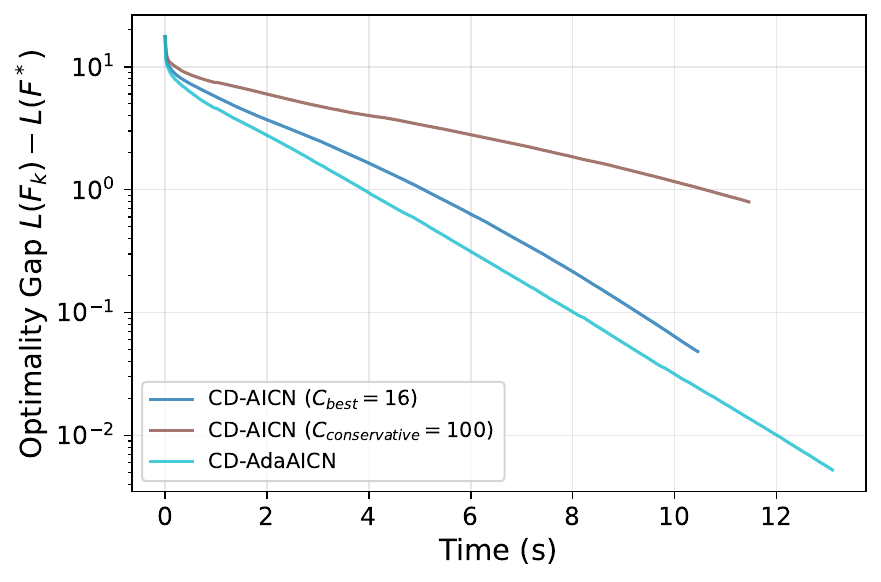} 
    \end{subfigure}
    
    \caption{Performance of Newton coordinate descent AdaAICN versus non-adaptive AICN on Nesterov's second-order lower bound function ($d=20$). \textbf{Left:} Loss against the number of optimization iterations. \textbf{Right:} Loss against wall-clock time (seconds).}
    \label{fig:CDAdaAICN}
\vskip 0.3in
\end{figure}

The results, presented in \cref{fig:AdaAICN} and \cref{fig:CDAdaAICN}, demonstrate that the adaptive variant significantly accelerates convergence across both settings, improving both the iteration complexity and the total wall-clock time. In the exact setting, AdaAICN achieves a $2\times$ reduction in wall-clock time and in the number of iterations required to reach the minimum value, rapidly improving the performance of the statically tuned $C_{\mathrm{best}}$ model without requiring costly hyperparameter sweeps. In the weak learner coordinate descent setting (\cref{fig:CDAdaAICN}), the difference is significant but smaller compared to the exact setting.

\newpage
\section{Supplementary Tree Boosting Experiments on Real-World Data}\label{app:ExperimentsGRNComparison}

In this appendix, we compare the training loss trajectories of Newton boosting methods and study the behavior of the weak learner alignment metrics used by weak-GRN \cite{2026GradientRegularizedNewtonBoostingTreeswithGlobalConvergence}, weak-AICN, and vanilla first-order gradient boosting. This section parallels the analysis of \cite{2026GradientRegularizedNewtonBoostingTreeswithGlobalConvergence}, using the same losses and datasets as presented there. Our code builds upon the JAX \cite{2018jaxGithub} boosting implementation released alongside weak-GRN, adding the weak-AICN cubic regularization as an optional parameter within the Newton boosting class.

\subsection{Charbonnier Loss: Non-convergence of Vanilla Newton Boosting}

A classical example of the divergence of Newton's method is the Charbonnier loss function, $l(\widehat{y}, y) = \sqrt{1+(y-\widehat{y})^2}-1$. For this experiment, we use the UCI Wine Quality dataset \cite{UCIWineQuality} and fit decision trees of depth $J=4$. We compare weak-AICN to weak-GRN \cite{2026GradientRegularizedNewtonBoostingTreeswithGlobalConvergence}, vanilla Newton boosting, and vanilla first-order gradient boosting across two learning rates: $\eta=1.0$ and $\eta=0.1$. We apply a static $\ell_2$ regularization of $\lambda_{\mathrm{base}}=0.1$. 

As discussed in \cref{sec:AICNWithWeakLearners}, weak-AICN relies on the $C_{\mathrm{est}}$ parameter determining the affine-invariant cubic regularization used by the method. Weak-GRN on the other hand, uses a different mechanism to globalize Newton's method. It instead applies extra adaptive gradient-based $\ell_2$ regularization equal to $\lambda_k = \sqrt{M\|g_k\|}$ at each iteration $k$, where $g_k$ is the gradient and $M$ is the Lipschitz constant of the Hessian operator. For these experiments, we select the smallest values of $C_{\mathrm{est}}$ and $M$ that successfully prevent divergence for both algorithms, resulting in $C_{\mathrm{est}} = 10$ and $M=0.1$.

\begin{figure}[H]
    \centering
    \includegraphics[width=1.0\linewidth]{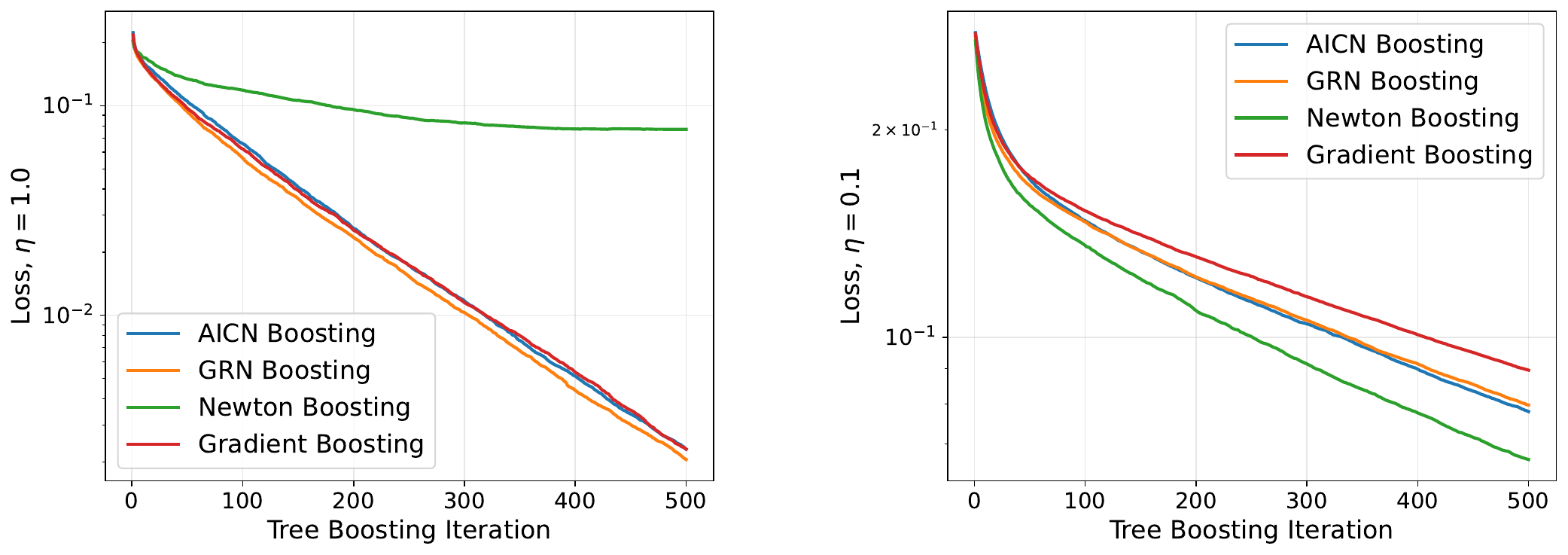}
    \caption{Training loss trajectories on the UCI Wine Quality dataset using the Charbonnier loss. We compare weak-AICN, weak-GRN, vanilla Newton boosting, and first-order gradient boosting at two learning rates: $\eta=1.0$ (left) and $\eta=0.1$ (right). Vanilla Newton boosting fails to converge without sufficient dampening ($\eta=1.0$), while weak-AICN and weak-GRN converge robustly across both settings.}
    \label{fig:charbonnier-aicn-grn-train}
\end{figure}

The training loss trajectories are presented in \cref{fig:charbonnier-aicn-grn-train}. Consistent with the findings in \cite{2026GradientRegularizedNewtonBoostingTreeswithGlobalConvergence}, we observe that vanilla Newton boosting fails to converge in the undamped setting ($\eta=1.0$), while weak-GRN, weak-AICN, and first-order boosting all converge robustly. Under this fast learning rate, the three convergent methods perform comparably. When the learning rate is reduced to $\eta=0.1$, vanilla Newton boosting successfully converges. For consistency, we maintain the same $C_{\mathrm{est}}$ and $M$ parameters across both learning rate settings. Because weak-GRN dynamically adds $\ell_2$ regularization and weak-AICN dynamically increases dampening, both methods converge slightly slower than the vanilla Newton update in the $\eta=0.1$ regime. Notably, if the parameters $C_{\mathrm{est}}$ and $M$ were set to $0$, both weak-GRN and weak-AICN would reduce to vanilla Newton boosting. First-order boosting naturally lags behind the second-order methods, as it fails to leverage curvature information for the Charbonnier loss.

\subsection{Binary Cross Entropy: Effects of Tree Depth and Weak Learner Edge}

In this section, we compare the weak learner alignment metrics that govern the convergence of weak-AICN and weak-GRN. These metrics are the Hessian-induced cosine angle $\Theta_k$ (used by weak-AICN) and the weak gradient edge $\gamma_k$ (used by weak-GRN \cite{2026GradientRegularizedNewtonBoostingTreeswithGlobalConvergence}), defined below for the reader's convenience.

\begin{definition}[Weak Gradient Edge \cite{2026GradientRegularizedNewtonBoostingTreeswithGlobalConvergence}]\label{defWeakGradientEdgeInAICNAppendix}
    Let $f_{k+1}^{\mathrm{wGRN}}$ be the weak-GRN update, and let $\lambda_k = \lambda_{\mathrm{base}}+\sqrt{M\|g_k\|}$ be the effective weak-GRN regularization. The \textbf{implied weak gradient} is defined as
    \begin{align}\label{eqWeakGradientInAICNAppendix}
        g_k^w : = - (H_k + \lambda_k I)f_{k+1}^{\mathrm{wGRN}}.
    \end{align}
    We say that the iterates defined by $f_{k+1}^{\mathrm{wGRN}}$ have \textbf{weak gradient edge} $\gamma = \inf_k \gamma_k$ if there exist real numbers $0<\gamma_k \leq 1$ satisfying 
    \begin{align}\label{eqWeakGradientEdgeInAICNAppendix}
        \|g_k^w - g_k\|^2 \leq (1-\gamma_k^2) \|g_k\|^2.
    \end{align}
\end{definition}

We also include comparisons against first-order gradient boosting \cite{2011GeneralizedBoostingAlgorithmsForConvexOptimization}, whose convergence relies on Euclidean cosine angles, denoted here as $\Theta_k^E$. We define this cosine angle analogously to \cref{def:CosineAngle}, but evaluated using the standard Euclidean (Hilbert) norm rather than the Hessian-induced norm, and using first-order trees rather than second-order trees.

Following the experimental setup of \cite{2026GradientRegularizedNewtonBoostingTreeswithGlobalConvergence}, we minimize the binary cross entropy loss on a $10,000$-sample subset of the HIGGS classification dataset \cite{UCIHiggs}. Because the binary cross entropy loss is well-behaved (specifically, Hessian-dominated per \cref{defHessianDominatedLoss}), aggressive regularization is unnecessary; we therefore fix a small baseline penalty of $\lambda_{\mathrm{base}} = 1\times 10^{-4}$. Throughout these experiments, we set the learning rate to $\eta=0.1$, the tree depth to $J=4$, and the global regularization constants to $C_{\mathrm{est}}=1.0$ and $M=0.01$. As in the original weak-GRN experiments \cite{2026GradientRegularizedNewtonBoostingTreeswithGlobalConvergence}, the static $\ell_2$ term regularizes
only the quadratic step: each sample Hessian is replaced by
$h_{(i)}+\lambda_{\mathrm{base}}I$, while the sample gradients remain unchanged. This differs from adding $\ell_2$-regularization to the loss itself, which modifies both the sample gradients and Hessians.

\begin{figure}[p]
    \centering
    \begin{subfigure}[b]{0.46\textwidth}
        \centering
        \includegraphics[width=\textwidth]{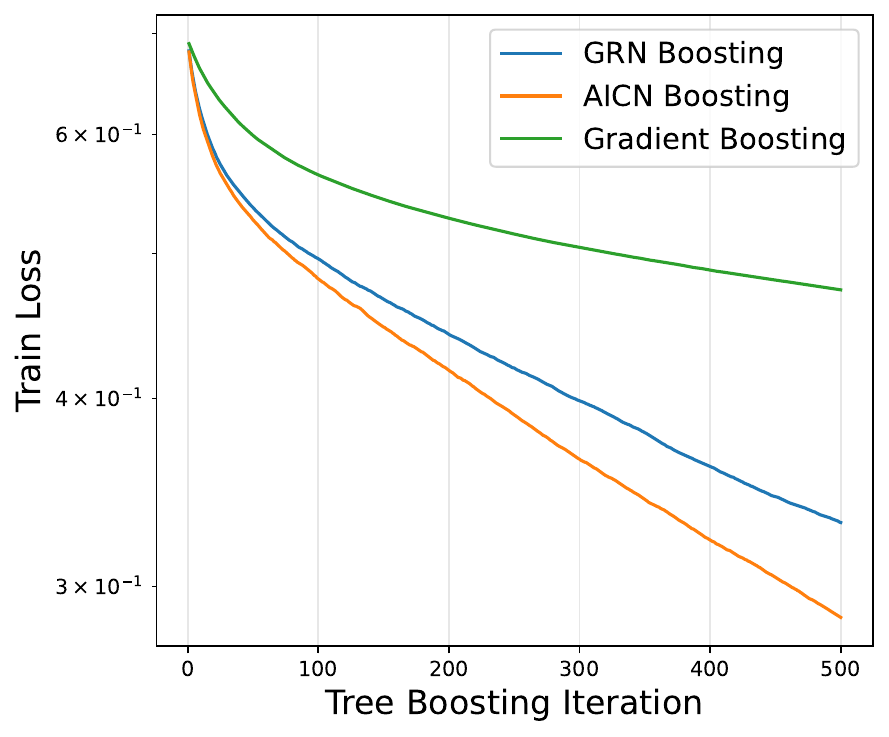}
    \end{subfigure}
    \hfill
    \begin{subfigure}[b]{0.46\textwidth}
        \centering
        \includegraphics[width=\textwidth]{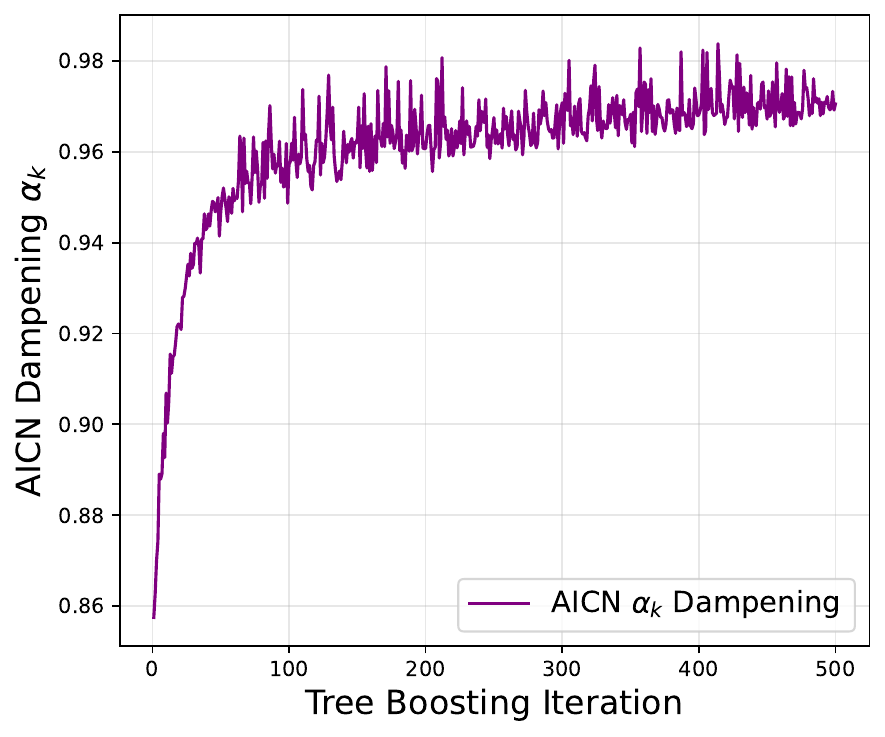}
    \end{subfigure}
    \par\medskip
    \begin{subfigure}[b]{0.46\textwidth}
        \centering
        \includegraphics[width=\textwidth]{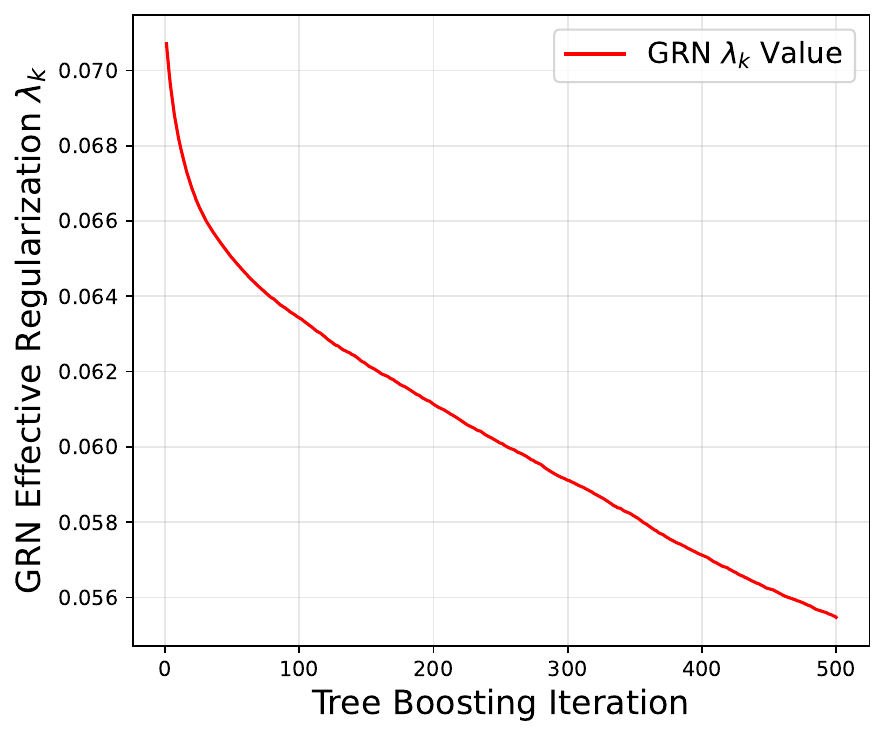}
    \end{subfigure}
    \hfill
    \begin{subfigure}[b]{0.46\textwidth}
        \centering
        \includegraphics[width=\textwidth]{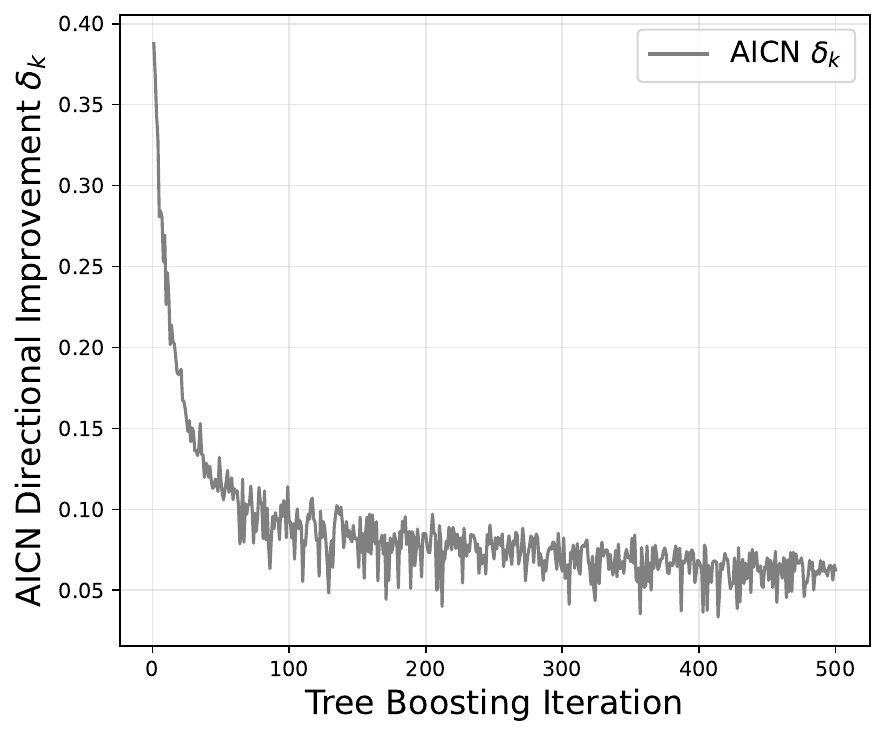}
    \end{subfigure}

    \caption{Convergence metrics on the HIGGS dataset using Binary Cross Entropy ($\eta=0.1, J=4$). \textbf{Top left:} Training loss convergence for weak-AICN, weak-GRN, and first-order gradient boosting. \textbf{Top right:} Evolution of the weak-AICN dampening factor $\alpha_k$, which approaches $1.0$ as optimization progresses. \textbf{Bottom left:} Evolution of the effective adaptive $\ell_2$ regularization $\lambda_k$ used by weak-GRN. \textbf{Bottom right:} Directional improvement $\delta_k$ of the weak Newton update used by weak-AICN.}
    \label{fig:higgs_loss}

    \par\vspace{3em}

    \centering
    \begin{subfigure}[b]{0.32\textwidth}
        \centering
        \includegraphics[width=\textwidth]{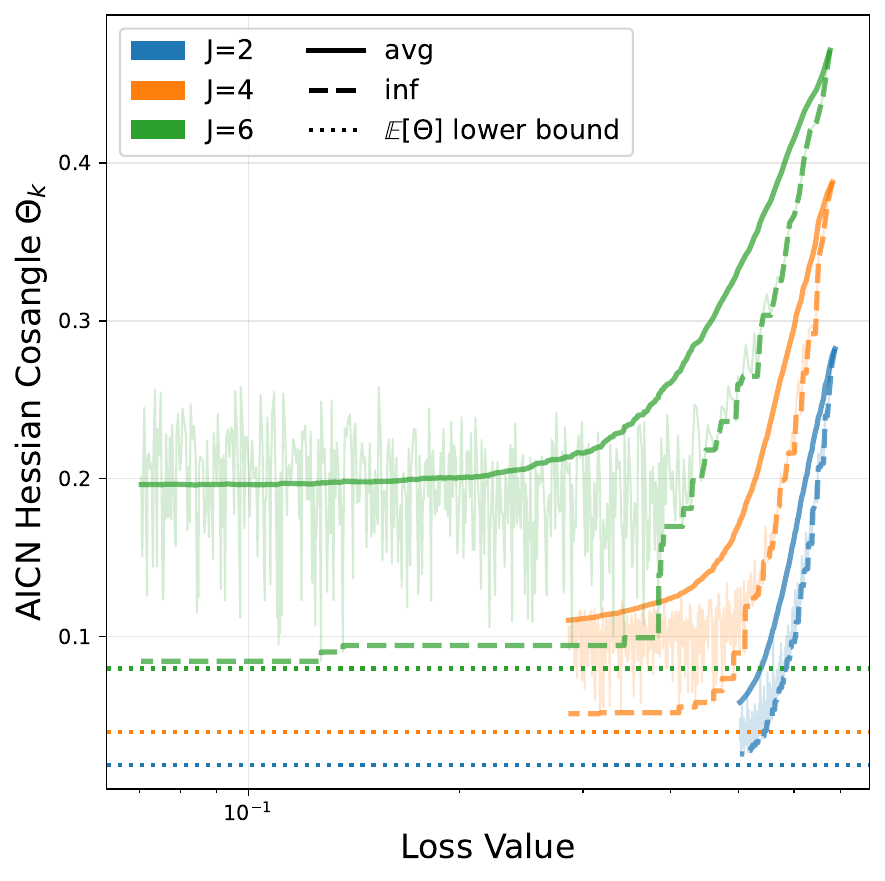}
    \end{subfigure}
    \hfill 
    \begin{subfigure}[b]{0.32\textwidth}
        \centering
        \includegraphics[width=\textwidth]{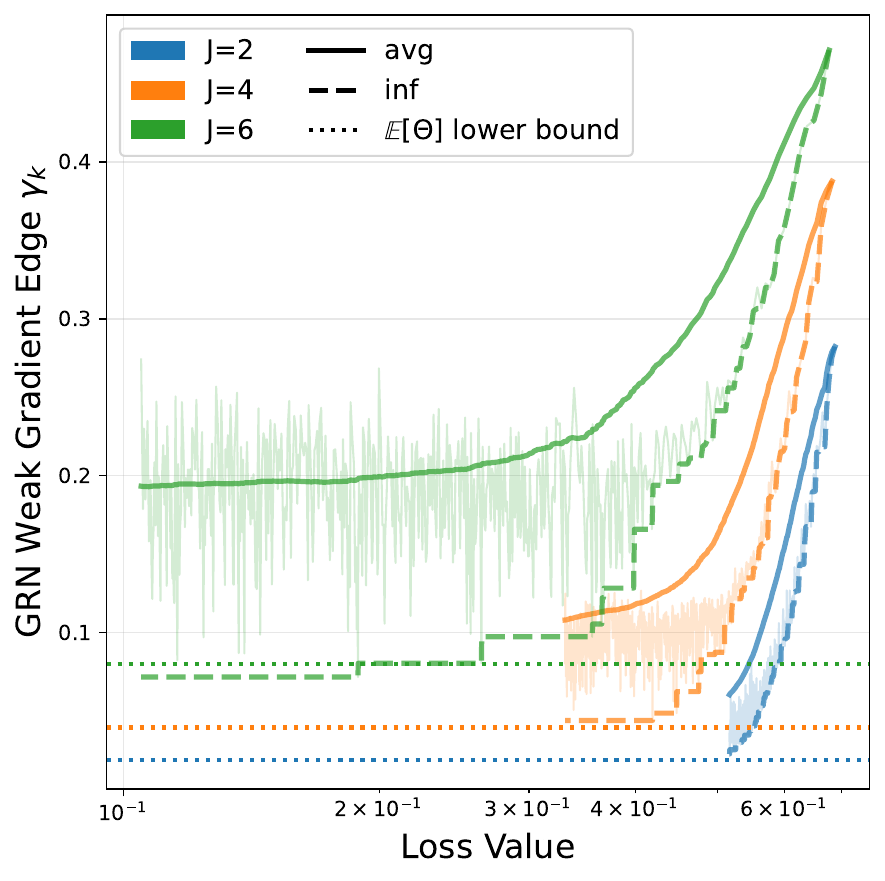} 
    \end{subfigure}
    \hfill 
    \begin{subfigure}[b]{0.32\textwidth}
        \centering
        \includegraphics[width=\textwidth]{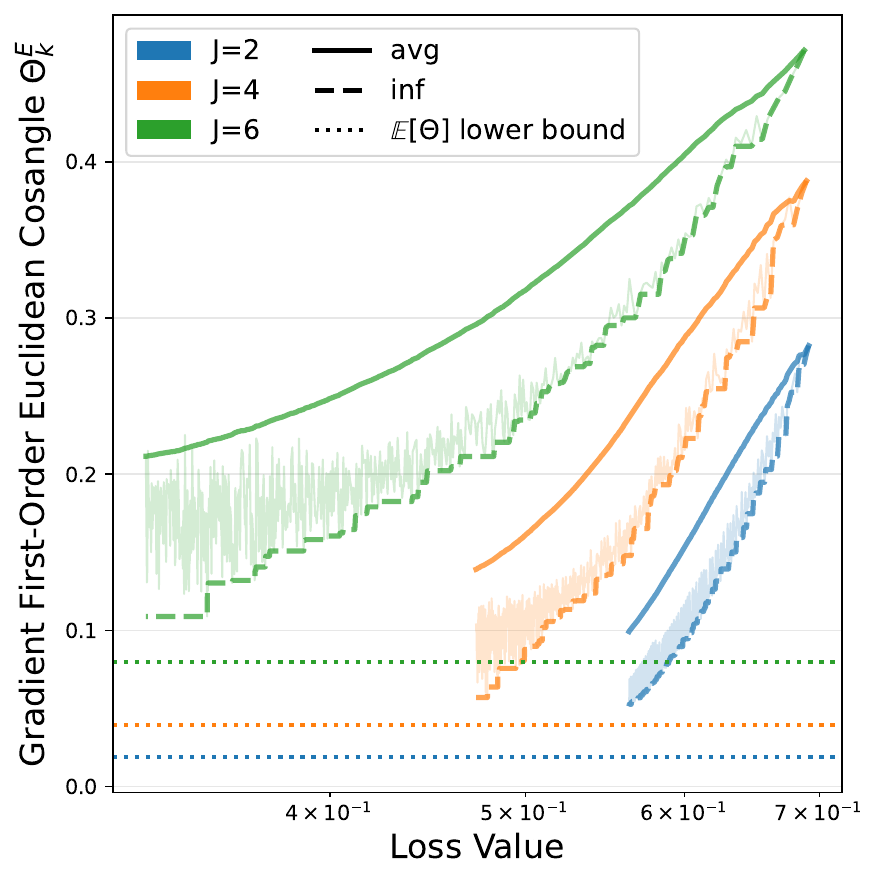} 
    \end{subfigure}
    
    \caption{Effect of tree depth ($J \in \{2, 4, 6\}$) on weak learner alignment metrics on the HIGGS dataset. \textbf{Left:} Hessian-induced cosine angle $\Theta_k$ for weak-AICN. \textbf{Middle:} Weak gradient edge $\gamma_k$ for weak-GRN. \textbf{Right:} Euclidean cosine angle $\Theta_k^E$ for first-order gradient boosting. Across all metrics, the stabilized average alignment roughly doubles with every two additional levels of tree depth. Dotted lines show the expectation lower bounds for the random tree-angle proxy in \cref{thm:tree-average-main}, included as reference values.}
    \label{fig:higgs_tree_depth}
\end{figure}

\cref{fig:higgs_loss} visualizes the training loss trajectories, the weak-AICN dampening factor $\alpha_k$ and directional improvement $\delta_k$, and the weak-GRN regularization $\lambda_k$. As optimization progresses, the weak-AICN dampening factor $\alpha_k$ gradually approaches $1.0$. Recall that $\alpha_k = \frac{2}{1+\sqrt{1+2C_{\mathrm{est}}\delta_k}}$, where $\delta_k = \delta_k(f_{k+1}^{\mathrm{wN}})$ is the directional improvement of the weak Newton iterate. As the residual error becomes harder to fit during the later stages of training, $\delta_k$ shrinks, causing $\alpha_k$ to relax towards $1.0$. This increasing difficulty to fit the residual is further illustrated in the weak learner edge comparison plots of \cref{fig:higgs_weak_learner_edge_comparisons}. As optimization progresses, there is less signal and more noise in the residuals, causing the cosine angles and weak gradient edge to decrease. These plots additionally highlight the differences between the pessimistic worst-case edges (infimums) and the running averages computed from the start of optimization. On this dataset, both the Hessian cosine angle $\Theta_k$ and the weak gradient edge $\gamma_k$ plateau at approximately $0.1$ on average. In contrast, their respective infimums are near $0.05$. This difference is critical for theoretical convergence bounds: our weak-AICN global rate relies on the path-averaged angle raised to the fourth power $\overline{\Theta}_k^4$, whereas the weak-GRN bound suffers from a twelfth-power dependence on the worst-case edge $\gamma^{12}$. Furthermore, the left plot of \cref{fig:higgs_weak_learner_edge_comparisons} shows that tracking weak learner alignment against iteration count can be visually misleading. Plotting them against the training loss (right plot) clarifies the underlying trend: evaluated at equivalent loss values, weak-AICN ($\Theta_k$) maintains slightly higher weak learner edge compared to the other methods, though all three metrics are very similar overall.

Finally, in \cref{fig:higgs_tree_depth}, we test the effect of tree depth ($J \in \{2, 4, 6\}$) on each weak learner edge, similar to \cite{2026GradientRegularizedNewtonBoostingTreeswithGlobalConvergence}. We observe a clear monotonic relationship between tree depth and weak learner alignment. For all metrics, the weak learner edge is high during the initial phase of optimization before stabilizing. The stabilized running averages roughly equal $0.2$ for $J=6$, $0.1$ for $J=4$, and $0.05$ for $J=2$, effectively doubling every two additional levels of tree depth. This dyadic scaling (after squaring $\Theta_k$) is consistent with our theoretical tree-angle proxy bounds presented in \cref{thm:tree-average-main}. The empirical angles along the actual optimization path remain above the no-signal reference values.

\clearpage
\section{Tree Weak Learners Cosine Angle Lower Bounds}
\label{app:tree-proxy-experiments}

We continue in the empirical boosting setting of \cref{sec:gradient-boosting}.
Thus \(X=\{x_1,\ldots,x_n\}\subset \mathbb R^p\) is fixed,
\[
    \mathcal H_X := \mathcal{L}^2(\widehat\nu_n)\cong \mathbb R^X,
\]
and \(\mathcal H_X\) is equipped with the empirical inner product
\(\langle\cdot,\cdot\rangle_X\). Throughout this appendix, \(g_k,H_k,Q_k\), and
\(f^N_{k+1}\) are understood in this empirical space. We work with scalar outputs; the
vector-valued case is analogous.

\subsection{Fixed-sample tree-angle geometry}
\label{app:fixed-sample-tree-geometry}

Fix \(J\geq 0\). For a depth-\(J\) axis-aligned tree \(T\), let
\(V_T\subset \mathcal H_X\) denote the subspace of functions on \(X\) that are constant
on the leaves of \(T\). Thus
\[
    \mathcal T_J(X)=\bigcup_T V_T,
\]
where \(T\) ranges over depth-\(J\) axis-aligned trees, viewed only through their induced
partitions of \(X\).

Let \(A:\mathcal H_X\to\mathcal H_X\) be self-adjoint and positive definite. Define
\[
    \langle f,h\rangle_A := \langle f,Ah\rangle_X,
    \qquad
    \|f\|_A^2 := \langle f,f\rangle_A .
\]
For each tree \(T\), let \(\Pi^A_{V_T}\) denote the \(A\)-orthogonal projection onto
\(V_T\). We define the primitive depth-\(J\) tree cosine by
\[
    \Theta_J(X;f\mid A)
    :=
    \max_{g\in\mathcal T_J(X)\setminus\{0\}}
    \frac{\langle f,g\rangle_A}{\|f\|_A\|g\|_A},
    \qquad
    f\in\mathcal H_X\setminus\{0\}.
\]
Equivalently,
\[
    \Theta_J(X;f\mid A)
    =
    \max_T
    \frac{\|\Pi^A_{V_T}f\|_A}{\|f\|_A},
\]
where \(T\) ranges over depth-\(J\) axis-aligned trees on \(X\). Hence
\[
    \Theta_J(X;f\mid A)^2
    =
    1-
    \inf_{g\in\mathcal T_J(X)}
    \frac{\|f-g\|_A^2}{\|f\|_A^2}.
\]
When \(A=I\) with respect to \(\langle\cdot,\cdot\rangle_X\), we abbreviate
\[
    \Theta_J(X;f) := \Theta_J(X;f\mid I).
\]

Along the realized boosting iterates, the depth-\(J\) weak Newton tree cosine from
Sections~2--3 is the special case
\[
    \Theta^{(k)}_J(X,y,F_0)
    :=
    \Theta_J(X;f^N_{k+1}\mid H_k).
\]
When the realized iterates are fixed, we also write
\(\Theta_{k,J}:=\Theta^{(k)}_J(X,y,F_0)\).
All deterministic statements below therefore apply to weak Newton tree cosines by setting
\(A=H_k\) and \(f=f^N_{k+1}\).

\subsection{Deterministic empirical tree geometry}

\begin{proposition}[Leaf-count obstruction and worst-case exactness]\label{prop:tree-finite-depth-exactness}
For every configuration \(X\) of \(n\) distinct points, exact representation of all empirical
functions \(f\in\mathcal H_X\) by depth-\(J\) axis-aligned trees is impossible when \(2^J<n\).
Conversely, exact representation of all empirical functions is possible for every such \(X\)
once \(J\ge n-1\).
Equivalently, for every self-adjoint positive definite \(A\) on \(\mathcal H_X\),
\[
\Theta_J(X;f\mid A)=1
\qquad
\text{for every nonzero }f\in\mathcal H_X
\]
whenever \(J\ge n-1\).
\end{proposition}

\begin{proof}
A depth-\(J\) binary tree has at most \(2^J\) leaves. If \(2^J<n\), then no depth-\(J\)
tree can separate \(n\) distinct sample points into singleton leaves, so it cannot represent
all empirical functions on \(X\).

For the converse, argue by induction on \(n\). The case \(n=1\) is trivial. If \(n\ge2\),
choose two sample points and a coordinate on which they differ. A threshold between two
consecutive values in that coordinate gives a nontrivial split into two subsets of sizes
\(n_1,n_2\ge1\) with \(n_1+n_2=n\). By induction, the two children can be separated
with depths at most \(n_1-1\) and \(n_2-1\), so the whole sample can be separated with
depth at most
\[
1+\max\{n_1-1,n_2-1\}\le n-1.
\]
Assigning arbitrary leaf values then realizes every empirical function. The equivalent
statement for \(\Theta_J(X;f\mid A)\) is immediate from the definition.
\end{proof}

\begin{example}[Parity obstruction]\label{ex:tree-parity-obstruction}
For the cube configuration
\[
X=\{0,1\}^d,
\]
take \(J=d-1\) and define
\[
f(x):=(-1)^{x_1+\cdots+x_d}.
\]
Then
\[
\Theta_J(X;f)=0.
\]
Hence below the leaf-count regime there is, in general, no universal positive lower bound
on the intrinsic tree cosine for empirical axis-aligned trees.
\end{example}

\begin{proof}
Every depth-\((d-1)\) leaf fixes at most \(d-1\) coordinates, so at least one coordinate
remains free on each leaf. The parity function sums to zero over any such leaf, and
therefore has zero inner product with every function that is constant on that leaf. Since
a depth-\((d-1)\) tree is constant on each of its leaves, it follows that \(f\) is orthogonal
to every such tree.
\end{proof}

This example should be read as a worst-case obstruction, not as a model of typical boosting
iterations. In particular, for squared loss \(H_k=I\), so the construction directly rules out
assumption-free deterministic lower bounds for the iteration-wise quantity \(\Theta_{k,J}\)
below the leaf-count regime. For general Hessian metrics, it still shows why worst-case
tree-angle analysis is too pessimistic and motivates the probabilistic proxy studied next.

\subsection{Random model and dyadic lower bounds}

The deterministic obstruction above shows that worst-case tree-angle bounds are too pessimistic.
We therefore introduce a tractable random proxy for tree alignment. The proxy is not meant to
assert that an actual normalized Newton direction is uniformly distributed on the empirical unit
sphere. Rather, it is an uncoupled null model: it retains the sample and metric geometry, but
replaces the Newton direction by a generic isotropic direction in that metric.

Let
\[
X_1,\dots,X_n \stackrel{\mathrm{iid}}{\sim}\mathrm{Unif}([0,1]^p),
\]
and write \(X=\{X_1,\dots,X_n\}\), which is almost surely a set of \(n\) distinct points.
A \emph{metric model} \(\mathsf Q\) consists of this sample law together with an arbitrary
conditional law, given \(X\), for a self-adjoint positive definite operator
\[
    A:\mathcal H_X\to\mathcal H_X.
\]
For a realized pair \((X,A)\), define the \(A\)-unit sphere
\[
S_A(\mathcal H_X):=\{f\in\mathcal H_X:\|f\|_A=1\},
\]
and let
\[
U_A\mid (X,A)\sim \mathrm{Unif}(S_A(\mathcal H_X)),
\]
where uniformity is understood with respect to the inner product \(\langle\cdot,\cdot\rangle_A\).
We define the random metric proxy depth-\(J\) tree cosine by
\[
\widehat\Theta^{\mathrm{prox},\mathsf Q}_{n,p,J}
:=
\Theta_J(X;U_A\mid A).
\]
The identity-metric static proxy is the special case where \(A=I\) deterministically; in that case
we write simply
\[
\widehat\Theta^{\mathrm{prox}}_{n,p,J}
:=
\widehat\Theta^{\mathrm{prox},\mathsf Q_I}_{n,p,J}
=
\Theta_J(X;U_I\mid I).
\]

The intended comparison is with randomized iteration-level quantities generated by a chosen
probabilistic model \(\mathsf M\) for the data, labels, initialization, and fixed loss. Such a model
induces coupled quantities
\[
\widehat\Theta^{(k;\mathsf M)}_{n,p,J}
:=
\Theta_J(X;f^N_{k+1}\mid H_k),
\]
where \(H_k\) and \(f^N_{k+1}\) are produced by the same optimization state. The corresponding
uncoupled metric-matched proxy is
\[
\widehat\Theta^{\mathrm{prox},(k;\mathsf M)}_{n,p,J}
:=
\Theta_J(X;U_{H_k}\mid H_k),
\qquad
U_{H_k}\mid(X,H_k)\sim \operatorname{Unif}(S_{H_k}(\mathcal H_X)).
\]
The role of the proxy is to replace the coupled Newton direction by a generic direction in the same
realized Hessian geometry. The first-iteration comparison motivating this interpretation is recorded
next.

\begin{remark}[Proxy comparisons and asymptotic beta domination]\label{rem:first-iteration-proxy-comparison}
At a fixed depth \(J\), the basic model-dependent proxy comparison is the first-iteration
expectation inequality
\[
\mathbb E\!\left[
\bigl(\widehat\Theta^{\mathrm{prox},(0;\mathsf M)}_{n,p,J}\bigr)^2
\right]
\le
\mathbb E\!\left[
\bigl(\widehat\Theta^{(0;\mathsf M)}_{n,p,J}\bigr)^2
\right].
\]
This should be read as a structured-signal hypothesis: it is expected when the first Newton direction
contains feature-aligned, tree-readable signal, while unstructured null models should give approximate
equality and parity-type targets need not satisfy it. This first-iteration comparison is conjectural,
and \cref{app:tree-angle-experiments} provides empirical evidence for precisely this interpretation
of the proxy.

\vspace{10pt}
A stronger, genuinely dynamical conjecture is that, for non-adversarial boosting models
\(\mathsf M\), the actual coupled weak Newton tree angle should asymptotically dominate the
proxy beta law itself. More precisely, if \(2^J<n\) and if the limiting coupled tree angle
\(\widehat\Theta^{(\infty;\mathsf M)}_{n,p,J}\) exists in distribution, one may conjecture that
\[
\bigl(\widehat\Theta^{(\infty;\mathsf M)}_{n,p,J}\bigr)^2
\succeq_{\mathrm{st}}
Z_J,
\qquad
Z_J\sim
\operatorname{Beta}\!\left(2^{J-1},\frac{n-2^J}{2}\right),
\]
where \(\succeq_{\mathrm{st}}\) denotes stochastic domination. Equivalently,
\[
\mathbb P\!\left(
\bigl(\widehat\Theta^{(\infty;\mathsf M)}_{n,p,J}\bigr)^2\ge t
\right)
\ge
\mathbb P(Z_J\ge t),
\qquad t\in[0,1].
\]
This distributional statement is substantially stronger than anything proved or directly verified
experimentally here. The consequence in expectation would be
\[
\mathbb E\!\left[
\bigl(\widehat\Theta^{(\infty;\mathsf M)}_{n,p,J}\bigr)^2
\right]
\ge
\frac{2^J}{n}.
\]
More cautiously, at finite time this suggests the path-averaged expected-value heuristic
\[
\liminf_{K\to\infty}
\frac1K\sum_{k=0}^{K-1}
\mathbb E\!\left[
\bigl(\widehat\Theta^{(k;\mathsf M)}_{n,p,J}\bigr)^2
\right]
\ge
\frac{\min(2^J,n)}{n}.
\]
The experiments in \cref{app:ExperimentsGRNComparison} should be interpreted as evidence for
this weaker along-trajectory scaling: the observed dependence of realized weak learner angles on
tree depth follows the dyadic behavior predicted by the proxy theorem. Thus
\cref{app:tree-angle-experiments} supports the first-iteration proxy comparison, while
\cref{app:ExperimentsGRNComparison} supports the weaker path-averaged expected-value
consequence. Neither appendix should be read as establishing the full stochastic domination
conjecture, which remains a stronger structural hypothesis about the late-time optimization dynamics. A rigorous treatment of this conjecture would require additional control of the limiting weak-Newton residual geometry; we leave this as an interesting direction for future work. 
\end{remark}

\begin{remark}[Distributional viewpoint]\label{rem:distributional-proxy-viewpoint}
Let \(\sigma_{X,A}\) denote the uniform probability measure on \(S_A(\mathcal H_X)\). The metric
proxy \(\widehat\Theta^{\mathrm{prox},\mathsf Q}_{n,p,J}\) is obtained by first sampling \((X,A)\)
and then sampling a generic \(A\)-unit direction. Equivalently, its distribution function can be
written as
\[
\mathbb P(\widehat\Theta^{\mathrm{prox},\mathsf Q}_{n,p,J}\le t)
=
\int
\left(
\int_{S_A(\mathcal H_X)}
\mathbf 1_{\{\Theta_J(X;u\mid A)\le t\}}\,d\sigma_{X,A}(u)
\right)\,d\mathsf Q(X,A),
\qquad 0\le t\le 1.
\]
Thus the proxy distribution averages the fixed-sample tree-angle observable first over generic
directions in the realized metric and then over random sample--metric pairs.

\vspace{7pt}
Similarly, if \(\mathsf M\) is a probabilistic model generating the iterates, then the coupled
iteration-level tree cosine has distribution function
\[
\mathbb P(\widehat\Theta^{(k;\mathsf M)}_{n,p,J}\le t)
=
\int
\mathbf 1_{\{\Theta_J(X;f^N_{k+1}(X,y,F_0)\mid H_k(X,y,F_0))\le t\}}\,
d\mathsf M(X,y,F_0).
\]
The distinction is that \(\widehat\Theta^{\mathrm{prox},(k;\mathsf M)}_{n,p,J}\) preserves the
realized Hessian geometry but randomizes the direction, while
\(\widehat\Theta^{(k;\mathsf M)}_{n,p,J}\) uses the Newton direction coupled to that geometry.
\end{remark}

We now prove the main stochastic domination and expected value bounds in the form of the subsequent two results; the combination of \cref{lem:tree-dyadic-lower-bound} and \cref{thm:tree-dyadic-lower-bound} is what entails \cref{thm:tree-average-main}.

\begin{lemma}[Dyadic stochastic lower bound for the random tree cosine]
\label{lem:tree-dyadic-lower-bound}
Let \(n\ge2\), \(p\ge1\), and \(J\ge0\), and let \(\mathsf Q\) be any metric model as above.
If \(2^J<n\), then there exists a random variable
\[
Z_J\sim \operatorname{Beta}\!\left(2^{J-1},\frac{n-2^J}{2}\right),
\]
where \(\operatorname{Beta}(a,b)\) denotes the distribution on \([0,1]\) with density
proportional to \(z^{a-1}(1-z)^{b-1}\), such that
\[
\bigl(\widehat\Theta^{\mathrm{prox},\mathsf Q}_{n,p,J}\bigr)^2\ge Z_J
\qquad\text{almost surely.}
\]
If \(2^J\ge n\), then
\[
\widehat\Theta^{\mathrm{prox},\mathsf Q}_{n,p,J}=1
\qquad\text{almost surely.}
\]
\end{lemma}

\begin{proof}
Because the sample law is continuous, the first-coordinate values are pairwise distinct almost surely.
Condition on such a realization of \((X,A)\) and relabel the points so that
\[
x_1^{(1)}<x_2^{(1)}<\cdots<x_n^{(1)}.
\]
Using only the first coordinate, recursively bisect the ordered list into consecutive blocks as evenly
as possible. After depth \(J\), this yields a depth-\(J\) axis-aligned tree, and let
\[
V_J^{\mathrm{dyad}}(X)\subseteq \mathcal H_X
\]
denote its associated leaf-constant subspace. By construction,
\[
\dim V_J^{\mathrm{dyad}}(X)=\min(2^J,n).
\]

First suppose \(2^J<n\). Since
\(\widehat\Theta^{\mathrm{prox},\mathsf Q}_{n,p,J}=\Theta_J(X;U_A\mid A)\) is defined using the best
depth-\(J\) tree function in the \(A\)-metric, we have
\[
\widehat\Theta^{\mathrm{prox},\mathsf Q}_{n,p,J}
\ge
\max_{g\in V_J^{\mathrm{dyad}}(X)\setminus\{0\}}
\frac{\langle U_A,g\rangle_A}{\|g\|_A}
=
\|\Pi^A_{V_J^{\mathrm{dyad}}(X)}U_A\|_A.
\]
Hence
\[
\bigl(\widehat\Theta^{\mathrm{prox},\mathsf Q}_{n,p,J}\bigr)^2
\ge
\|\Pi^A_{V_J^{\mathrm{dyad}}(X)}U_A\|_A^2
\qquad\text{almost surely.}
\]

Choose any orthonormal basis of \((\mathcal H_X,\langle\cdot,\cdot\rangle_A)\) adapted to the
decomposition
\[
V_J^{\mathrm{dyad}}(X)\oplus_A \bigl(V_J^{\mathrm{dyad}}(X)\bigr)^{\perp_A}.
\]
In this basis, the \(A\)-unit-sphere law is the standard spherical law on \(S^{n-1}\), so
\[
\|\Pi^A_{V_J^{\mathrm{dyad}}(X)}U_A\|_A^2
\stackrel d=
\frac{G_1^2+\cdots+G_{2^J}^2}{G_1^2+\cdots+G_n^2},
\]
where \(G_1,\dots,G_n\stackrel{\mathrm{iid}}{\sim}N(0,1)\). Define
\[
Z_J:=\|\Pi^A_{V_J^{\mathrm{dyad}}(X)}U_A\|_A^2.
\]
Then, conditionally on \((X,A)\), and hence also unconditionally,
\[
Z_J\sim \operatorname{Beta}\!\left(2^{J-1},\frac{n-2^J}{2}\right),
\]
which proves the stochastic domination.

If \(2^J\ge n\), then the same first-coordinate construction isolates every sample point into a
singleton leaf. Hence the resulting depth-\(J\) tree class equals \(\mathcal H_X\), and therefore
\[
\widehat\Theta^{\mathrm{prox},\mathsf Q}_{n,p,J}=1
\qquad\text{almost surely.}
\]
\end{proof}

\begin{theorem}[Dyadic expectation lower bounds]
\label{thm:tree-dyadic-lower-bound}
For every metric model \(\mathsf Q\), and for all \(n\ge2\), \(p\ge1\), and \(J\ge0\),
\[
\mathbb E\!\left[\bigl(\widehat\Theta^{\mathrm{prox},\mathsf Q}_{n,p,J}\bigr)^2\right]
\ge
\frac{\min(2^J,n)}{n}.
\]
Moreover, if \(2^J<n\), then
\[
\mathbb E[\widehat\Theta^{\mathrm{prox},\mathsf Q}_{n,p,J}]
\ge
\frac{
\Gamma\!\left(\frac{2^J+1}{2}\right)
\Gamma\!\left(\frac n2\right)
}{
\Gamma\!\left(2^{J-1}\right)
\Gamma\!\left(\frac{n+1}{2}\right)
}.
\]
In particular, if \(2^J<n\) and \(J,n\to\infty\), then
\[
\frac{
\Gamma\!\left(\frac{2^J+1}{2}\right)
\Gamma\!\left(\frac n2\right)
}{
\Gamma\!\left(2^{J-1}\right)
\Gamma\!\left(\frac{n+1}{2}\right)
}
=
\left(1+O\!\left(2^{-J}+\frac1n\right)\right)
\frac{2^{J/2}}{\sqrt n}.
\]
If \(2^J\ge n\), then
\[
\widehat\Theta^{\mathrm{prox},\mathsf Q}_{n,p,J}=1
\qquad\text{almost surely,}
\]
and consequently
\[
\mathbb E[\widehat\Theta^{\mathrm{prox},\mathsf Q}_{n,p,J}]
=
\mathbb E\!\left[\bigl(\widehat\Theta^{\mathrm{prox},\mathsf Q}_{n,p,J}\bigr)^2\right]
=
1.
\]
\end{theorem}

\begin{proof}
If \(2^J<n\), \cref{lem:tree-dyadic-lower-bound} yields
\[
\bigl(\widehat\Theta^{\mathrm{prox},\mathsf Q}_{n,p,J}\bigr)^2\ge Z_J
\qquad\text{almost surely,}
\]
where
\[
Z_J\sim \operatorname{Beta}\!\left(2^{J-1},\frac{n-2^J}{2}\right).
\]
Therefore
\[
\mathbb E\!\left[\bigl(\widehat\Theta^{\mathrm{prox},\mathsf Q}_{n,p,J}\bigr)^2\right]
\ge
\mathbb E[Z_J]
=
\frac{2^{J-1}}{2^{J-1}+((n-2^J)/2)}
=
\frac{2^J}{n}
=
\frac{\min(2^J,n)}{n}.
\]
Also,
\[
\mathbb E[\widehat\Theta^{\mathrm{prox},\mathsf Q}_{n,p,J}]
=
\mathbb E\!\left[\bigl((\widehat\Theta^{\mathrm{prox},\mathsf Q}_{n,p,J})^2\bigr)^{1/2}\right]
\ge
\mathbb E[Z_J^{1/2}].
\]
For \(Z\sim\operatorname{Beta}(a,b)\),
\[
\mathbb E[Z^{1/2}]
=
\frac{\Gamma(a+\frac12)\Gamma(a+b)}
{\Gamma(a)\Gamma(a+b+\frac12)}.
\]
Applying this with
\[
a=2^{J-1},
\qquad
b=\frac{n-2^J}{2},
\]
gives
\[
\mathbb E[\widehat\Theta^{\mathrm{prox},\mathsf Q}_{n,p,J}]
\ge
\frac{
\Gamma\!\left(\frac{2^J+1}{2}\right)
\Gamma\!\left(\frac n2\right)
}{
\Gamma\!\left(2^{J-1}\right)
\Gamma\!\left(\frac{n+1}{2}\right)
}.
\]

If \(2^J\ge n\), then \cref{lem:tree-dyadic-lower-bound} gives
\(\widehat\Theta^{\mathrm{prox},\mathsf Q}_{n,p,J}=1\) almost surely. Hence both expectations equal \(1\),
which is precisely \(\min(2^J,n)/n\) for the squared-angle bound. It remains only to justify the final asymptotic statement. We use the standard
gamma-ratio estimate
\[
    \frac{\Gamma(x+a)}{\Gamma(x+b)}
    =
    x^{a-b}\left(1+O(x^{-1})\right),
    \qquad x\to\infty,
\]
for fixed \(a,b\). With \(m=2^J\), applying this first with
\(x=m/2\), \(a=1/2\), and \(b=0\), gives
\[
    \frac{\Gamma\!\left(\frac{m+1}{2}\right)}
         {\Gamma\!\left(\frac m2\right)}
    =
    \left(\frac m2\right)^{1/2}
    \left(1+O(m^{-1})\right).
\]
Applying the same estimate with \(x=n/2\), \(a=0\), and \(b=1/2\), gives
\[
    \frac{\Gamma\!\left(\frac n2\right)}
         {\Gamma\!\left(\frac{n+1}{2}\right)}
    =
    \left(\frac n2\right)^{-1/2}
    \left(1+O(n^{-1})\right).
\]
Multiplying these two estimates yields
\[
\frac{
\Gamma\!\left(\frac{m+1}{2}\right)
\Gamma\!\left(\frac n2\right)
}{
\Gamma\!\left(\frac m2\right)
\Gamma\!\left(\frac{n+1}{2}\right)
}
=
\left(1+O\!\left(m^{-1}+n^{-1}\right)\right)
\sqrt{\frac mn}.
\]
Substituting \(m=2^J\) gives the claimed
\[
    \left(1+O\!\left(2^{-J}+\frac1n\right)\right)
    \frac{2^{J/2}}{\sqrt n}.
\]
\end{proof}

\begin{remark}
The squared-angle estimate is the main quantity used in the paper, since the weak-AICN
rates in \cref{sec:AICNWithWeakLearners} depend on squared cosine factors such as \(\Theta_{k,J}^2\). \cref{thm:tree-dyadic-lower-bound}
therefore gives a lower benchmark for the uncoupled metric proxy
\(\widehat\Theta^{\mathrm{prox},\mathsf Q}_{n,p,J}\), uniformly over the metric model
\(\mathsf Q\). In particular, choosing \(\mathsf Q\) to be the marginal law of \((X,H_k)\)
induced by a probabilistic boosting model gives the corresponding benchmark for the
metric-matched first-iteration or iteration-level proxy. The additional bound for
\(\mathbb E[\widehat\Theta^{\mathrm{prox},\mathsf Q}_{n,p,J}]\) is included only as a
distributional consequence of the same Beta lower bound.
\end{remark}

\newpage

\newpage

\providecommand{\ExpPar}[1]{\par\noindent\textbf{#1.}\quad}

\providecommand{\ExpPar}[1]{\par\smallskip\noindent\textbf{#1.}\quad}

\providecommand{\ExpPar}[1]{\par\smallskip\noindent\textbf{#1.}\quad}

\begingroup
\setlength{\textfloatsep}{7pt plus 2pt minus 2pt}
\setlength{\floatsep}{7pt plus 2pt minus 2pt}
\setlength{\intextsep}{7pt plus 2pt minus 2pt}
\setlength{\abovecaptionskip}{4pt}
\setlength{\belowcaptionskip}{0pt}
\captionsetup[figure]{labelfont=it,textfont=it}

\section{Experimental Appendix: Proxy Tree-Angle Evaluation}
\label{app:tree-angle-experiments}

This appendix tests the first-iteration proxy comparison from
\cref{rem:first-iteration-proxy-comparison}. For a probabilistic first-iteration model
\(\mathsf M\) and a tree-fitting rule \(\mathsf A\), the experimental analogue of the
conjectural comparison is
\[
\Delta_{J,\mathsf A}(\mathsf M)
:=
\mathbb E_{\mathsf M}\!\left[
\Theta_{J,\mathsf A}(X;f_1^{\mathrm N}\mid H_0)^2
-
\mathbb E_{U_{H_0}\mid X,H_0}
\Theta_{J,\mathsf A}(X;U_{H_0}\mid H_0)^2
\right]\ge 0.
\]
Here the second term is the uncoupled metric-isotropic baseline in the same Hessian geometry. The
comparison is not meant to hold for arbitrary signals: we test it on \emph{tree-readable} structured
models, meaning signals generated from low-complexity coordinate functions which shallow
axis-aligned trees can plausibly approximate. The Rademacher model serves as an unstructured
control.

\ExpPar{Data model and algorithmic tree cosines}
Throughout, \(X_1,\dots,X_n\stackrel{\mathrm{iid}}{\sim}\mathrm{Unif}([0,1]^p)\),
\(X=\{X_1,\dots,X_n\}\), \(\mathcal H_X:=\mathbb R^X\), and
\(\langle f,h\rangle_X:=n^{-1}\sum_{x\in X}f(x)h(x)\). All empirical Hessians in these
experiments are diagonal. For \(A=\operatorname{diag}(a_i)\), write
\(\langle f,h\rangle_A:=n^{-1}\sum_i a_i f(x_i)h(x_i)\) and
\(\|f\|_A^2:=\langle f,f\rangle_A\). Since exact maximization over all depth-\(J\) trees is not
computationally feasible, we evaluate algorithmic tree cosines. For a fitting procedure \(\mathsf A\),
let \(\widehat g_{J,\mathsf A}(t;A)\) be its fitted depth-\(J\) tree approximation to
\(t\in\mathcal H_X\), computed in the \(A\)-metric, and define
\[
\Theta_{J,\mathsf A}(X;t\mid A)
:=
\frac{\langle t,\widehat g_{J,\mathsf A}(t;A)\rangle_A}
{\|t\|_A\|\widehat g_{J,\mathsf A}(t;A)\|_A},
\]
with value \(0\) if \(\widehat g_{J,\mathsf A}(t;A)=0\). For the weighted least-squares tree fits
used below, equivalently
\(\Theta_{J,\mathsf A}(X;t\mid A)^2
=
1-\|t-\widehat g_{J,\mathsf A}(t;A)\|_A^2/\|t\|_A^2\).
We use two fitting procedures: the dyadic construction from
Algorithm~\ref{alg:dyadic-tree-benchmark}, which is theorem-aligned, and a greedy CART
regression tree fitted with sample weights \(a_i\).

\ExpPar{Coupled and uncoupled targets}
For each sampled problem instance, compute \(g_0:=\nabla L(F_0)\),
\(H_0:=\nabla^2L(F_0)\), and \(f_1^{\mathrm N}:=-H_0^{-1}[g_0]\). We compare
\[
\theta^{2,\mathrm{cpl}}_{J,\mathsf A}
:=
\Theta_{J,\mathsf A}(X;f_1^{\mathrm N}\mid H_0)^2
\]
with the metric-matched proxy
\[
\theta^{2,\mathrm{prox}}_{J,\mathsf A}
:=
\frac1{M_{\mathrm{prox}}}
\sum_{m=1}^{M_{\mathrm{prox}}}
\Theta_{J,\mathsf A}(X;U_{H_0}^{(m)}\mid H_0)^2,
\qquad
U_{H_0}^{(m)}\mid(X,H_0)\sim\mathrm{Unif}(S_{H_0}(\mathcal H_X)).
\]
If \(H_0=\operatorname{diag}(h_i)\), we sample \(U_{H_0}\) by drawing
\(z_i\stackrel{\mathrm{iid}}{\sim}N(0,1)\), setting \(u_i=z_i/\sqrt{h_i}\), and normalizing so that
\(\|u\|_{H_0}=1\).

\ExpPar{Label and loss models}
We use \(n=64\), \(p=16\), and \(J=0,\dots,6\). For \(v=(v_1,\dots,v_n)\), write
\(\bar v:=n^{-1}\sum_i v_i\) and
\(\operatorname{std}(v)_i:=(v_i-\bar v)/(n^{-1}\sum_j(v_j-\bar v)^2)^{1/2}\). Define
\[
c:=\frac{(1,2,\dots,p)^\top}{\|(1,2,\dots,p)\|_2},
\qquad
s_{\mathrm{lin}}:=\operatorname{std}\bigl((c^\top x_i)_{i=1}^n\bigr),
\]
and
\[
s_{\mathrm{sm}}
:=
\operatorname{std}\bigl(
(\sin(2\pi x_i^{(1)})
+0.75\cos(2\pi x_i^{(2)})
+0.50\sin(4\pi x_i^{(3)}))_{i=1}^n
\bigr).
\]
For squared loss, \(\ell_{\mathrm{sq}}(z,y)=\frac12(z-y)^2\), we set \(F_0=0\), hence
\(H_0=I\) and \(f_1^{\mathrm N}=y\), and use
\(y_i\stackrel{\mathrm{iid}}{\sim}\mathrm{Unif}\{-1,1\}\),
\(y_i=(s_{\mathrm{lin}})_i\), and \(y_i=(s_{\mathrm{sm}})_i\). For binary logistic loss,
\(\ell_{\mathrm{log}}(z,y)=\log(1+\exp z)-yz\), \(y\in\{0,1\}\), with
\(\operatorname{sig}(z)=(1+e^{-z})^{-1}\), we use
\(y_i\stackrel{\mathrm{iid}}{\sim}\operatorname{Bernoulli}(1/2)\),
\(y_i\sim\operatorname{Bernoulli}(\operatorname{sig}(2(s_{\mathrm{lin}})_i))\), and
\(y_i\sim\operatorname{Bernoulli}(\operatorname{sig}(2(s_{\mathrm{sm}})_i))\), independently
conditional on \(X\). The logistic warm start is
\(F_0(x_i)=1.5\,\operatorname{std}(0.75s_{\mathrm{lin}}+0.75s_{\mathrm{sm}}+0.15\xi)_i\),
with \(\xi_i\stackrel{\mathrm{iid}}{\sim}N(0,1)\), so
\(h_i=\operatorname{sig}(F_0(x_i))(1-\operatorname{sig}(F_0(x_i)))\) is nonconstant.

\ExpPar{Dyadic constructive benchmark}
The proof of \cref{lem:tree-dyadic-lower-bound} uses an explicit depth-\(J\) tree obtained by
sorting the sample by the first coordinate and recursively bisecting into consecutive blocks. In the
metric setting, the leaf values are weighted averages.

\begin{algorithm}[!htbp]
\small
\caption{Dyadic constructive tree benchmark}
\label{alg:dyadic-tree-benchmark}
\begin{algorithmic}[1]
\STATE \textbf{Input:} Sample \(X=\{x_1,\dots,x_n\}\subset[0,1]^p\), target \(t\in\mathcal H_X\), metric weights \(a_i>0\), depth \(J\)
\STATE \textbf{Output:} Dyadic fit \(g_J^{\mathrm{dyad}}\) and squared cosine \((\theta_J^{\mathrm{dyad}})^2\)
\STATE Relabel the sample so that \(x_1^{(1)} < x_2^{(1)} < \cdots < x_n^{(1)}\)
\STATE Initialize \(\mathcal B_0 \leftarrow \bigl\{\{1,\dots,n\}\bigr\}\)
\FOR{\(j=1,2,\dots,J\)}
    \STATE Initialize \(\mathcal B_j \leftarrow \emptyset\)
    \FOR{each block \(B\in\mathcal B_{j-1}\)}
        \IF{\(|B|=1\)}
            \STATE Add \(B\) to \(\mathcal B_j\)
        \ELSE
            \STATE Split \(B\) into two consecutive sub-blocks \(B_{\mathrm L},B_{\mathrm R}\) as evenly as possible
            \STATE Add \(B_{\mathrm L}\) and \(B_{\mathrm R}\) to \(\mathcal B_j\)
        \ENDIF
    \ENDFOR
\ENDFOR
\STATE Let \(\mathcal B_J\) be the final partition
\FOR{each block \(B\in\mathcal B_J\)}
    \STATE For every \(i\in B\), set
    \(g_J^{\mathrm{dyad}}(x_i)\leftarrow
    \bigl(\sum_{k\in B}a_k t(x_k)\bigr)/\bigl(\sum_{k\in B}a_k\bigr)\)
\ENDFOR
\STATE Set
\((\theta_J^{\mathrm{dyad}})^2
\leftarrow
\bigl(\sum_i a_i t(x_i)g_J^{\mathrm{dyad}}(x_i)\bigr)^2/
\bigl((\sum_i a_i t(x_i)^2)(\sum_i a_i g_J^{\mathrm{dyad}}(x_i)^2)\bigr)\),
with value \(0\) if \(g_J^{\mathrm{dyad}}=0\).
\end{algorithmic}
\end{algorithm}

\ExpPar{Reported experiments and uncertainty}
We report three figures. For each loss, label model, algorithm, and depth, we estimate
\(\mathbb E[\theta^{2,\mathrm{cpl}}_{J,\mathsf A}]\),
\(\mathbb E[\theta^{2,\mathrm{prox}}_{J,\mathsf A}]\), and
\(\Delta_{J,\mathsf A}:=\mathbb E[\theta^{2,\mathrm{cpl}}_{J,\mathsf A}
-\theta^{2,\mathrm{prox}}_{J,\mathsf A}]\). All curves use \(2000\) independent problem
instances and \(M_{\mathrm{prox}}=16\) proxy directions per instance. Shaded bands are
nonparametric bootstrap \(95\%\) confidence intervals for the mean, computed by resampling outer
problem instances. The curve \(\min(2^J,n)/n\) is included only as theorem reference from
\cref{thm:tree-dyadic-lower-bound}, not as an algorithmic lower bound for CART.

\ExpPar{Squared-loss proxy comparison}
Since \(F_0=0\), the squared-loss Newton direction is the label function itself and \(H_0=I\).
Thus this experiment compares structured labels to an uncoupled identity-isotropic proxy.

\begin{figure}[!tbp]
    \centering
    \includegraphics[width=.94\linewidth]{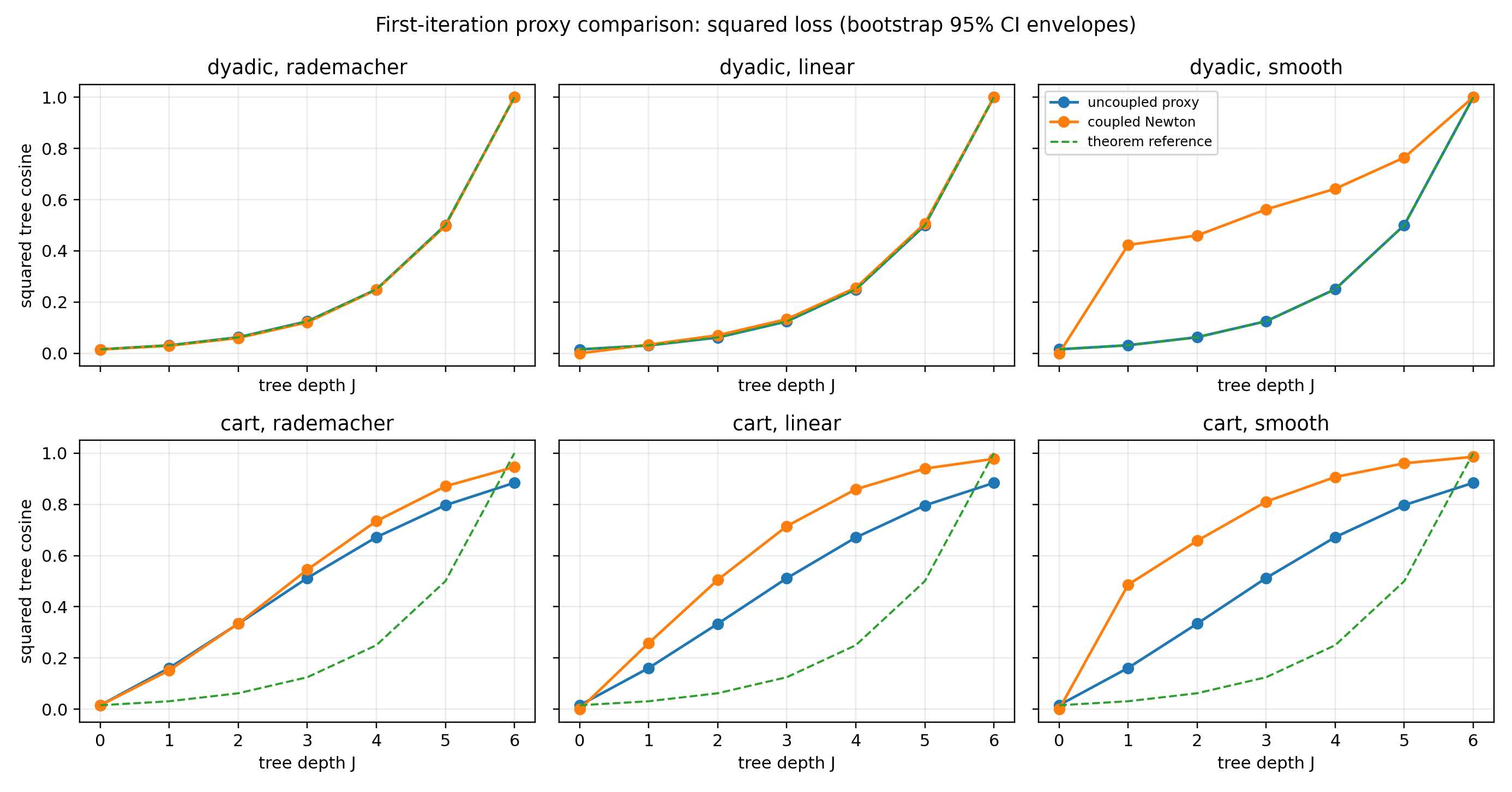}
    \caption{Squared-loss first-iteration proxy comparison for \(n=64\), \(p=16\), and
    \(J=0,\dots,6\). Rows correspond to dyadic and CART fitting; columns correspond to
    Rademacher, linear, and smooth label models. Each panel compares the coupled target
    \(\Theta_{J,\mathsf A}(X;f_1^{\mathrm N}\mid I)^2\) with the uncoupled proxy
    \(\Theta_{J,\mathsf A}(X;U_I\mid I)^2\). Shaded bands are bootstrap \(95\%\) confidence
    intervals for the mean.}
    \label{fig:tree-squared-proxy-comparison}
\end{figure}

\noindent
The Rademacher model is an unstructured control. Positive separation for the linear and smooth
models indicates that tree-readable structure improves alignment relative to an isotropic proxy.

\ExpPar{Logistic-loss metric-matched proxy comparison}
For logistic loss, the warm start gives nonconstant \(H_0\). The uncoupled baseline is therefore
sampled from the \(H_0\)-unit sphere.

\begin{figure}[!tbp]
    \centering
    \includegraphics[width=.94\linewidth]{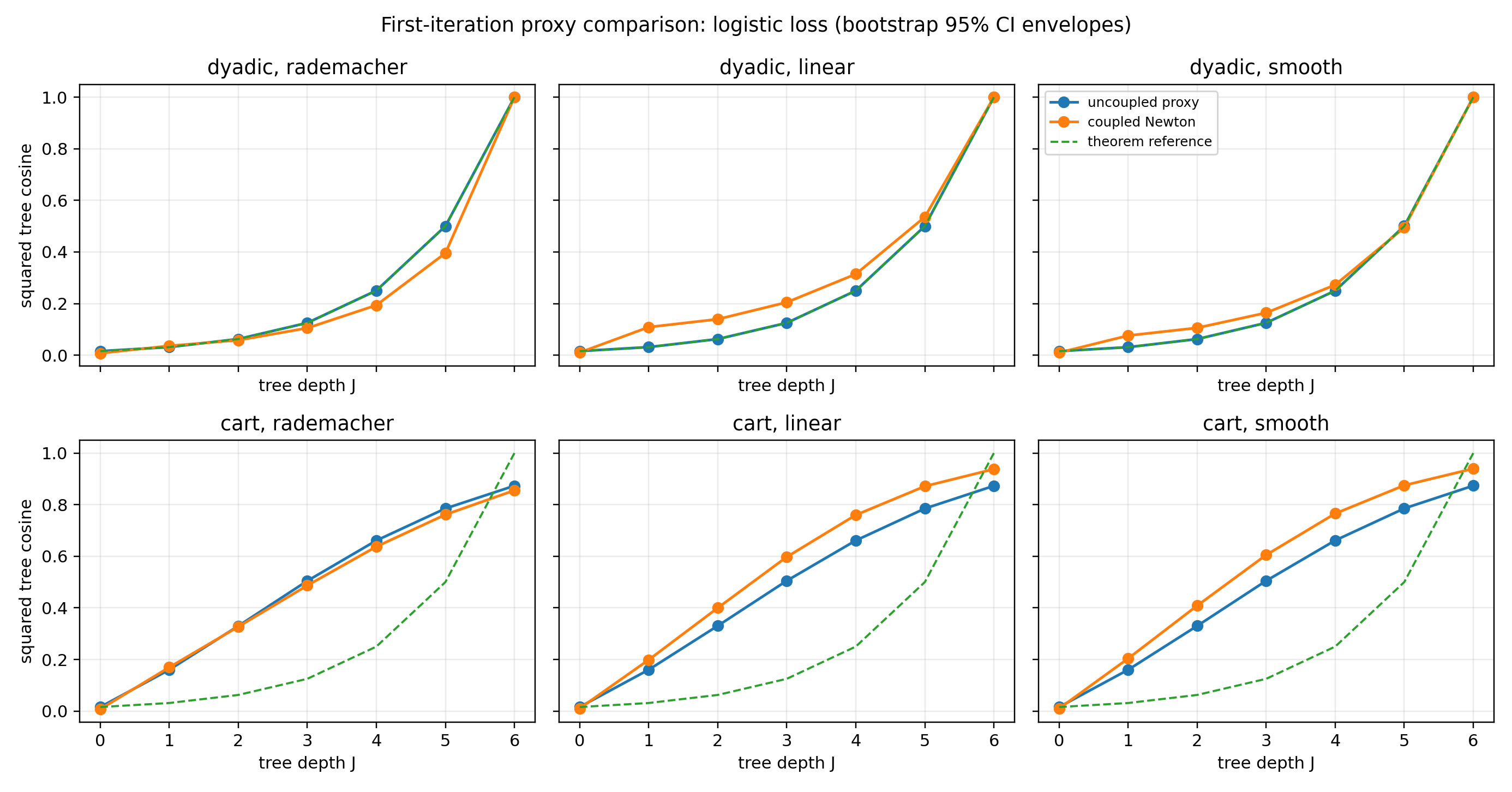}
    \caption{Binary logistic-loss first-iteration proxy comparison for \(n=64\), \(p=16\), and
    \(J=0,\dots,6\). Rows correspond to dyadic and CART fitting; columns correspond to
    Rademacher, linear, and smooth label models. Each panel compares the coupled Newton direction
    \(\Theta_{J,\mathsf A}(X;f_1^{\mathrm N}\mid H_0)^2\) with the uncoupled metric-matched proxy
    \(\Theta_{J,\mathsf A}(X;U_{H_0}\mid H_0)^2\). Shaded bands are bootstrap \(95\%\) confidence
    intervals for the mean.}
    \label{fig:tree-logistic-proxy-comparison}
\end{figure}

\noindent
This tests the proxy comparison in nontrivial Hessian geometry. The adaptive CART fits show a clear
positive gap for the structured label models, while the unstructured control remains close to the proxy.

\ExpPar{Coupling advantage}
Finally, we plot the paired difference
\(\Delta_{J,\mathsf A}
=
\mathbb E[\theta^{2,\mathrm{cpl}}_{J,\mathsf A}-\theta^{2,\mathrm{prox}}_{J,\mathsf A}]\).
This is the finite-sample algorithmic version of the proxy comparison above; positive values indicate
that the coupled first Newton direction is more tree-alignable than the uncoupled metric-isotropic
baseline.

\begin{figure}[!tbp]
    \centering
    \includegraphics[width=.94\linewidth]{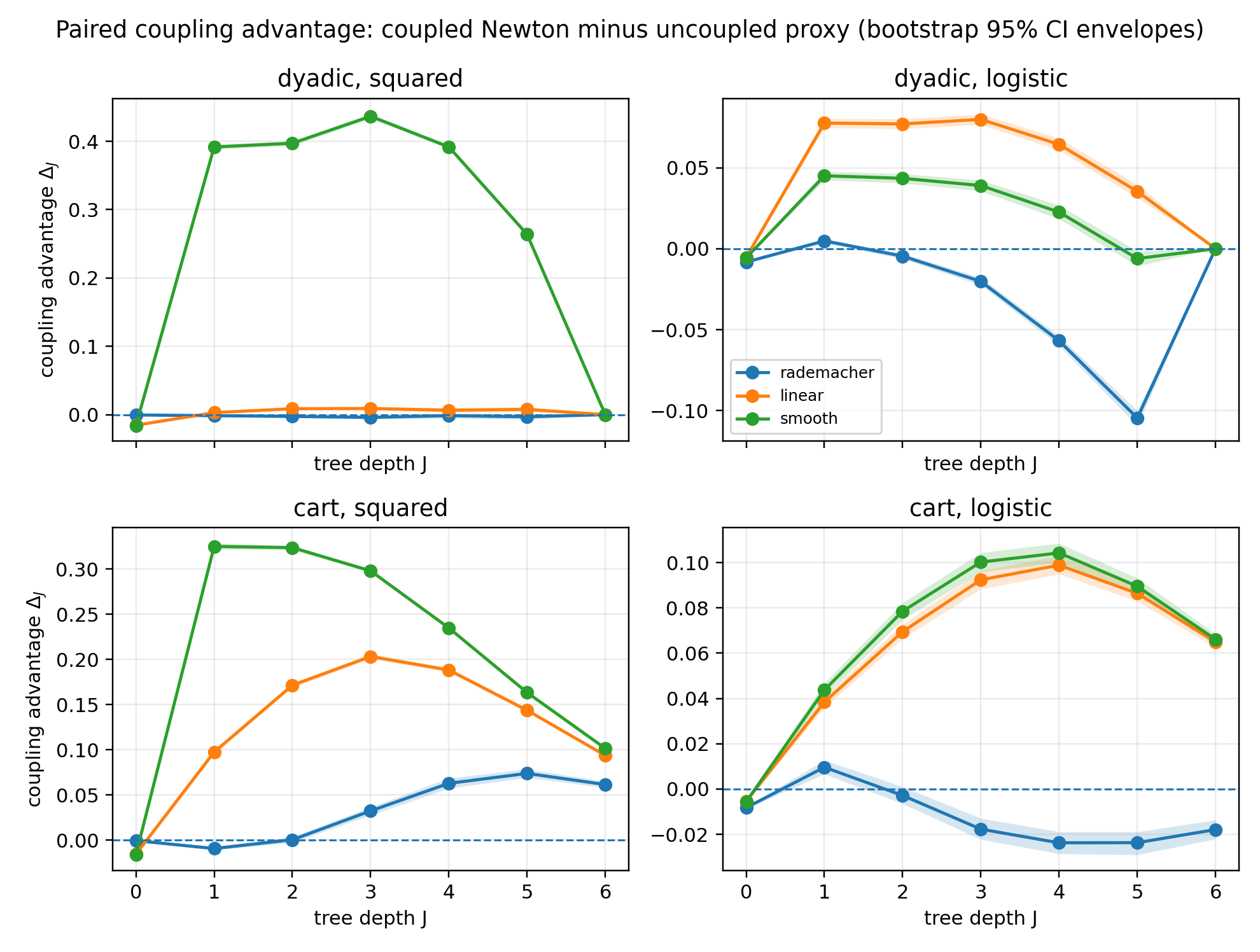}
    \caption{Coupling advantage \(\Delta_{J,\mathsf A}\) as a function of tree depth. Rows correspond
    to dyadic and CART fitting; columns correspond to squared and logistic losses. Shaded bands are
    bootstrap \(95\%\) confidence intervals for the mean paired difference. Positive values indicate
    that the coupled first Newton direction is more tree-alignable than an uncoupled metric-isotropic
    direction in the same geometry.}
    \label{fig:tree-coupling-advantage}
\end{figure}

\noindent
Taken together, \cref{fig:tree-squared-proxy-comparison,fig:tree-logistic-proxy-comparison,fig:tree-coupling-advantage}
support this first-iteration proxy comparison in the intended tree-readable regime. The Rademacher
controls remain close to the uncoupled proxy, while the linear and smooth models show a clear
positive coupling advantage, especially for the adaptive CART fits. The dyadic method is
theorem-aligned but rigid, and therefore mainly benefits when the signal aligns with its fixed
first-coordinate splitting rule. Thus the experiments support the proxy comparison as a
model-dependent structured-signal phenomenon, not as a universal statement over all targets,
all fitting rules, or later optimization iterations.

\FloatBarrier
\endgroup

\clearpage
\section{Alternative Decision Tree Angle Bound}

\begin{theorem}
\label{thm:alternative_tree_angle}
Consider the decision tree specialization in \cref{sec:gradient-boosting} and \cref{app:fixed-sample-tree-geometry}. Fix $X$, a positive definite Hessian $H_k$, and a depth-$J$ tree $T^*$ with exactly $2^J$ non-empty leaves, where $2^J<n$. Suppose $f^{\mathrm N}_{k+1}=\mathfrak{t}+\epsilon$, where $\mathfrak{t}\in V_{T^*}$ is a fixed tree-aligned signal and $\epsilon$ has independent $\mathcal N(0,\sigma^2)$ coordinates in an $H_k$-orthonormal basis, with $\sigma>0$.
Then the weak learner cosine angle $\Theta_k$ satisfies
\begin{equation}
    \mathbb{E} \left[\Theta_k^2\right] \ge \mathbb{E} \left[ \frac{\|\mathfrak{t} + \epsilon_V\|_{H_k}^2}{\|\mathfrak{t} + \epsilon_V\|_{H_k}^2 + B} \right] \ge \mathbb{E} \left[ \frac{A}{A + B} \right] = \frac{2^J}{n},
\end{equation}
where $A \sim \sigma^2 \chi^2_{m}$ and $B \sim \sigma^2 \chi^2_{n-m}$ are independent scaled chi-squared variables with $m = \dim V_{T^*} = 2^J$, and $\epsilon_V = \Pi_{T^*}^{H_k} \epsilon$ is the projected noise. Additionally, if $\mathfrak{t} = 0$, then $\Theta_k^2$ dominates a random variable following a Beta distribution:
\begin{equation}
    \Theta_k^2 \ge Z, \quad \text{where} \quad Z \sim \operatorname{Beta}\left(2^{J-1}, \frac{n - 2^J}{2}\right).
\end{equation}
\end{theorem}

\begin{proof}
All analysis and projections take place in the Hessian-induced norm $\| \cdot \|_{H_k}$.
Following the discussion in \cref{app:fixed-sample-tree-geometry}, $\Theta_k$ can be expressed by projecting the exact Newton direction to the tree subspace that maximizes the projection of the exact Newton step, yielding:
\begin{equation}
    \Theta_k^2 = \max_{T \in \mathcal{T}_J(X)} \frac{\|\Pi_T^{H_k} f^N_{k+1}\|_{H_k}^2}{\|f^N_{k+1}\|_{H_k}^2} \ge \frac{\|\Pi_{T^*}^{H_k} f^N_{k+1}\|_{H_k}^2}{\|f^N_{k+1}\|_{H_k}^2},
\end{equation}
where $T^*$ is the fixed tree from the hypothesis, and $\Pi_T^{H_k}$ denotes orthogonal projection in the $H_k$-inner product. By assumption, $T^*$ has exactly $2^J$ non-empty leaves, so $\dim V_{T^*}=2^J$.

We decompose the noise $\epsilon$ into components within the subspace $V_{T^*}$ and its $H_k$-orthogonal complement $V_{T^*}^\perp$:
\begin{equation}
    \epsilon = \epsilon_V + \epsilon_\perp, \quad \text{where} \quad \epsilon_V = \Pi_{T^*}^{H_k} \epsilon, \quad \epsilon_\perp = \Pi_{V_{T^*}^\perp}^{H_k} \epsilon.
\end{equation}
Because $\epsilon$ has independent $\mathcal N(0,\sigma^2)$ coordinates in an $H_k$-orthonormal basis, the orthogonal projections $\epsilon_V$ and $\epsilon_\perp$ are independent. Letting $m = \dim V_{T^*} = 2^J$, their squared $H_k$-norms follow scaled chi-squared distributions:
\begin{equation}
    A := \|\epsilon_V\|_{H_k}^2 \sim \sigma^2 \chi^2_m, \quad B := \|\epsilon_\perp\|_{H_k}^2 \sim \sigma^2 \chi^2_{n-m}.
\end{equation}
Since the signal $\mathfrak{t}$ lies entirely in $V_{T^*}$, we have $\Pi_{T^*}^{H_k} f^N_{k+1} = \mathfrak{t} + \epsilon_V$. Consequently, the total squared norm decomposes as $\|f^N_{k+1}\|_{H_k}^2 = \|\mathfrak{t} + \epsilon_V\|_{H_k}^2 + \|\epsilon_\perp\|_{H_k}^2$. Substituting this into our lower bound for $\Theta_k^2$ yields
\begin{equation}
    \Theta_k^2 \ge \frac{\|\mathfrak{t} + \epsilon_V\|_{H_k}^2}{\|\mathfrak{t} + \epsilon_V\|_{H_k}^2 + \|\epsilon_\perp\|_{H_k}^2} = \frac{\|\mathfrak{t} + \epsilon_V\|_{H_k}^2}{\|\mathfrak{t} + \epsilon_V\|_{H_k}^2 + B}.
\end{equation}
To bound the expectation of this ratio, observe that since $B > 0$ almost surely, the scalar function $b(a) = \frac{a}{a + B}$ is increasing for $a \ge 0$. For any realization of the signal $\mathfrak{t}$, the term $\|\mathfrak{t} + \epsilon_V\|_{H_k}^2$ follows a scaled non-central chi-squared distribution $\sigma^2 \chi^2_m(\lambda)$ with non-centrality parameter $\lambda = \|\mathfrak{t}\|_{H_k}^2 / \sigma^2$. A well-known property of the non-central chi-squared distribution is that it stochastically dominates the central chi-squared distribution. Because $b(a)$ is an increasing function, this stochastic dominance preserves the inequality in expectation:
\begin{equation}
    \mathbb{E}\left[ \Theta_k^2 \right] \ge \mathbb{E}\left[ \frac{\|\mathfrak{t} + \epsilon_V\|_{H_k}^2}{\|\mathfrak{t} + \epsilon_V\|_{H_k}^2 + B} \right] \ge \mathbb{E}\left[ \frac{A}{A + B} \right].
\end{equation}
Since $A$ and $B$ are independent scaled chi-squared variables, the ratio $Z := \frac{A}{A+B}$ follows a Beta distribution: $ Z \sim \operatorname{Beta}\left(\frac{m}{2}, \frac{n-m}{2}\right) = \operatorname{Beta}\left(2^{J-1}, \frac{n-2^J}{2}\right)$. The expected value of $Z$ is $\frac{m/2}{m/2 + (n-m)/2} = \frac{2^J}{n}$, completing the proof of the general expectation bound.

Finally, for the special case where there is no signal ($\mathfrak{t} = 0$), the ratio instead directly evaluates to the Beta-distributed random variable $Z$:
\begin{equation}
    \Theta_k^2 \ge \frac{A}{A + B} =: Z \sim \operatorname{Beta}\left(2^{J-1}, \frac{n-2^J}{2}\right).
\end{equation}
\end{proof}

\end{document}